\documentclass[lettersize,journal]{IEEEtran}
\usepackage{subfigure}
\usepackage{amsmath,amsfonts}
\usepackage{algorithmic}
\usepackage{algorithm}
\usepackage{array}
\usepackage{textcomp}
\usepackage{stfloats}
\usepackage{url}
\usepackage{verbatim}
\usepackage{graphicx}
\usepackage{cite}
\usepackage{hyperref}
\usepackage{cleveref}

\usepackage{bm}
\usepackage{amsmath,amssymb,amsthm}
\usepackage{multirow,array}
\usepackage{color}
\usepackage{algorithmic,algorithm}
\usepackage{lipsum}
\usepackage{graphicx}

\usepackage{diagbox}
\usepackage{bbding}

\usepackage{amsmath,amsthm}
\usepackage{amssymb,amsfonts}
\usepackage[utf8]{inputenc} % allow utf-8 input
\usepackage[T1]{fontenc}    % use 8-bit T1 fonts
\usepackage{hyperref}       % hyperlinks
\usepackage{url}            % simple URL typesetting
\usepackage{booktabs}       % professional-quality tables
\usepackage{amsfonts}       % blackboard math symbols
\usepackage{nicefrac}       % compact symbols for 1/2, etc.
\usepackage{microtype}      % microtypography
\usepackage[table,xcdraw]{xcolor}       % colors
\usepackage{amsmath}
\usepackage{bm}
\usepackage{array}
\usepackage{caption}
\usepackage{cleveref}
\usepackage{array}
\usepackage{gensymb}
\usepackage{multicol}
\usepackage{amsfonts,amssymb}
\usepackage{multirow}
\usepackage{enumitem}
\usepackage{graphicx}% microtypography
\usepackage{adjustbox} 
\usepackage{amsmath}
\usepackage{pifont} 
\usepackage{graphicx}
\usepackage{caption}
\usepackage{threeparttable}
\usepackage{colortbl}

\PassOptionsToPackage{numbers, compress}{natbib}
\begin{document}

\title{3D Gaussian Splatting: Survey, Technologies, Challenges, and Opportunities}

\author{Yanqi~Bao,
        Tianyu~Ding,
        {Jing~Huo}*,
        Yaoli~Liu,
        Yuxin~Li,
        Wenbin~Li,
        Yang~Gao,~\IEEEmembership{Member,~IEEE,}
        and~Jiebo~Luo,~\IEEEmembership{Fellow,~IEEE,}
        % <-this % stops a space
\thanks{Yanqi Bao, Jing Huo, Yaoli Liu, Yuxin Li, Wenbin Li and Yang Gao are with the State Key Laboratory for Novel Software Technology, Nanjing University, China, 210023 (e-mail: \{yq\_bao, yaoliliu, liyuxin16\}@smail.nju.edu.cn; \{huojing, liwenbin, gaoy\}@nju.edu.cn).}% <-this % stops a space
\thanks{Tianyu Ding is with the Applied Sciences Group, Microsoft Corporation, Redmond, USA (e-mail: tianyuding@microsoft.com).}
\thanks{Jiebo Luo is with the Department of Computer Science, University of Rochester, America (e-mail: jluo@cs.rochester.edu).}
\thanks{*Corresponding authors: Jing Huo.}}

\markboth{Journal of \LaTeX\ Class Files,~Vol.~14, No.~8, September~2024}%
{Shell \MakeLowercase{\textit{et al.}}: A Sample Article Using IEEEtran.cls for IEEE Journals}


\maketitle

\begin{abstract}
3D Gaussian Splatting (3DGS) has emerged as a prominent technique with the potential to become a mainstream method for 3D representations. It can effectively transform multi-view images into explicit 3D Gaussian through efficient training, and achieve real-time rendering of novel views. This survey aims to analyze existing 3DGS-related works from multiple intersecting perspectives, including related tasks, technologies, challenges, and opportunities. The primary objective is to provide newcomers with a rapid understanding of the field and to assist researchers in methodically organizing existing technologies and challenges. Specifically, we delve into the optimization, application, and extension of 3DGS, categorizing them based on their focuses or motivations. Additionally, we summarize and classify nine types of technical modules and corresponding improvements identified in existing works. Based on these analyses, we further examine the common challenges and technologies across various tasks, proposing potential research opportunities.
\end{abstract}


\section{Introduction}\label{sec:introduction}
\IEEEPARstart{T}{he} advent of Neural Radiance Fields (NeRF)~\cite{mildenhall2021nerf} has ignited considerable interest in the pursuit of photorealistic 3D content. Despite substantial recent advancements that have markedly enhanced NeRF's potential for practical applications, its inherent efficiency challenges have remained unresolved. The introduction of 3D Gaussian Splatting (3DGS) has decisively addressed this bottleneck, enabling high-quality real-time ($\geq$30 fps) novel view synthesis at 1080p resolution. This rapid development has quickly attracted researchers and led to a proliferation of related works.

Owing to the efficiency and controllable explicit representation of 3DGS, its applications extend across a diverse array of fields. These include enhancing immersive environments in Virtual Reality (VR) and Augmented Reality (AR), improved spatial awareness in robotics and autonomous systems, and urban planning and architecture, etc.

\begin{figure}
\begin{center}
\includegraphics[width=0.5\textwidth]{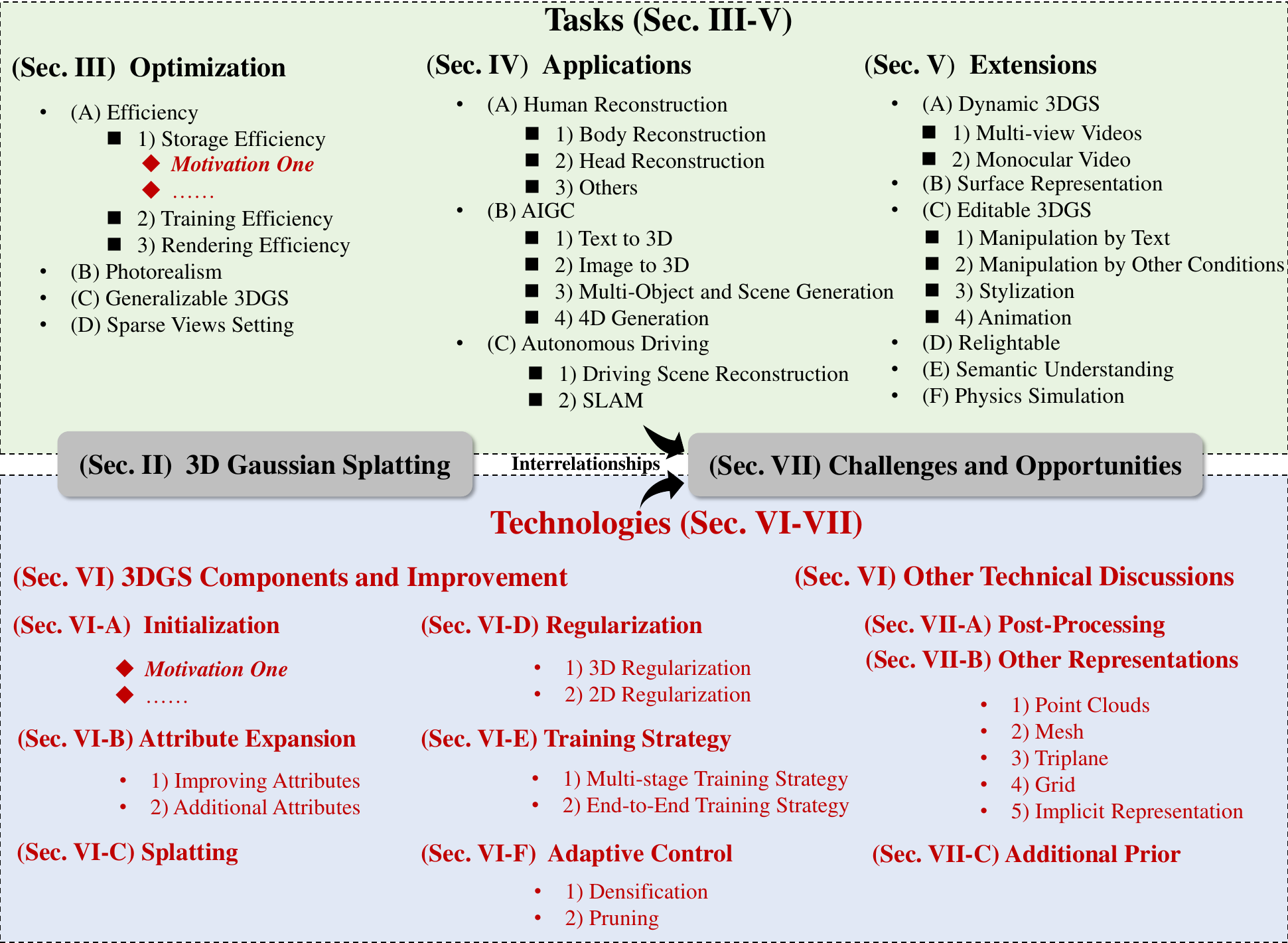}
\end{center}
\vspace{-0.3cm}
\caption{{The introduction of this survey, with the \textbf{RED} parts indicating the new content compared to existing reviews.}}
\label{fig0}
\vspace{-0.5cm}
\end{figure}

{To assist readers in quickly grasping the progress in 3DGS research, we provide a comprehensive survey of 3DGS and its derivative works. This survey systematically compiles the most important and recent literature on the subject, offering detailed classifications and discussions of their tasks and techniques. In examining the technological commonalities across numerous 3DGS variants, we present a structured analysis of technical improvements in fundamental components of vanilla 3DGS, such as initialization, attribute configurations and regularization, etc. Based on this summary of techniques, we aim to help readers synthesize the connections among different improved techniques and provide approaches to enhance various components of vanilla 3DGS to meet their customized tasks. Moreover, we conduct a systematic investigation into the relationships between downstream tasks and their enabling technologies in 3DGS, identifying and analyzing four fundamental challenges. Through careful examination of these challenges, we propose promising research directions to advance this rapidly evolving field, providing a roadmap for future innovations in 3DGS.}

{Although some comprehensive reviews have documented recent advances in 3DGS\cite{chen2024survey,wu2024recent,fei20243d}, they focus on categorizing and discussing existing works by downstream tasks, overlooking technical connections between different tasks, which leads to redundant discussions. Our work is distinctive in providing discussions at two levels: Tasks and Techniques. Specifically, we categorize and discuss existing downstream tasks according to their different motivations or focuses, rather than reviewing all works sequentially. More significantly, we present a thorough examination of \textbf{technical improvements} implemented across various modules of the vanilla 3DGS by existing variants, enabling readers to establish clear relationships between different research fields that share similar methodological foundations. Building upon these, we further investigate the their underlying \textbf{commonalities} and delineate core challenges and opportunities, as shown in Fig.~\ref{fig0}. Through this approach, we aim to synthesize recent technical breakthroughs in the 3DGS field and direct researchers' attention to the core unique challenges facing 3DGS. The primary contributions of this survey can be summarized as follows}:

\begin{enumerate}[leftmargin=12pt]
    \item {This survey discusses 3DGS and its various derivative tasks, including the optimization, application, and extension of 3DGS. We provide a classification and discussion based on focuses or motivations, which enables readers to gain a more comprehensive understanding of tasks and establish research directions.}
    \item {\textbf{More importantly}, we comprehensively analyze the enhancements in various techniques within 3DGS across different works, offering detailed classifications and in-depth discussions. This enables readers to discern the interrelationships among various improved techniques, thereby assisting in their application to customized tasks.}
    \item Based on the analysis of existing works and techniques, we identify the commonalities and associations among 3DGS-related tasks and summarize the core challenges.
    \item In addressing common challenges, this survey elucidates potential opportunities and provides insightful analyses.
    \item We have published an open-source project on GitHub that compiles 3DGS-related articles, and will continue to add new works and technologies into this project. \url{https://github.com/qqqqqqy0227/awesome-3DGS}. We hope that researchers can use it to access the latest research information.
\end{enumerate}

\section{Preliminaries}\label{3D Gaussian Splatting}

In this section, we primarily introduce 3DGS to the readers. More background, including Neural Implicit Fields~\cite{liao2024ov} and Point-based Rendering~\cite{zhang2021progressive}, is described in \textbf{Supplementary-A}.

3DGS~\cite{kerbl20233d} combines the advantages of neural implicit field and point-based rendering methods, achieving the high-fidelity rendering quality of the former while maintaining the real-time rendering capability of the latter. Specifically, as shown in Fig.~\ref{fig12}, 3DGS defines points in the point cloud as 3D Gaussian primitives with volumetric density:\\
\begin{equation}
    G(\bm{x})=\exp{\left( -\frac{1}{2} (\bm{x})^T \bm{\Sigma^{-1}} (\bm{x}) \right)},
\end{equation}
where $\bm{\Sigma}$ is the 3D covariance matrix and $\bm{x}$ is the position from the point (Gaussian mean) $\bm{\mu}$. To ensure the semi-positive definiteness of the covariance matrix, 3DGS reparameterizes the covariance matrix as a combination of a rotation matrix $\bm{R}$ and a scaling matrix $\bm{S}$:\\
\begin{equation}
    \bm{\Sigma} = \bm{R} \bm{S} \bm{S}^T \bm{R}^T,
\end{equation}
where the 3D scaling matrix $\bm{S}$ can be represented by a 3D vector $\bm{s}$, and the rotation matrix $\bm{R}$ is obtained through a learnable quaternion $\bm{q}$, resulting in a total of 7 learnable parameters. Compared to the commonly employed Cholesky decomposition, which guarantees the semi-positive-definiteness of matrices, the reparameterization method utilized by 3DGS, albeit introducing an additional learnable parameter, facilitates the imposition of geometric constraints on Gaussian primitives (e.g., constraining the scaling vector to give Gaussian primitives a flattened characteristic). In addition to geometric attributes, each Gaussian primitive also stores an opacity $\alpha$ and a set of learnable Spherical Harmonic (SH) parameters to represent view-dependent appearance. Thus, the collection of all primitives can be regarded as a discretized representation that only stores the non-empty parts of the neural field.

At the beginning of training, the initial Gaussian primitives are either initialized from a sparse point cloud provided by Structure-from-Motion or randomly initialized. The initial number of Gaussian primitives may be insufficient for high-quality scene reconstruction; hence, 3DGS offers a method for adaptively controlling Gaussian primitives. This method evaluates whether a primitive is "under-reconstructed" or "over-reconstructed" by observing the gradient of each Gaussian primitive's position attributes in view space. Based on this evaluation, the method increases the number of Gaussian primitives by cloning or splitting the primitives to enhance scene representation capability. Additionally, the opacity of all Gaussian primitives is periodically reset to zero to mitigate the presence of artifacts during the optimization process. This adaptive process allows 3DGS to start optimization with a smaller initial set of Gaussians, thus alleviating the dependency on dense point clouds that previous point-based differentiable rendering methods required.

During rendering, 3DGS projects 3D Gaussian primitives onto the 2D imaging plane using the EWA splatting method \cite{zwicker2001ewa} and employs $\alpha$ blending to compute the final pixel color. This process is described in detail in the \textbf{Supplementary-A}.

\section{Optimization of 3D Gaussian Splatting}\label{Optimization}
\subsection{Efficiency}\label{Efficiency}
Efficiency is one of the core metrics for evaluating 3D reconstruction~\cite{ding2024ray}. In this section, we describe it from three perspectives: storage, training, and rendering efficiency. 
\subsubsection{Storage Efficiency}\label{Storage Efficiency}
3DGS requires millions of different Gaussian primitives to fit the geometry and appearance in a scene, leading to high storage overhead: a typical reconstruction of an outdoor scene often requires several hundred megabytes to multiple gigabytes of explicit storage space. Given that the geometric and appearance attributes of different Gaussian primitives may be highly similar, storing attributes for each primitive individually can lead to potential redundancy. Some quantitative reconstruction results are reported in Table~\ref{compression comparison}.

{Existing works~\cite{girish2023eagles, navaneet2023compact3d, niedermayr2023compressed} primarily focus on applying \textit{\textbf{Vector Quantization~\cite{equitz1989new} (VQ) techniques}} to compress 3DGS.}

{Among them, Compact3D~\cite{navaneet2023compact3d} applies VQ to compress different attributes into four corresponding codebooks and stores the index of each Gaussian in these codebooks to reduce the storage overhead. After establishing the codebooks, the training gradients are copied and backpropagated to the original non-quantized Gaussian parameters via the codebooks, updating both the quantized and non-quantized parameters, and discarding the non-quantized parameters when the training is done. Despite achieving efficient 3DGS compression, these methods inevitably encounter quantization errors following discretization and remain sensitive to hyperparameter configurations.}

% Additionally, Compact3D employs run-length coding to compress the sorted index values further, thereby enhancing storage efficiency. 

\begin{figure*}
\begin{center}
\includegraphics[width=0.8\textwidth]{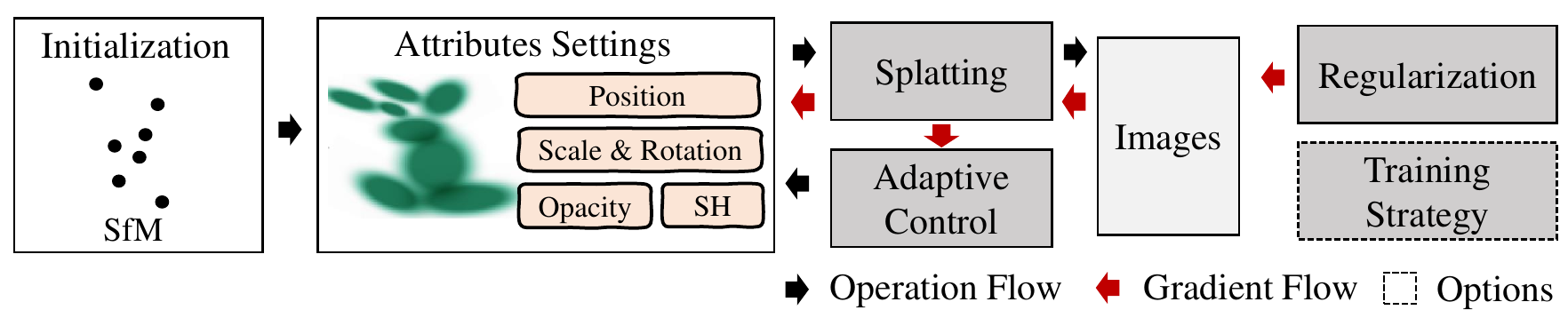}
% \vspace{-0.7cm}
\end{center}
\caption{Pipeline and Technologies of 3D Gaussian Splatting.}
\label{fig12}
% \vspace{-0.6cm}
\end{figure*}

% Similarly, the work by Niedermayr et al.\cite{niedermayr2023compressed} proposes a sensitivity-aware vector quantization technique to construct codebooks based on sensitivity-aware k-means\cite{sculley2010web} and utilizes the DEFLATE compression algorithm~\cite{deutsch1996deflate} to further compress the trained codebooks. After training, it proposes a quantization-aware fine-tuning strategy to recover information lost due to VQ.

{Furthermore, some works~\cite{lee2023compact, fan2023lightgaussian} aim at \textit{\textbf{developing efficient pruning strategies.}}}

LightGaussian~\cite{fan2023lightgaussian} introduces a Gaussian pruning strategy based on the global significance score and a distillation strategy for high-degree spherical harmonic parameters. Similarly, the work by Lee et al.\cite{lee2023compact} introduces a learnable mask to reduce the number of original Gaussians. {Such methods heavily rely on determining which primitives are non-essential.}

% and a unified hash-grid-based appearance field\cite{muller2022instant} to compress the color parameters. 

% Unlike the aforementioned works, Self-Organizing Gaussian~\cite{morgenstern2023compact} does not use traditional nontopological VQ codebooks for compressing large numbers of Gaussians. Instead, it employs the concept of self-organizing mapping to map Gaussian attributes into corresponding 2D grids. The topological relationships in the 2D grids reflect those in the original attribute space, allowing compression algorithms for topologically structured 2D data to be applied to the unordered Gaussian primitives.

{Furthermore, there are works~\cite{lu2023scaffold, hamdi2024ges, morgenstern2023compact, chen2025hac} focused on \textit{\textbf{improving efficient Gaussian representations or attributes.}}}

% Scaffold-GS~\cite{lu2023scaffold}, based on sparse point clouds reconstructed by COLMAP~\cite{schonberger2016structure}, introduces voxels and anchors that are the centers of voxels. It utilizes learnable offsets and neural networks to decode the positions and other attributes of the local neural Gaussians around the predefined anchors, thereby reducing redundant Gaussian primitives. Additionally, Scaffold-GS~\cite{lu2023scaffold} proposes a set of strategies for the growth and pruning of anchors based on multi-resolution voxel grids.

Scaffold-GS~\cite{lu2023scaffold} designs anchors and additional attributes for efficient representation, which have the capability to convert to 3DGS. Based on this representation, Scaffold-GS proposes a set of strategies for the growth and pruning of anchors on multi-resolution voxel grids. {Despite their widespread implementation, these approaches face inherent compression limitations due to their unstructured characteristics, which HAC~\cite{chen2025hac} later mitigated through the incorporation of structured hash grid. }

\begin{table}[t]
\caption{Comparison of compression on MipNeRF360~\cite{barron2022mip}.}
\centering
\begin{tabular}{p{3cm}p{0.5cm}p{0.5cm}p{0.6cm}>{\centering\arraybackslash}p{2cm}}
\toprule
\textbf{Method} & \textbf{PSNR↑} & \textbf{SSIM↑} & \textbf{LPIPS↓} & \textbf{Size (MB)↓} \\ 
\midrule
3DGS & 27.49 & \textbf{0.813} & \textbf{0.222} & 744.7\\ 
Scaffold-GS~\cite{lu2023scaffold} & 27.50 &   0.806 &  0.252 & 253.9\\ 
HAC~\cite{chen2025hac} & \textbf{27.53} &  0.807 &  0.238 & \textbf{15.26} \\
Compact-3DGS~\cite{lee2024compact} & 27.08 &   0.798 &  0.247 & 48.80 \\
EAGLES~\cite{girish2023eagles} & 27.15 &  0.808 & 0.238 &  68.89\\
LightGaussian~\cite{fan2023lightgaussian} & 27.00 & 0.799 &  0.249 & 44.54\\
Gaussian-SLAM~\cite{morgenstern2023compact} & 26.01 &  0.772 & 0.259 & 23.90 \\
Compact3d~\cite{navaneet2023compact3d} & 27.16 &   0.808 & 0.228 & 50.30 \\
\bottomrule
\end{tabular}
\vspace{-3mm}
\label{compression comparison}
\end{table}

% GES~\cite{hamdi2024ges} introduces the Generalized Exponential (GEF) Mixture to replace Gaussian representations, which has the capability to efficiently fit arbitrary signals. By designing a Fast Differentiable Rasterizer and Frequency-Modulated Image Loss for GEF, GES is capable of utilizing a smaller number of GEF primitives while maintaining performance.

\subsubsection{Training Efficiency}\label{Training Efficiency}

Improving training efficiency is also important for 3DGS. DISTWAR~\cite{durvasula2023distwar} introduces an advanced technique aimed at accelerating atomic operations in raster-based differentiable rendering applications, which typically encounter significant bottlenecks during gradient computation due to the high volume of atomic updates. By leveraging intra-warp locality in atomic updates and addressing the variability in atomic traffic among warps, DISTWAR implements warp-level reduction of threads at the SM sub-cores using registers. {These strategies enable DISTWAR to achieve an average 2.44× acceleration in performance.}

\subsubsection{Rendering Efficiency}\label{Rendering Efficiency}

Real-time rendering is one of the core advantages of Gaussian-based methods. Some works that improve storage efficiency can simultaneously enhance rendering performance, for example, by reducing the number of Gaussian primitives. {Several studies~\cite{jo2024identifying,lee2024gscore} have specifically addressed this issue.} After training the 3DGS, the work by~\cite{jo2024identifying} involves pre-identifying and excluding unnecessary Gaussian primitives through offline clustering based on their spatial proximity and potential impact on the final rendered 2D image. Furthermore, this work introduces a specialized hardware architecture designed to support this technique, achieving a speedup of 10.7× compared to a GPU.

% GSCore~\cite{lee2024gscore} proposes a hardware acceleration unit for optimizing the rendering pipeline of 3DGS in radiance field rendering. Based on the analysis of performance bottlenecks in Gaussian sorting and rasterization, GSCore introduces optimization techniques such as Gaussian shape-aware intersection tests, hierarchical sorting, and subtile skipping. Implementing these techniques within GSCore results in a 15.86× average speedup over mobile GPUs.

\subsection{Photorealism}\label{Photorealism}

Photorealism is also a topic worth attention~\cite{verbin2022ref}. 3DGS is expected to achieve realistic rendering in various scenarios. Some~\cite{cheng2024gaussianpro, zhang2024fregs, radl2024stopthepop} focus on \textbf{\textit{optimizing under vanilla settings.}} {Among them, GaussianPro~\cite{cheng2024gaussianpro} introduces an innovative paradigm for joint 2D-3D training. Building upon the 3D plane definition and patch matching technology, it proposes a progressive Gaussian propagation strategy, which harnesses the consistency of 3D views and projection relationships to refine the rendered 2D depth and normal maps. Although these methods demonstrate superior capabilities in handling artifacts and 3D inconsistencies compared to 3DGS, further exploration is still needed for complex geometric reconstruction. To further mitigate this issue, the work~\cite{diolatzis2024n} introduces a scalable and efficient N-dimensional Gaussian Mixture Model for fast, accurate high-dimensional modeling without domain-specific heuristics or sacrificing computational efficiency but is limited by its reliance on dense data and conservative culling.}

% FreGS~\cite{zhang2024fregs} transitions the supervision process into the frequency domain and employs amplitude and phase attributes from the 2D discrete Fourier transform to alleviate over-reconstruction in 3DGS. Based on this idea, FreGS introduces a frequency-domain-guided coarse-to-fine annealing technique to eliminate undesirable artifacts.

{The sharp decline in \textit{\textbf{multi-scale rendering performance}} is also a topic worth attention~\cite{yan2023multi, yu2024mip,song2024sa,liang2025analytic}.} {Among them, Mip-splatting~\cite{yu2024mip}, addressing the issue from the perspective of the sampling rate, introduces a Gaussian low-pass filter and 2D Mip filter based on Nyquist's theorem to constrain the frequency of the Gaussians according to the maximal sampling rate across all observed samples. Then, to address the over-smoothing issue in Mip-splatting, Analytic-Splatting~\cite{liang2025analytic} proposes a novel method that analytically approximates the integral of 2D Gaussian signals within pixel window areas through a conditioned logistic function. While it enhances detail fidelity, it comes at the expense of efficiency.}

% The work~\cite{yan2023multi} first analyzes the causes of aliasing in the frequency domain at low resolutions and in far-away rendering, utilizing Multi-Scale Gaussians to address this issue. Then, Pixel Coverage is defined to reflect the Gaussian size compared to the current pixel size. Based on this concept, it identifies small Gaussians and aggregates them into larger Gaussian for multi-scale training and selective rendering.

% Unlike modifications in the training phase, SA-GS~\cite{song2024sa} operates solely by a 2D scale-adaptive filter during test time, making it applicable to any pretrained 3DGS.

{Other works attempt to reconstruct challenging scenes, such as \textbf{\textit{reflective surfaces~\cite{jiang2024gaussianshader, meng2024mirror, yang2024spec, ye20243d, 10.1145/3680528.3687659, liu2025mirrorgaussian} and Deblurring~\cite{oh2024deblurgs, lee2025deblurring, seiskari2025gaussian, peng2025bags}.}}}

% , and Relightable~\cite{gao2023relightable}

{GaussianShader~\cite{jiang2024gaussianshader} reconstructs reflective surfaces by employing a hybrid color representation and integrating the specular GGX~\cite{walter2007microfacet} and normal estimation module, which encompasses diffuse color, direct specular reflection, and a residual color component that accounts for phenomena such as scattering and indirect light reflections. The work~\cite{10.1145/3680528.3687659} introduces a local Gaussian density mixture representation that combines geometry-guided ray interpolation with neural blending and end-to-end optimization to model high-frequency view-dependent reflections and refractions. Although these methods effectively manage complex reflective surfaces, they unavoidably sacrifice computational efficiency relative to vanilla 3DGS. Recently, the work~\cite{ye20243d} proposes a deferred shading method for 3DGS, which overcomes the challenge of normal estimation for environment map reflection and demonstrates enhanced efficiency by propagating accurate normals across neighboring Gaussians and per-pixel shading in screen space.}

% Mirror-3DGS~\cite{meng2024mirror} adds a learnable mirror attribute to determine the location of mirrors and introduces a Virtual Mirrored Viewpoint to aid in the reconstruction of mirror scenes based on the original 3DGS. And, Spec-Gaussian~\cite{yang2024spec} replaces the original 3DGS with Anisotropic Spherical Gaussian to construct scenes with specular and anisotropic components and introduces an Anchor-based representation~\cite{lu2023scaffold} for efficiency.

% Relightable 3D Gaussian (R3DG)~\cite{gao2023relightable} represents 3D scenes using relightable points, each characterized by a normal direction, BRDF parameters, and incident lighting, where the incident lights are decomposed into global and local components with view-dependent visibilities. Then, a novel point-based ray-tracing technique, based on bounding volume hierarchy, is designed in R3DG for efficient visibility baking and real-time rendering with accurate shadow effects.

{Existing deblurring approaches predominantly focus on motion blur or lens defocusing~\cite{oh2024deblurgs, lee2025deblurring, seiskari2025gaussian}, introducing blur process modeling to enable sharp reconstruction supervised by blurred images. To address a wider range of scenarios, BAGS~\cite{peng2025bags} introduces a Blur Agnostic robust modeling by a Blur Proposal Network and coarse-to-fine optimization scheme.}
% DeblurGS~\cite{oh2024deblurgs} tackles the challenge of inaccurate camera poses caused by severe blur, which optimizes sharp 3D scenes by estimating the 6-Degree-of-Freedom (6-DoF) camera motion for each blurry observation and synthesizing corresponding blurry renderings. 

\subsection{Generalizable 3DGS}\label{Generalization}

The objective of existing generalizable 3D reconstruction or novel view synthesis tasks is to leverage extensive auxiliary datasets to learn scene-agnostic representations. 

{The 3DGS's explicit representation has led to a substantial body of works focused on \textbf{\textit{using reference images to directly infer corresponding Gaussian primitives on a per-pixel basis}}, which are subsequently employed to render images from target views~\cite{szymanowicz2023splatter,charatan2023pixelsplat}. To achieve this, early works such as Splatter Image~\cite{szymanowicz2023splatter} propose a novel paradigm for converting images into Gaussian attribute images. MVSplat~\cite{chen2024mvsplat} proposes representing the cost volume using plane sweeps in 3D space and predicting the depths in sparse reference inputs, precisely locating the centers of Gaussian primitives. While demonstrating generalization ability, this technique's application is limited by its restricted synthesis range and the emergence of distractor-data. Subsequent works, FreeSplat~\cite{wang2024freesplat} and DGGS~\cite{bao2024distractor}, address these limitations through Pixel-wise Triplet Fusion strategy and distractor-free training and inference paradigms, respectively. Most recently, G3R~\cite{chen2025g3r} extends generalizable 3DGS to dynamic scenes, achieving generalized dynamic scene reconstruction through extra LiDAR data.}

{Furthermore, some studies~\cite{zou2023triplane, xu2024agg} focus on \textit{\textbf{introducing triplane}} to achieve generalization capabilities , which infers Gaussian attributes by querying triplane features. Recent works~\cite{gslrm2024,xu2024grm} seek to extend similar paradigms to large-scale 3D datasets, utilizing transformer-based architectures to enable direct 3D asset inference from sparse image inputs.}

\subsection{Sparse Views Setting}\label{Sparse Views}

% Reconstructing from sparse inputs presents significant challenges, wherein the methodology of 3DGS is fundamentally analogous to that of NeRF. It seeks to design new regularization strategies or incorporate additional information, such as depth data, to aid in supervising scene optimization. The difference is that such supplementary information or conditioning can be used not only for constructing loss functions but also, more importantly, for initializing 3DGS. This is particularly crucial under conditions of sparse views, where traditional initialization methods often prove ineffective.

Reconstructing from sparse inputs presents significant challenges, wherein the methodology of 3DGS is fundamentally analogous to that of NeRF~\cite{guo2024depth}, which aim to {\textit{\textbf{develop novel regularization strategies and integrate supplementary information, such as depth data or Diffusion model}}~\cite{chung2023depth,zhu2023fsgs,swann2024touch,li2024dngaussian,liu2024georgs,liu2025deceptive, paliwal2025coherentgs, zhang2025cor}}.

{Specifically, the incorporation of depth data is crucial for 3DGS as an explicit representation, alleviating the model's demand for spatial comprehension under sparse inputs. And multi-view trained diffusion models also can provide important prior knowledge for expanding sparse views into refined dense information. Nevertheless, these methods typically depend on the diffusion models' ability to preserve 3D consistency. Additional regularization is commonly achieved through incorporating explicit spatial distribution rules or pseudo-views supervision. In addition to these common methods, some studies have focused on the \textit{\textbf{initialization and optimization strategy}}}. GaussianObject~\cite{yang2024gaussianobject} introduces an initialization strategy based on Visual Hull and an optimization method using distance statistical data to eliminate floaters. 

% Chung et al.~\cite{chung2023depth} propose employing a monocular depth estimation model to predict depth maps, which are subsequently refined using SfM~\cite{schonberger2016structure} for precise depth range. 

% Building upon the depth supervision, FSGS~\cite{zhu2023fsgs} introduces a proximity-guided Gaussian upsampling method to increase the quantity and integrates new pseudo-views through a 2D prior model to further mitigate overfitting.

% Subsequently, Touch-GS~\cite{swann2024touch} extends this paradigm with tactile sensing in robotic perception applications. After aligning with monocular depth information, tactile sensing data effectively predict corresponding depth and uncertainty maps based on implicit surface representations, which are used to enhance the initialization and optimization processes.

% Furthermore, DNGaussian~\cite{li2024dngaussian} explores the problem from a regularization perspective, proposing two different regularization: hard depth and soft depth, to address the degradation of scene geometry. Then, DNGaussian introduces both global and local depth normalization methods to enhance sensitivity to subtle local depth variations.

\section{Applications of 3D Gaussian Splatting}\label{Applications}

% 3DGS excels in various application domains due to its efficiency and photorealistic renderings, which include digital human reconstruction, Artificial Intelligence Generated Content (AIGC), and autonomous driving, among others. Building on prior explorations, 3DGS can be directly applied as a core technology in diverse research areas, effectively replacing traditional 3D representations.

\subsection{Human Reconstruction}\label{Human Reconstruction}

% The high-speed rendering and representation capabilities of 3D Gaussian Splatting provide real-time rendering and enhanced detail optimization and motion control for tasks such as human reconstruction, animation, and human generation. 

% The application of 3DGS for digital human-related tasks, including human reconstruction, animation, and human generation, has garnered substantial attention within the research community. Recent works can be categorized according to the reconstructed parts.

\subsubsection{Body Reconstruction}

% % 定义勾和叉的命令
% \newcommand{\cmark}{\ding{51}}  % 需要 pifont 包
% \newcommand{\xmark}{\ding{55}}  % 需要 pifont 包

% % 如果上面的命令不起作用，可以尝试以下替代：
% % \newcommand{\cmark}{\checkmark}  % 需要 amssymb 包
% % \newcommand{\xmark}{$\times$}

\newcommand{\myparagraph}[1]{\noindent\textbf{#1}}

\newcommand{\cmark}{\textcolor{blue}{\ding{51}}} % Green checkmark
\newcommand{\ccmark}{\cmark\kern-0.4em\cmark} % Green double checkmark
\newcommand{\xmark}{\textcolor{red}{\ding{55}}} % Red cross

\newcolumntype{R}[2]{%
    >{\adjustbox{angle=#1,lap=\width-(#2)}\bgroup}%
    l%
    <{\egroup}%
}
\newcommand*\rot{\multicolumn{1}{R{35}{1em}}}

\begin{table}
\caption{Comparison to works for Body Reconstruction.}
\vspace{-4mm}
    \begin{center}
    \small
    \begin{tabular}{ccccccc|l}
    \rot{Pose-dependent Deformation} 
    & \rot{Novel Pose Animation}  
    & \rot{Fast Training} 
    & \rot{Rendering > 60FPS}  
    & \rot{Monocular Input} 
    & \rot{Super-resolution} 
    & \rot{Generalization} & \\
    \hline\hline
    \cmark
    & \cmark
    & \xmark
    & \xmark
    & \cmark
    & \xmark
    & \xmark
    & HuGS~\cite{moreau2023human}
    \\
    -
    & -
    & -
    & -
    & -
    & \cmark
    & \cmark
    & GPS-Gaussian~\cite{zheng2024gps-gaussian}
    \\
    \xmark
    & \cmark
    & \cmark
    & \cmark
    & \cmark
    & \xmark
    & \xmark
    & HUGS~\cite{kocabas2023hugs}
    \\
    \cmark
    & \cmark
    & \cmark
    & \cmark
    & \cmark
    & \xmark
    & \xmark
    & GaussianAvatar~\cite{hu2023gaussianavatar}
    \\
    \xmark
    & \xmark
    & \cmark
    & \cmark
    & \cmark
    & \xmark
    & \xmark
    & GauHuman~\cite{hu2023gauhuman}
    \\
    \cmark
    & \cmark
    & \cmark
    & \xmark
    & \cmark
    & \xmark
    & \xmark
    & 3DGS-Avatar~\cite{qian20233dgsavatar}
    \\
    \cmark
    & \cmark
    & -
    & \xmark
    & \cmark
    & \xmark
    & \xmark
    & ASH~\cite{pang2023ash}
    \\
    \cmark
    & \cmark
    & \xmark
    & \xmark
    & \xmark
    & \xmark
    & \xmark
    & Animatable Guassian~\cite{li2023animatable}
    \\
    \end{tabular}
    \end{center}
    \vspace{-7mm}
    \label{tab:comparison_concurrent}
\end{table}

Body reconstruction mainly focuses on reconstructing deformable human avatars from multi-view or monocular videos~\cite{sheng2024open}, as well as providing real-time rendering. We list comparisons of recent works in Tab.~\ref{tab:comparison_concurrent}.

% HuGS~\cite{moreau2023human} uses multi-view video as input to generate human models. This method introduces a coarse-to-fine process for human animation: Linear Blend Skinning (LBS) is initially employed to synthesize new human motions, while a shallow neural network captures and improves local deformations for non-linear elements, such as the movement of loose clothing during motion. Another notable work, HUGS~\cite{kocabas2023hugs}, enables training from monocular videos and employs triplane features for implicit representation. The Gaussian parameters—color and opacity, position and rotation, and LBS weights—are predicted by three separate MLP neural networks, which helps avoid overfitting and enhances generalization, especially for synthesizing novel motions.

Most works~\cite{moreau2023human, kocabas2023hugs, hu2023gauhuman, hu2023gaussianavatar, qian20233dgsavatar} prefer to use {\textit{\textbf{well-preconstructed human models like SMPL~\cite{Loper2015smpl} or SMPL-X~\cite{pavlakos2019smplx}}}} as strong prior knowledge. Nevertheless, SMPL is limited to introducing prior knowledge about the human body itself, thus posing challenges for the reconstruction and deformation of outward features such as garments and hair. \textit{\textbf{For the reconstruction of outward appearance,}} HUGS~\cite{kocabas2023hugs} utilizes SMPL and LBS only at the initial stage, allowing Gaussian primitives to deviate from the initial mesh to accurately represent garments and hair. Additionally, Animatable Gaussians~\cite{li2023animatable} uses a template that can fit outward appearance as guidance and leverages StyleGAN to learn the pose-dependent Gaussian maps, enhancing the capability for modeling detailed dynamic appearances. 

{\textit{\textbf{Some studies project the problem space from 3D to 2D}}, thereby reducing complexity and introducing well-established 2D networks for parameter learning~\cite{pang2023ash,li2023animatable}. Among them, ASH~\cite{pang2023ash} generates a motion-related template mesh via a deformation network and predicts Gaussian parameters through a 2D network using a motion-related texture map derived from this mesh. These methods are often limited by the difficulty of isolating clothing from the reconstructed 3DGS.}

{Although existing methods achieve high-fidelity reconstruction, their applications are limited by scene-specific training requirements, prompting research efforts toward generalizable human reconstruction~\cite{zheng2024gps-gaussian}.}

% Similarly, Animatable Gaussians~\cite{li2023animatable} projects a template mesh human model from canonical space onto two 2D planes, frontal and dorsal, learning Gaussian attributes in these spaces.

% Animatable Gaussians also utilizes Principal Component Analysis (PCA) to project new pose signals into the trained pose space, improving generalization for new poses.

% GPS-Gaussian~\cite{zheng2024gps-gaussian} addresses the generalizable human novel view synthesis by introducing Gaussian Parameter Maps that can be directly regressed without per-subject optimizations. This approach is complemented by a depth estimation module that lifts 2D parameter maps into 3D space.

% It employs a depth prediction module to obtain depth information from two selected input images, which helps elevate the Gaussian features learned and predicted in 2D space to 3D space.

% GPS-Gaussian~\cite{zheng2024gps-gaussian} uses Gaussian parameter maps

\subsubsection{Head Reconstruction}

{In the domain of human head reconstruction, the works~\cite{chen2024monogaussianavatar,qian2023gaussianavatars,zhao2024psavatar,dhamo2025headgas,teotia2024gaussianheads}, following the prevalent approach of utilizing SMPL as a strong prior, incorporates FLAME~\cite{siggraphAsia2017flame} meshes to provide geometric guidance or coarse reconstruction for 3DGS, achieving superior rendering quality.} However, Gaussian Head Avatar~\cite{xu2023gaussian} challenges the conventional use of FLAME meshes and LBS for facial deformation, arguing that these simplified linear operations are inadequate for capturing intricate facial expressions. As an alternative, it introduces an MLP-based approach that directly predicts Gaussian displacements during the transition from neutral to target expressions, enabling high-resolution head rendering at up to 2K resolution. {And then, the works~\cite{xu2024gaussian, xu20253d} replace FLAME-based initialization with geometric guidance derived from signed distance field and DMTet. Several works~\cite{ma20243d,xiang2024flashavatar} focus on rendering efficiency, achieving frame rates exceeding 300 FPS. For downstream tasks, various studies~\cite{li2025talkinggaussian,cho2024gaussiantalker,yu2024gaussiantalker} seek to integrate audio for controlling dynamic head reconstruction, thereby achieving audio-visual synchronization.}

%During initialization, a Gaussian is placed on each triangle of the FLAME mesh, and these Gaussians are optimized during the training process. To maintain control over adaptive density adjustments, GaussianAvatars employs a unique inheritance mechanism, ensuring that each Gaussian is associated with a specific triangle in the FLAME mesh, which deforms accordingly when the FLAME mesh is animated.%

\subsubsection{Others}

3DGS has introduced innovative solutions in other human-related areas~\cite{luo2024gaussianhair,bolanos2024charshadow}. GaussianHair~\cite{luo2024gaussianhair} focuses on the reconstruction of human hair, using linked cylindrical Gaussian modeling for the strands. {Additionally, some research~\cite{abdal2023gaussian,liu2023humangaussian, kirschstein2024gghead} have explored the integration of 3DGS with generative models. }

\subsection{Artificial Intelligence-Generated Content (AIGC)}\label{AIGC}

Artificial Intelligence Generated Content (AIGC) leverages artificial intelligence technologies to autonomously produce content. In this sector, we systematically classify contemporary algorithms based on the types of prompts and the objects they generate. The categories include Image-to-3D Object Generation, Text-to-3D Object Generation, Multi-Object and Scene Generation, and 4D Generation.

\begin{figure*}[!tp]
\centering
\subfigure[Text to 3D Objects~\cite{tang2023dreamgaussian}.]{
           \includegraphics[height=0.20\textheight]{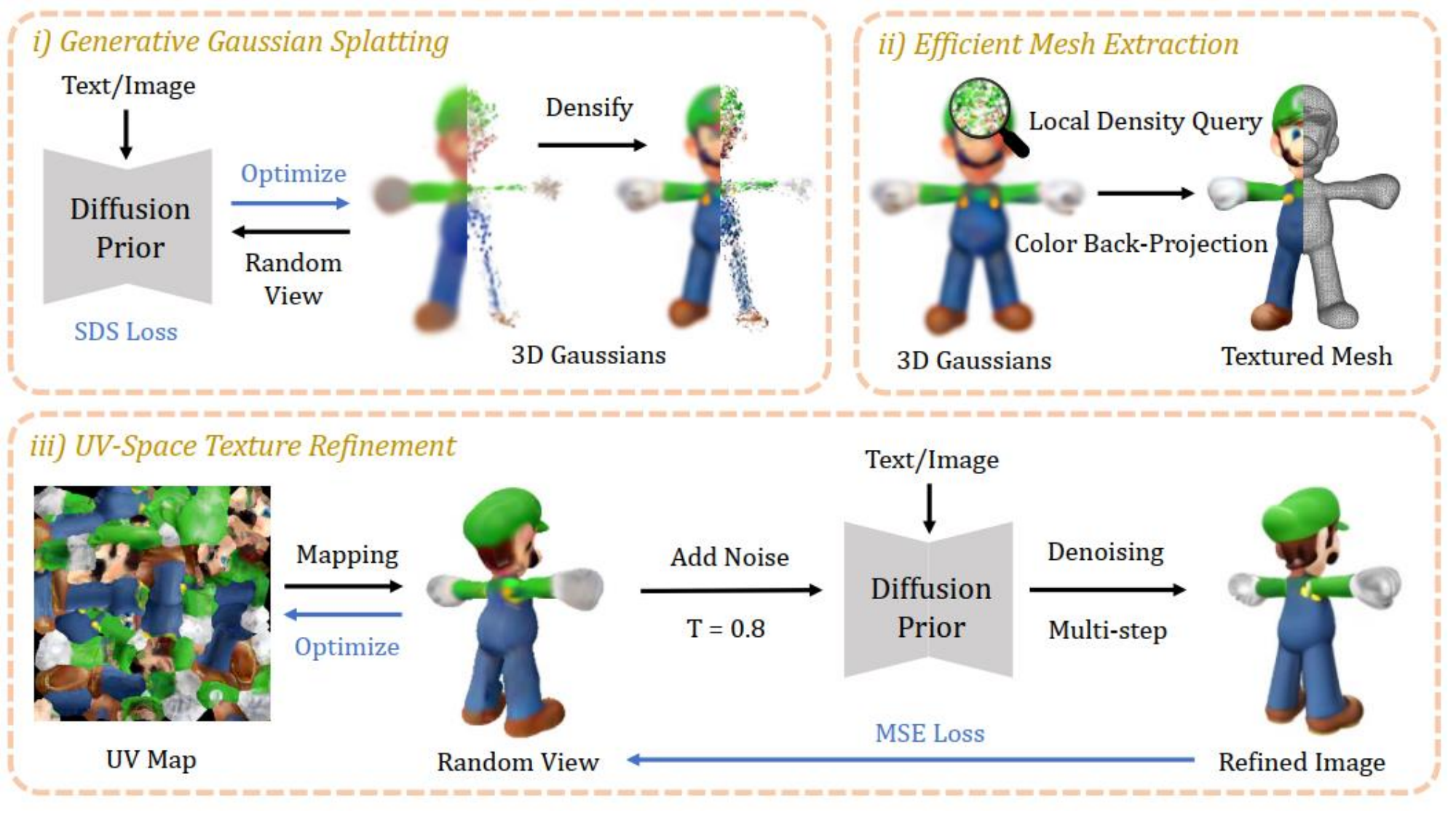}}
\subfigure[4D Generation~\cite{ren2023dreamgaussian4d}.]{
           \includegraphics[height=0.18\textheight]{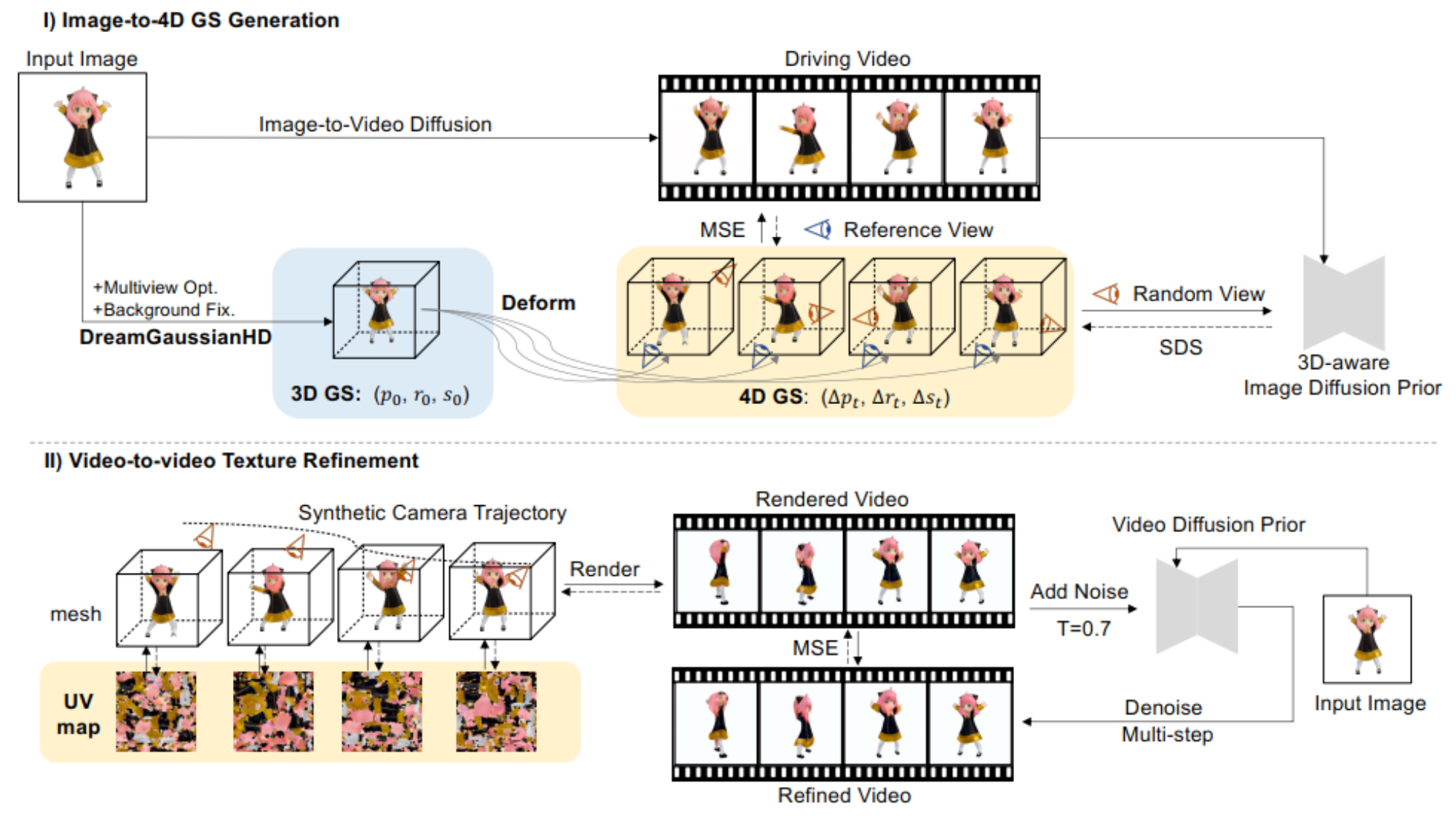}}
\subfigure[Image to 3D Object~\cite{zhang2023repaint123}.]{
           \includegraphics[height=0.12\textheight]{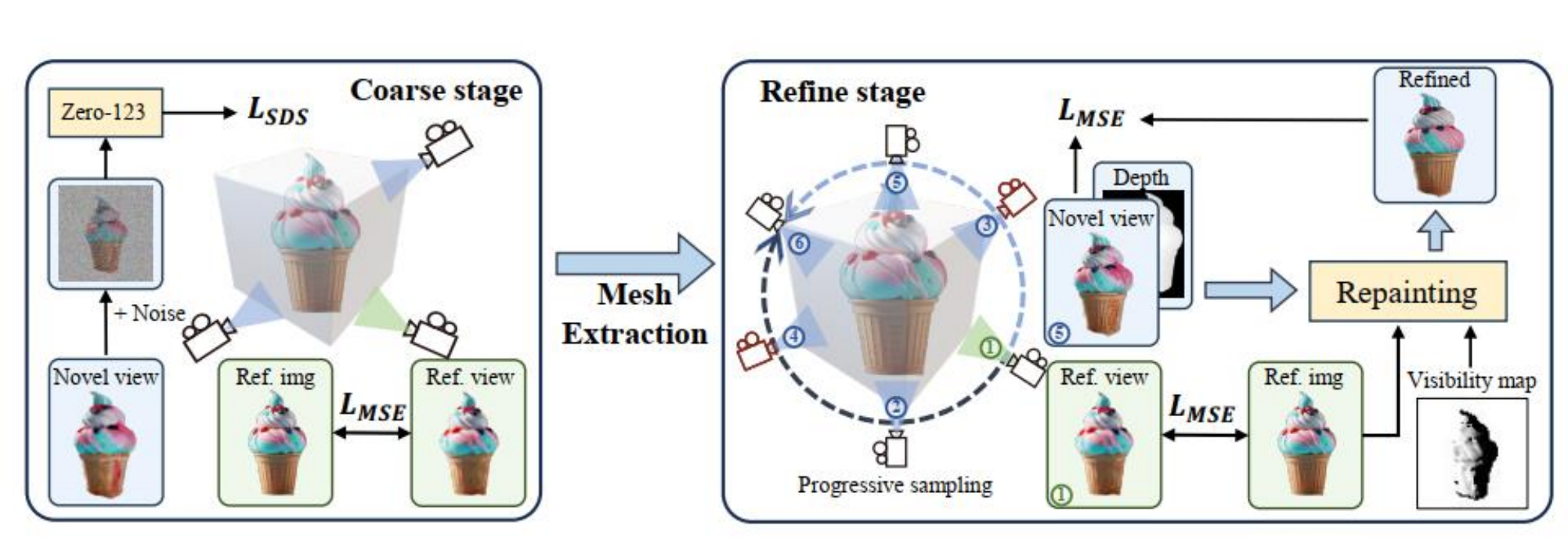}}
\subfigure[Multi-Object Generation~\cite{zhou2024gala3d}.]{
           \includegraphics[height=0.12\textheight]{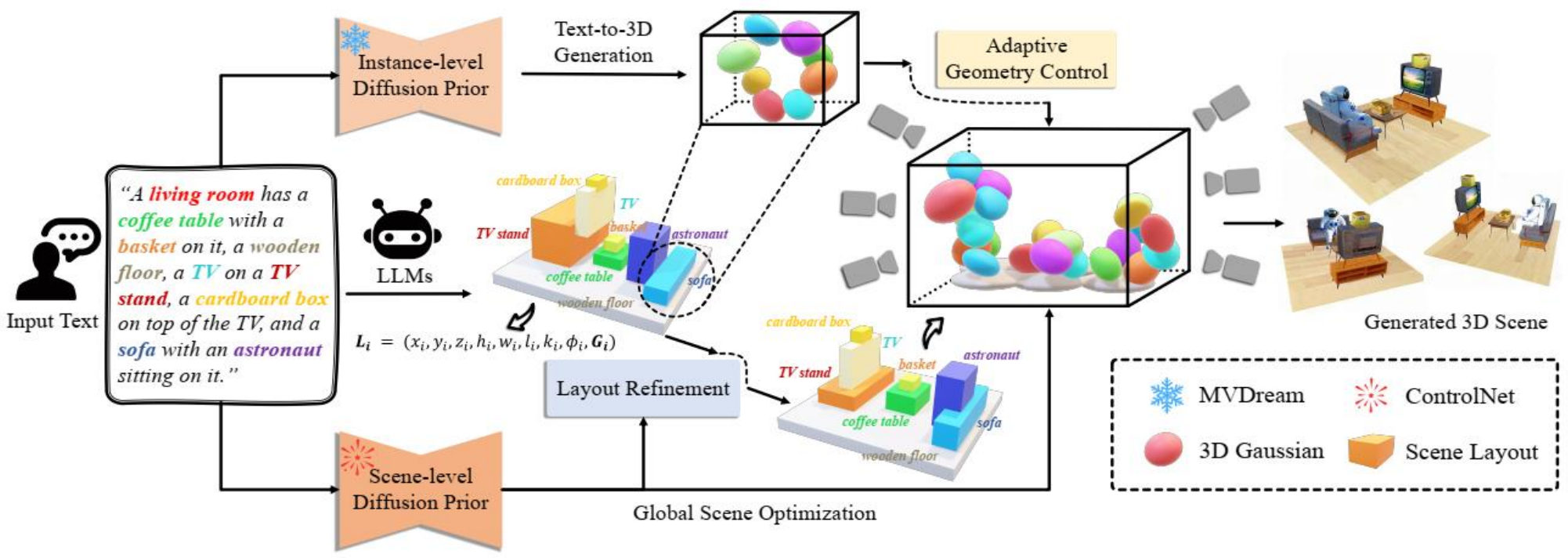}}
% \vspace{-0.3cm}
\caption{Four typical tasks of 3DGC in AIGC Applications.}
\label{fig2}
\vspace{-0.5cm}
\end{figure*}

\subsubsection{Text to 3D Objects}\label{Text to 3D Objects}
% At present, a substantial body of research is dedicated to extending Score Distillation Sampling (SDS)~\cite{poole2022dreamfusion}, which plays a crucial role in this context. To further elucidate SDS, we denote the 3D representations as $\theta$ and the differentiable rendering process as $g(\cdot)$, thereby representing the rendered image as $g(\theta)$. DREAMFUSION\cite{poole2022dreamfusion} ensures that the rendered images from each camera viewpoint adhere to the credible samples derived from the pre-trained diffusion model $\varphi$. In practice, they leverage the score estimation function $\epsilon_{\phi}(\bm{x_{t}},t, y)$ of the existing diffusion model, where $\epsilon_{\phi}$ predicts the sampled noise based on the noisy image $\bm{x_{t}}$ and the textual condition $y$. Therefore, the gradient of score distillation loss for $\theta$,
% % To encourage sampled representations to converge towards denser regions of the true data distribution,
% \begin{equation}
%      \nabla_{\theta } =\mathbb{E} \left [ w_{t}\left ( \epsilon_{\phi}\left ( \bm{x_{t}},t,y \right )-\epsilon   \right ) \frac{\partial \bm{x}}{\partial \theta}  \right ]. \label{eq3}
% \end{equation}
{Extensive existing research endeavors to utilize the superior generative capabilities of 2D generative models to achieve coherent 3D content creation. Benefiting from reduced dependency on extensive 3D training data, Score Distillation Sampling-based paradigms garnered significant attention in early research~\cite{poole2022dreamfusion}. Some works~\cite{tang2023dreamgaussian, chen2023text, yi2023gaussiandreamer,shen2025vista3d} focus on improving the framework to \textbf{\textit{apply score distillation loss to 3DGS.}}} Building upon Score Distillation Sampling (SDS), DreamGaussian~\cite{tang2023dreamgaussian} ensures the geometric consistency of the generated models by extracting explicit Mesh representations from the 3DGS and refines texture in the UV space to enhance the quality of the renderings. {However, the mode-seeking paradigm of score distillation frequently leads to \textbf{\textit{ oversaturation, excessive smoothing, and lack-detail}} in the generated outcomes~\cite{li2023gaussiandiffusion,liang2023luciddreamer,di2024hyper,yang2023learn,zhuo2025vividdreamer}.} 
Among them, LucidDreamer~\cite{liang2023luciddreamer} addresses the challenges of over-smoothing and insufficient sampling steps inherent in traditional SDS. By introducing deterministic diffusion trajectories (DDIM~\cite{song2020denoising}) and interval-based score matching mechanisms Eq.~\ref{eq9}, it achieves superior generation quality and efficiency.

{To further mitigate the inherent limitations of SDS, several works~\cite{melas20243d, voleti2025sv3d,gao2024cat3d,han2025vfusion3d,yang2024hi3d} seek to leverage \textbf{\textit{video or multi-view generative models}} to obtain more data for reconstruction. Although these approaches introduce direct prior guidance for 3D generation, the inherent lack of guaranteed 3D consistency in both multi-view and video generation still leads to instability in reconstruction. To enhance the efficiency of 3D asset generation, works~\cite{jiang2024brightdreamer, tang2024lgm,zhang2024gaussiancube} aim to generate using only \textit{\textbf{feed-forward networks}} without the need for scene-specific training. BrightDreamer~\cite{jiang2024brightdreamer} predicts positional offsets following fixed initialization and employs a text-guided triplane generator to process extracted textual features for predicting additional 3DGS attributes, achieving text-to-3D model conversion in 77ms, albeit at the cost of some reconstruction quality. For enhanced geometric detail control, SketchDream~\cite{liu2024sketchdream} proposes a framework for sketch-based text-to-3D generation and editing, which integrates hand-drawn sketches and text prompts to achieve fine-grained geometry and appearance control, enabling high-quality 3D content creation and local editing through a two-stage coarse-to-fine approach.} 

{More directly, several works~\cite{he2024gvgen,lu2024large,roessle2024l3dg,yu2024get3dgs} aim to incorporate 3DGS representations into 3D generative models. Among these, L3DG~\cite{roessle2024l3dg} proposes a latent diffusion framework for compressed 3D Gaussian representation, achieving superior visual quality and real-time rendering. However, these approaches are inherently limited by the availability of 3D data, which impacts their reconstruction capability for complex targets.}

% Among them, IM-3D~\cite{melas20243d} identifies optimization difficulties associated with score distillation loss. Thus, it attempts to fine-tune existing image-to-video generation models to enable the generation of multi-view spatially consistent images (videos). These generated multi-view images are then used as supervision for 3DGS generation. 

% Similarly, LGM~\cite{tang2024lgm} proposes a novel paradigm to generate 3DGS from text or a single image. It generates multi-view images of the target using existing networks and reconstructs 3D models under different inputs using an asymmetric U-Net based architecture with cross-view self-attentions.

% GVGEN~\cite{he2024gvgen} focuses on a feed-forward setup without triplane, proposing GaussianVolumes as a structured lightweight representation for generalizable generation. Based on this representation, GVGEN generates a Gaussian Distance Field with a trained diffusion model and utilizes it to guide the prediction of corresponding attributes.

 {Some works~\cite{abdal2023gaussian, liu2023humangaussian,yuan2024gavatar} also attempt to apply this generative paradigm to areas such as \textbf{\textit{digital human generation.}}} HumanGaussian~\cite{liu2023humangaussian} combines RGB and depth rendering to improve the SDS, thereby jointly supervising the optimization of the structural perception of human appearance and geometry. 

% Additionally, it introduces annealed negative prompt guidance and a scaling-based pruning strategy to address oversaturation and floating artifacts. 

% In addition to the numerous works relying heavily on diffusion models, the work by Abdal et al.~\cite{abdal2023gaussian} proposes a novel paradigm that combines 3DGS with Shell Maps~\cite{porumbescu2005shell} and a 3D Generative Adversarial Networks (GANs) framework. By leveraging Gaussian Shell Maps, this approach rapidly represents human bodies and their corresponding deformations.

\subsubsection{Image to 3D Object}\label{Image to 3D Object}

{Similar to works on NeRF, recent studies~\cite{feng2024fdgaussian,zhang2023repaint123,chen2025cascade} have also focused on generating entire 3DGS from a single image. These approaches share fundamental similarities with text-to-3D object methods. As an example, following a process similar to DreamGaussian~\cite{tang2023dreamgaussian}, Repaint123~\cite{zhang2023repaint123} employs Zero-123~\cite{liu2023zero} and SDS for coarse 3DGS, followed by a fine stage where mesh representation is extracted and refined using depth-guided and visibility-aware repainting on novel views for consistent 3DGS fine-tuning. }

% Repaint123~\cite{zhang2023repaint123} attempts to utilize a progressive image repainting strategy, generating multi-view images with 3D consistency. Following a process similar to DreamGaussian~\cite{tang2023dreamgaussian}, Repaint123 divides this procedure into coarse and fine optimization stages. In the coarse stage, it uses the pre-trained Zero-123~\cite{liu2023zero} as supervision and employs score distillation loss to optimize a coarse 3DGS. During the fine stage, Repaint123 extracts the mesh representation from the first stage and proposes combining depth images and reference images based on DDIM inversion to guide the denoising process on novel view images, ensuring consistency across views. For overlapping and occluded regions between views, Repaint123 employs visibility-aware adaptive repainting methods to enhance repainting quality in these areas. The repainted images are then used for fine-tuning the 3DGS, resulting in more refined outputs. 

% FDGaussian~\cite{feng2024fdgaussian} proposes a more straightforward approach by dividing the entire generation process into multi-view image generation and 3DGS reconstruction. During the generation phase, FDGaussian extracts 3D features from the image by decoupling orthogonal planes, optimizing the multi-view generation model based on Zero-1-to-3~\cite{liu2023zero}. In the reconstruction phase, it enhances the efficiency and performance of 3DGS through control optimization based on inter-Gaussian distances and a fusion strategy based on epipolar attentions.

\subsubsection{Multi-Object and Scene Generation} \label{Multi-Object and Scene Generation}

In addition to single-object generation, multi-object and scene generation is more crucial in most application scenarios.

\textbf{Multi-Object Generation:} Several studies~\cite{vilesov2023cg3d, zhou2024gala3d,li2025dreamscene,chen2025comboverse} have explored the generation of multiple composite objects, which not only \textit{\textbf{concentrate on the individual objects}} but also aim to \textit{\textbf{investigate the interactions between multiple objects.}} {For predicting the interactions between multiple objects, CG3D~\cite{vilesov2023cg3d} leverages SDS and probabilistic graph models extracted from text to predict the relative relationships between objects and incorporates priors such as gravity and contact relationships between objects, CG3D achieves models with realistic physical interactions.} 

% To simplify such a problem, GALA3D~\cite{zhou2024gala3d} uses layouts generated by large language models (LLMs) to guide multiple object reconstructions. By exploring the distribution of positions and optimizing the shapes of Gaussian primitives according to the layout, GALA3D generates scenes that conform to the specified layouts. Additionally, by supervising the generation of individual objects and entire scenes through SDS and introducing a Layout Refinement module, GALA3D achieves more realistic and text-consistent generation results.

\textbf{Scene Generation:} {Unlike object-centric generation, scene generation typically requires the \textbf{\textit{incorporation of additional information}}, such as pre-trained depth estimation models~\cite{chung2023luciddreamer,ouyang2023text2immersion} or Large Language Models~\cite{zhou2025dreamscene360,li2025dreamscene}, where the former provides spatial understanding for projecting images into 3D space, while the latter enhances the quality of text prompts. LucidDreamer$^2$\cite{chung2023luciddreamer} employs a two-stage approach: first initializing point clouds using text-to-image and depth estimation models with inpainting\cite{lugmayr2022repaint} for consistency, then converting to 3DGS with extended image supervision.}

% To achieve this, LucidDreamer$^2$~\cite{chung2023luciddreamer} designs a two-stage generative paradigm. In the first stage, LucidDreamer$^2$ leverages pre-trained text-to-image and monocular depth estimation models to initialize the point clouds, and introduces inpainting models~\cite{lugmayr2022repaint} to complete a multi-view consistent scene point cloud. In the second stage, the generated point cloud is used to initialize 3DGS, with extended supervision images to ensure a smoother training process. 

% Based on a similar paradigm, Text2Immersion~\cite{ouyang2023text2immersion} introduces a poses-progressive generation strategy to achieve a more stable training process, and incorporates enlarged viewpoints and pre-trained super-resolution models to optimize the generated scenes. 

\subsubsection{4D Generation}~\label{4D Generation}
{Analogous to static scene generation using text-to-image SDS, it is natural to consider that \textbf{\textit{text-to-video SDS could potentially generate dynamic scenes~\cite{ling2023align, ren2023dreamgaussian4d, gao2024gaussianflow, bahmani20244d,bahmani2025tc4d,zheng2024unified}.}} These works primarily focus on designing video-based SDS losses or exploring hybrid supervision with T2I (Text to Image) and T2V (Text to Video) models. As an example, Align Your Gaussians \cite{ling2023align} adopts a two-stage approach: first reconstructing static 3DGS using MVDream\cite{shi2023mvdream} and text-to-image supervision, then extending to 4DGS with text-to-video guidance and simplified score distillation loss. Although these methods are effective, the inherent limitations of the aforementioned SDS-based paradigm persist.} 

{To mitigate this issue, subsequent works focus on \textbf{\textit{generating pseudo-labeled images}} from additional views to facilitate dynamic 3DGS reconstruction~\cite{yin20234dgen,pan2024fast,sun2024eg4d,zeng2025stag4d}. Among them, 4DGen~\cite{yin20234dgen} generates multi-view pseudo-labels per frame, while employing Hexplane's~\cite{cao2023hexplane} multi-scale features to maintain temporal consistency in 4DGS generation.}

{Furthermore, some studies focus on \textit{\textbf{animating static canonical 3DGS}}~\cite{zhang2024bags, wu2025sc4d} to achieve better control. Among them, }BAGS~\cite{zhang2024bags} introduces neural bones and skinning weights to describe the spatial deformation based on canonical space. Using diffusion model priors and rigid body constraints, BAGS can be manually manipulated to achieve novel pose rendering.

\subsection{Autonomous Driving}\label{Autonomous Driving}
% In the field of autonomous driving, 3DGS is primarily applied to the dynamic reconstruction of large-scale driving scenes and combined SLAM applications.

\subsubsection{Autonomous Driving Scene Reconstruction} \label{Autonomous Driving Scene Reconstruction}

Reconstruction driving scenes is a challenging task, involving multiple technical domains such as large-scale scene reconstruction, dynamic object reconstruction, static object reconstruction, and Gaussian mixture reconstruction. {Existing works~\cite{zhou2023drivinggaussian, yan2024street, zhou2024hugs} partitions the entire process into \textit{\textbf{static background and dynamic target reconstruction.}} DrivingGaussian~\cite{zhou2023drivinggaussian} reconstructs large-scale driving scenes by combining depth-binned static 3DGS for background and dynamic Gaussian graphs for multiple targets, utilizing multi-sensor data. Then StreetGaussians~\cite{yan2024street} extends this approach by incorporating semantic attributes and employing Fourier transforms for efficient SH temporal modeling in dynamic 3DGS. Subsequent works~\cite{zhao2024tclc}, aiming to further improve reconstruction efficiency, attempt to directly reconstruct entire scenes by Tightly Coupled LiDAR-Camera Gaussian Splatting.}

% Among them, DrivingGaussian~\cite{zhou2023drivinggaussian} aims to utilize multi-sensor data for reconstructing large-scale dynamic scenes in autonomous driving. In the static background, DrivingGaussian introduces incremental static 3D Gaussians under different depth bins to mitigate the scale confusion caused by distant street scenes. For dynamic objects, DrivingGaussian introduces dynamic Gaussian graphs to construct relationships among multiple targets (whose attributes include position, local-to-world coordinate transformation matrices, and orientation, etc.), jointly reconstructing the entire autonomous driving scene with the static background.

% StreetGaussians~\cite{yan2024street} adopts a similar approach, with the key difference being the introduction of semantic attributes during the reconstruction of background and foreground. Also, StreetGaussians uses Fourier transforms to efficiently represent the SH's temporal changes of dynamic 3DGS. Building on previous works, HUGS~\cite{zhou2024hugs} incorporates the Unicycle Model and the modeling of forward and angular velocities to assist in the dynamic reconstruction under physical constraints. Similar to previous dynamic 3DGS efforts~\cite{katsumata2023efficient, guo2024motion}, HUGS also employs optical flow supervision, combined with rendered RGB loss, semantic loss, and Unicycle Model losses, thus improving dynamic reconstruction accuracy.

Moreover, 3DGS have been applied to \textit{\textbf{multimodal spatiotemporal calibration tasks}}~\cite{herau20243dgs}. By leveraging the LiDAR point cloud as a reference for the Gaussians' positions, 3DGS-Calib~\cite{herau20243dgs} constructs a continuous scene representation and enforces both geometrical and photometric consistency across all sensors, achieving accurate and robust calibration {with improved efficiency compared to NeRF-based works.}

\subsubsection{Simultaneous Localization and Mapping (SLAM)}\label{SLAM}

% The technical approaches to SLAM can be broadly categorized into traditional methods, techniques involving NeRF, and approaches related to 3DGS. Among these, 3DGS methods stand out for their ability to offer continuous surface modeling, reduced memory requirements, improved noise and outlier handling, enhanced hole filling and scene repair, and flexible resolution in 3D mesh reconstruction~\cite{tosi2024nerfs}. 
SLAM is a key problem in robotics and computer vision, where a device builds a map of an unknown environment while locating itself within it.
{Some studies \cite{yan2023gs,huang2023photo,keetha2023splatam,yugay2023gaussian,hu2025cg,ha2024rgbd} have retained the traditional inputs and approached this from two perspectives: \textit{\textbf{online tracking and incremental mapping.}} In early work, GS-SLAM~\cite{yan2023gs} utilizes 3DGS for SLAM with adaptive primitive expansion and employs a coarse-to-fine optimization strategy: first optimizing camera poses using sparse pixels, then refining them through selective re-rendering of reliable Gaussians. In parallel, Photo-SLAM~\cite{huang2023photo} combines ORB features~\cite{rublee2011orb} and Gaussian attributes in a Hyper Primitives Map, utilizing LM optimization~\cite{more2006levenberg} and loop closure~\cite{rublee2011orb} for photorealistic SLAM reconstruction. While these approaches achieve higher efficiency than NeRF-based methods, further optimization of computational performance is essential for real-world deployment. Therefore, CG-SLAM~\cite{hu2025cg} leverages an uncertainty-aware 3D Gaussian field and a GPU-accelerated framework, achieving superior efficiency. Then, RGBD GS-ICP SLAM~\cite{ha2024rgbd} enhances efficiency by integrating G-ICP~\cite{segal2009generalized} with shared covariances and scale alignment techniques for faster convergence. However, these methods remain susceptible to sensor noise in practical applications. Some quantitative reconstruction
results are reported in Table~\ref{slam comparison}.}

{Incorporating scene understanding capabilities is equally crucial in SLAM tasks~\cite{li2024sgs,zhu2024semgauss,ji2024neds}, thus prompting several works to \textit{\textbf{integrate semantic information}}. Among them,} SGS-SLAM~\cite{li2024sgs} employs multi-channel geometric, appearance, and semantic features for rendering and optimization and proposes a keyframe selection strategy based on geometric and semantic constraints to enhance performance and efficiency.

% Based on it, SEMGAUSS-SLAM~\cite{zhu2024semgauss} designs a feature-level supervision for robustness and introduces a feature-based Bundle Adjustment to mitigate cumulative drift during tracking. Subsequent works, NEDS-SLAM~\cite{ji2024neds}, also adopts this concept, introducing semantic feature-assisted SLAM optimization and incorporating DepthAnything~\cite{yang2024depth} to learn semantically rich features with 3D spatial awareness. Additionally, NEDS-SLAM proposes a novel pruning method based on virtual multi-view consistency checks to identify and eliminate outliers.

{Additionally, there are several works focusing on related issues such as \textit{\textbf{ localization~\cite{jiang20243dgs} and navigation~\cite{lei2024gaussnav}.}}} Specifically, 3DGS-ReLoc~\cite{jiang20243dgs} leverages LiDAR initialization and 2D voxelized submaps with KD-tree for efficient memory usage, while achieving precise localization through feature-based PnP optimization. In the context of indoor navigation, GaussNav~\cite{lei2024gaussnav} focuses on the instance image navigation task. Based on reconstructed 3DGS maps, GaussNav proposes an image target navigation algorithm, achieving impressive performance through classification, matching, and path planning.

\begin{table}[t]
\caption{Comparison of 3DGS-SLAM on Replica~\cite{straub2019replica}.}
\centering
\begin{tabular}{p{3cm}p{0.5cm}p{0.5cm}p{0.6cm}>{\centering\arraybackslash}p{2.5cm}}
\toprule
\textbf{Method} & \textbf{PSNR↑} & \textbf{SSIM↑} & \textbf{LPIPS↓} & \textbf{Tracking RMSE↓} \\ 
\midrule
GS-SLAM~\cite{yan2024gs} & 34.27 & 0.98 & 0.08 & 0.50\\ 
Photo-SLAM~\cite{huang2023photo} & 34.96 &  0.94 &  0.06 & 0.60 \\ 
SplaTAM~\cite{keetha2023splatam} & 34.11 & 0.97 &  0.10 & 0.36 \\
GS-ICP SLAM~\cite{ha2024rgbd} & 38.83 &  0.98 & \textbf{0.04} & \textbf{0.16} \\
% Gaussian-SLAM~\cite{yugay2023gaussian} &  38.90 & \textbf{0.99} & 0.07 & 0.48 \\
MotionGS~\cite{guo2024motiongs} & \textbf{39.60} &  0.98 & \textbf{0.04} & 0.49\\
LoopSplat~\cite{zhu2024loopsplat} & 36.63 & \textbf{0.99} & 0.11 & 0.26\\
\bottomrule
\end{tabular}
\vspace{-5mm}
\label{slam comparison}
\end{table}

\section{Extensions of 3D Gaussian Splatting}\label{Extensions}

\subsection{Dynamic 3D Gaussian Splatting}\label{4DGS}
The study of dynamic 3DGS has recently garnered significant attention. Here, we categorize them into Multi-view Videos and Monocular Video based on different inputs.
% The study of dynamic 3DGS has recently garnered significant attention from researchers. The reconstruction of dynamic scenes transcends the limitations of static scene reconstruction and can be effectively applied to fields such as human motion capture and autonomous driving simulation. Unlike static 3DGS, dynamic 3DGS must account for consistency not only in the spatial dimension but also in the temporal dimension, ensuring continuity and smoothness over time. Here, we categorize them into Multi-view Videos and Monocular Video based on different reconstruction inputs.
\subsubsection{Multi-view Videos}\label{Multi-view Videos}
% This has led to numerous discussions on four-dimensional space (XYZ-t). Existing dynamic neural radiance fields rely on the inherent smoothness of implicit representations, while Gaussian elements struggle to represent smooth dynamic changes in the temporal dimension.
{Some works~\cite{luiten2023dynamic,sun20243dgstream} attempted to \textit{\textbf{directly construct dynamic 3DGS frame by frame.}}} {An early work~\cite{luiten2023dynamic} extends 3DGS to dynamic scenes by enabling temporal Gaussian motion while preserving static attributes. It employs online temporal reconstruction with previous-frame initialization and incorporates physical priors, including local rigidity, local rotational-similarity, and long-term local-isometry, for motion regularization, as shown in Eq.\ref{eq5}-\ref{eq7}. Similarly, 3DGStream~\cite{sun20243dgstream} employs a two-stage training with Neural Transformation Cache and I-NGP~\cite{muller2022instant} for dynamic reconstruction, followed by gradient-based adaptive Gaussian densification. Despite promising results, these methods are limited to reconstructing only scene elements visible in the initial frame.} To address this limitation, {other works~\cite{shaw2023swags,xiao2024bridging} aim to achieve such performance by \textit{\textbf{predicting deformations.}} SWAGS~\cite{shaw2023swags} proposes a window-based 4DGS with flow-guided adaptive window division and dynamic MLP optimization, employing inter-window consistency loss for seamless temporal reconstruction. }
% By employing a new Gaussian-Mesh hybrid representation and an optimized learning strategy, the method achieves better reconstruction results without sacrificing rendering efficiency.
\subsubsection{Monocular Video or Multi-view Images}\label{Monocular Video}
{Some works~\cite{yang2023deformable,liang2023gaufre,wu20234d,duisterhof2023md,guo2024motion, li2024st,lu20243d} tend to \textit{\textbf{divide into two stages: canonical reconstruction and deformation prediction.}} The studies~\cite{yang2023deformable, liang2023gaufre} reconstruct static 3DGS and predicts temporal deformation in terms of positions, rotations and scales through position-time encoding. Similarly, 4D-GS~\cite{wu20234d} introduces the multi-scale HexPlane~\cite{cao2023hexplane} as the foundational representation to encode temporal and spatial information. To further decouple motion and shape parameters in 4D-GS, ST-4DGS~\cite{li2024st} introduces a spatial-temporally consistent 4DGS framework that incorporates motion-aware shape regularization and spatial-temporal density control to learn compact 4D representations. Although these methods achieve stable performance, they struggle to handle abrupt motions and sudden object appearances.}

% Building on this foundation, GauFRe~\cite{liang2023gaufre} proposes a paradigm that decouples dynamic and static scene modeling, where the dynamic part utilizes a method similar to that of~\cite{yang2023deformable}. 4D-GS~\cite{wu20234d} introduces the multi-scale HexPlane~\cite{cao2023hexplane} as the foundational representation to encode temporal and spatial information. To optimize the training process, 4D-GS employs multi-head decoders to predict different attributes of Gaussian primitives separately. MD-Splatting~\cite{duisterhof2023md} also builds on this by incorporating the local rigidity loss and isometric loss as proposed in~\cite{luiten2023dynamic}, and designs a regularization term based on the law of conservation of momentum, making the dynamic motion trajectories smoother. Additionally, the authors include shadow prediction in the decoding part, further enhancing the realism of the reconstruction.

% The work by Guo et al.~\cite{guo2024motion} constructs a flow-augmentation method by analyzing the correspondence between 3D Gaussian motion and pixel-level flow, introducing additional optical flow-based supervision based on uncertainty and dynamic awareness. Additionally, it proposes a Motion Injector and Dynamic Map Refinement strategy based on a velocity field to mitigate the challenges associated with predicting deformations.

{Instead of discrete offsets, \textit{\textbf{exploring temporally continuous motion}} can promote smoothness in the time dimension~\cite{katsumata2023efficient,kratimenos2023dynmf,lin2023gaussian,huang2023sc}. Gaussian-Flow~\cite{lin2023gaussian} aims to develop a representation capable of fitting variable motion by analyzing the advantages and disadvantages of polynomial~\cite{kratimenos2023dynmf,li2023spacetime} and Fourier series fitting~\cite{katsumata2023efficient}. It then proposes a dual-domain deformation model with adaptive time-step scaling and temporal-rigid constraints for stable and continuous motion prediction.} 
\begin{table}[t]
\caption{Comparison of Dynamic 3DGS on D-NeRF~\cite{pumarola2021d}.}
\centering
\begin{tabular}{p{3cm}p{0.5cm}p{0.5cm}p{0.6cm}>{\centering\arraybackslash}p{0.8cm}p{0.6cm}}
\toprule
\textbf{Method} & \textbf{PSNR↑} & \textbf{SSIM↑} & \textbf{LPIPS↓} & \textbf{Train↓} & \textbf{FPS↑} \\ 
\midrule
K-Planes [NeRF-based] & 31.07 & 0.97 & 0.02 & 54 min & 1.20 \\ 
\midrule
Deformable3DGS~\cite{yang2023deformable} & 39.31 & 0.99 & 0.01 & 26 min & 85.45 \\ 
CoGS~\cite{yu2023cogs} & 37.90 & 0.983 & 0.027 & - & - \\
SC-GS~\cite{huang2023sc} & \textbf{43.31} & \textbf{0.997} & \textbf{0.0063} & - & - \\
GauFRe~\cite{liang2023gaufre} &  34.5 & 0.98 & 0.02 & 13mins  & 112 \\
Deformable4DGS~\cite{wu20234d} & 32.99 & 0.97 & 0.05 & 13 min & 104.00 \\ 
RealTime4DGS~\cite{yang2023real} & 32.71 & 0.97 & 0.03 & 10 min & 289.07 \\
4DRotorGS~\cite{duan20244d} & 34.26 & 0.97 & 0.03 & \textbf{5 min}  & \textbf{1257.63} \\ 
\bottomrule
\end{tabular}
\vspace{-4mm}
\label{4dgs comparison}
\end{table}

{Recent works aim to \textit{\textbf{extend 3DGS to 4D space}} for dynamic 3D scenes representation. Among them, the work~\cite{yang2023real} achieves end-to-end 4D training by jointly modeling spatial-temporal variables with 4D Gaussian primitives, incorporating 4D rotation, scaling, and temporal-aware spherical harmonics for color variation. Similarly, the work~\cite{duan20244d} introduces a rotor-based 4DGS representation with eight-component rotation decomposition, enabling temporal slicing for dynamic objects and enforcing 4D consistency through a dedicated loss. Although these approaches demonstrate robustness in complex scene reconstruction, the compressibility of their representations remains a notable consideration. Quantitative reconstruction
results are reported in Table~\ref{4dgs comparison}.}

{Dynamic 3DGS reconstruction demonstrate \textbf{\textit{broad applicability}} across diverse domains, extending beyond previously mentioned human reconstruction~\ref{Human Reconstruction}, AIGC~\ref{AIGC}, and autonomous driving~\ref{Autonomous Driving} to include robotic manipulation~\cite{lu2025manigaussian}, physical simulation~\cite{xie2023physgaussian}, and various downstream tasks.}

% Similarly, the work~\cite{duan20244d} proposes a rotor-based 4D Gaussian Splatting (4DGS) representation, wherein the rotation attributes of 4DGS are expressed by decomposing 4D rotors into eight components. These components, along with corresponding parameters, are used to describe the rotation in space-time. When representing dynamic 3DGS, it slices 4DGS at different timestamps, effectively addressing the sudden appearance or disappearance of objects in highly dynamic scenes. Furthermore, the approach enforces consistency in 4D space through the introduction of a 4D Consistency Loss.

\subsection{Surface Representation}\label{Surface Representation}

Although 3DGS enables highly realistic rendering, extracting surface representations remains challenging. {In this line of works, \textit{\textbf{Signed Distance Functions}} are an indispensable topic~\cite{guedon2023sugar,chen2023neusg,lyu20243dgsr,yu2024gsdf}.In early work, SuGaR~\cite{guedon2023sugar} proposes an idealized SDF to constrain Gaussian surface alignment, enabling efficient mesh extraction through Poisson reconstruction and optional mesh-guided Gaussian refinement for high-quality results. Similarly, 3DGSR~\cite{lyu20243dgsr} integrates neural implicit SDF with 3DGS through a differentiable SDF-to-opacity transformation, maintaining consistency between volumetric and 3DGS-derived depth properties. Another line of research~\cite{chen2023neusg, yu2024gsdf} focuses on jointly optimizing NeuS~\cite{wang2021neus} and 3DGS for surfaces. However, these methods exhibit limitations in handling unbounded scenes and computational overhead.}

{Other studies~\cite{dai2024high,yu2024gaussian,huang20242d,reiser2024binary} aim to address this issue by \textit{\textbf{improving 3DGS representation.}} The work~\cite{dai2024high} proposes Gaussian Surfels with depth-normal consistency loss and volumetric cutting for improved surface reconstruction, followed by screened Poisson mesh generation. Similarly, 2D Gaussian Splatting~\cite{huang20242d} (2DGS) replaces 3DGS with planar disks to represent surfaces, which are defined within the local tangent plane. Then, Gaussian Opacity Fields (GOF)~\cite{yu2024gaussian} are developed based on 3DGS, wherein 3DGS is normalized along the ray to form a 1DGS for volume rendering. Although these methods achieve precise geometry reconstruction, they inevitably compromise rendering fidelity and face challenges in handling semi-transparent surfaces.}

% The work~\cite{dai2024high} introduces a novel representation termed Gaussian Surfels, which exhibits an enhanced capability for surface reconstruction. Based on this representation, it proposes a depth-normal consistency loss to address the gradient vanishing problem and a volumetric cutting strategy to prune unnecessary voxels in regions with depth errors and discontinuities. Finally, this work applies screened Poisson reconstruction to generate the surface mesh.

%  GOF also incorporates depth distortion and normal consistency losses, facilitating the extraction of surface meshes from tetrahedral grids.

% By compressing the $z$-scale to zero and designating the $z$ direction as the normal direction for supervision, the Gaussian Surfels are better aligned with surface properties.

\subsection{Editable 3D Gaussian Splatting}\label{Editable}

% Editing 3D scenes or objects has long been crucial for achieving controllable, interactive generation. Traditional 3D representations, such as meshes~\cite{yu2004mesh} and point clouds~\cite{scheiblauer2011out}, are favored for their interactive editing capabilities. However, these methods face challenges in accurately rendering complex 3D scenes. With the advent of implicit representations like NeRF~\cite{mildenhall2021nerf}, which offer high-fidelity rendering and significant scalability, extensive research has been conducted on 3D editing within the NeRF framework~\cite{haque2023instruct,wang2022clip,wang2023nerf}. Nevertheless, the implicit multi-layer perceptron (MLP) networks limit direct explicit modification of specific scene parts, complicating tasks such as repairing missing parts and scene composition. In contrast, 3DGS, with its advantages of real-time rendering, representation of complex scenes, and explicit representation, has naturally attracted considerable attention from researchers focusing on 3DGS editing. Unfortunately, current editable 3DGS works often lack precise training supervision, which greatly poses a challenge for editing. 
3DGS, with its advantages of explicit representation, has naturally attracted considerable attention from researchers focusing on 3DGS editing. Unfortunately, current editable 3DGS works often lack precise training supervision, which poses a significant challenge for editing. In this section, we categorize existing works according to different tasks.

% Similar to 3DGS in AIGC~\ref{AIGC}, they typically introduce 2D diffusion models to assist in supervising the 3DGS editing process as in Eq.~\ref{eq1}.

\subsubsection{Manipulation by Text} {To address this challenge, the existing works can be classified into two distinct categories. The first type introduces the \textit{\textbf{score distillation loss}} as in Eq.~\ref{eq3}. These methods require editing prompts as additional conditions to guide the editing process~\cite{chen2023gaussianeditor,palandra2024gsedit}. GaussianEditor~\cite{chen2023gaussianeditor} enables semantic-controlled 3DGS editing through SDS, utilizing hierarchical 3DGS and anchor loss for stability, while incorporating 2D inpainting guidance for object manipulation. Following Dreamgaussian, GSEdit~\cite{palandra2024gsedit} uses the pre-trained Instruct-Pix2Pix~\cite{brooks2023instructpix2pix} model instead of the image generation model for SDS. However, such methods remain constrained by pre-trained diffusion models, particularly when handling complex editing prompts.}

{The second type focuses on \textit{\textbf{editing multi-view images}} before reconstructing. GaussianEditor$^2$~\cite{fang2023gaussianeditor} employs multi-modal, language, and segmentation models to locate editable regions from text inputs, then optimizes targeted Gaussians based on 2D-edited images. However, this paradigm introduces an intuitive problem: \textit{\textbf{how to ensure consistency in multi-view editing}}~\cite{wu2024gaussctrl,wang2024view}.} GaussCtrl~\cite{wu2024gaussctrl} introduces a depth-guided image editing network, ControlNet~\cite{zhang2023adding}, utilizing its ability to perceive geometry and maintain multi-view consistency in the editing network. It also introduces a latent code alignment strategy in the attention layers, ensuring that the edited multi-view images remain consistent with the references.  

% The work~\cite{wang2024view} aims to introduce inverse rendering and rendering in 3D latent space to maintain consistency on the attention map. Furthermore, it introduces an editing consistency module and an iterative optimization strategy, further enhancing the multi-view consistency and editing capabilities.

{Unlike editing methods for 3DGS, recent discussions have increasingly focused on \textit{\textbf{editing 4DGS}}. Recent work, Control4D~\cite{shao2023control4d} leverages 4D GaussianPlanes to structurally decompose four-dimensional space, ensuring spatiotemporal consistency through Tensor4D representation, while incorporating a super-resolution GAN-based 4D generator~\cite{goodfellow2020generative} that learns from diffusion-generated edited images. However, it remains challenged on non-rigid movements.}

\subsubsection{Manipulation by Other Conditions}

In addition to text-controlled editing, existing works have explored 3DGS editing methods under various conditions, such as mixed conditions~\cite{zhuang2024tip} and points~\cite{huang2023point}. {TIP-Editor~\cite{zhuang2024tip} enables fine-grained 3DGS control through text, reference image, and location inputs, utilizing stepwise 2D personalization and coarse-fine editing strategies to support diverse tasks like object insertion and stylization. And Point'n Move~\cite{huang2023point} enables object-level editing through point annotations, utilizing a dual-stage process of segmentation, inpainting, and recomposition, which demonstrates improved control capability. Recent research~\cite{feng2024evsplitting} has introduced a training-free 3DGS splitting paradigm that achieves editing plane control by formulating it as a constrained minimization problem, preserving visual fidelity through moment conservation while mitigating Gaussian overflow via an analytically derived closed-form solution.}

% Point’n Move~\cite{huang2023point} requires the user to provide annotated points for the objects to be edited. It achieves controlled editing (including inpainting the removal area) of the objects through dual-stage segmentation, inpainting, and recomposition steps.

\subsubsection{Stylization}
In the realm of style transfer for 3DGS, early explorations have been made by~\cite{saroha2024gaussian}. Similar to traditional style transfer works~\cite{huang2022stylizednerf}, this work designs a 2D stylization module on the rendered images and a 3D color module on the 3DGS. By aligning the stylized 2D results of both modules, this approach achieves multi-view consistent 3DGS stylization without altering the geometry.

\subsubsection{Animation}\label{Animation}

As described in Sec.~\ref{4DGS}, some dynamic 3DGS works, such as SC-GS~\cite{huang2023sc}, can 
achieve animation effects by animating sparse control points. AIGC-related works, such as BAGS~\cite{zhang2024bags}, aim to utilize video input and generation models to animate existing 3DGS. Similar research has also been mentioned in the context of Human Reconstruction. Additionally, CoGS~\cite{yu2023cogs} discusses how to control animation. Based on dynamic representations~\cite{luiten2023dynamic,yang2023deformable}, it uses a small MLP to extract relevant control signals and align the deformation of each Gaussian primitives. 

\subsection{Relightable}\label{Relightable}
{Relightable 3DGS has also emerged as one of the recent challenges gaining significant attention. \textbf{\textit{Decoupling texture and lighting}} represents a common approach in relighting tasks. In early work, Relightable 3D Gaussian~\cite{gao2023relightable} and GS-IR~\cite{liang2024gs} represent scenes using points with normal, BRDF, and decomposed lighting attributes for relighting. However, these methods face challenges in handling reflective scenes. Therefore, follow-up work~\cite{jiang2024gaussianshader} introduce accurate normal estimation and residual color terms to effectively model view-dependent reflections and complex lighting interactions. To address limitations in \textbf{\textit{complex materials like semi-transparent volumes and furs}}, OLAT-GS~\cite{kuang2024olat} decomposes observed color into the attenuated light intensity, received incident illumination and scattering value. And then, GS$^3$~\cite{bi2024gs3} combines spatial and angular Gaussians with a triple splatting process to model geometry and reflectance properties, incorporating neural networks for self-shadowing and global illumination. Despite their capability in handling challenging geometry and appearance, further exploration is needed for transparent materials and indirect lighting.}

% Relightable 3D Gaussian (R3DG)~\cite{gao2023relightable,jiang2024gaussianshader,liang2024gs,shi2023gir} represents 3D scenes using relightable points, each characterized by a normal direction, BRDF parameters, and incident lighting, where the incident lights are decomposed into global and local components with view-dependent visibilities. Then, a novel point-based ray-tracing technique, based on bounding volume hierarchy, is designed in R3DG for efficient visibility baking and real-time rendering with accurate shadow effects.

\subsection{Semantic Understanding}\label{Semantic Understanding}

Endowing 3DGS with semantic understanding capabilities allows for the extension of 2D semantic models into 3D space, thereby enhancing the model's comprehension in 3D environments. This can be applied to various tasks such as 3D detection, segmentation, and editing. Many works attempt to leverage \textit{\textbf{pre-trained 2D semantic-aware models}} for extra supervision on semantic attributes~\cite{ye2023gaussian,zhou2023feature,lan20232d,dou2024cosseggaussians}. {Feature 3DGS~\cite{zhou2023feature} leverages pre-trained 2D models to create joint 3DGS and feature fields, enabling spatial understanding through feature rasterization and regularization for downstream promptable tasks. However, these approaches remain constrained by multi-view consistency and open-world perception challenges. Subsequent research~\cite{gu2025egolifter, choi2025click} aims to incorporate contrastive learning losses as auxiliary supervision to achieve interactive 3D segmentation. To mitigate the dependence on 2D pre-trained models in the aforementioned works, the work~\cite{yue2025improving} attempts to leverage reconstructed Featured 3DGS to fine-tune the 2D pre-trained encoder.}

% Subsequently, Gaussian Grouping~\cite{ye2023gaussian} introduces the concept of Gaussian groups and extends Identity Encoding attributes to achieve it. This work proposes treating multi-view data as a video sequence with gradually changing views and utilizes a pre-trained object tracking model~\cite{cheng2023tracking} to ensure multi-view consistency of the segmentation labels obtained from SAM (Segment Anything)\cite{kirillov2023segment}. Additionally, Gaussian groups are supervised in both 2D and 3D spaces and applied directly for editing. Similarly, the work\cite{lan20232d} addresses semantic inaccuracy by introducing KNN clustering and Gaussian filtering, which can constrain nearby Gaussians and eliminate distant Gaussians.

% CoSSegGaussians~\cite{dou2024cosseggaussians} leverages a pre-trained point cloud segmentation model in conjunction with a dual-stream feature fusion module. This module integrates unprojected 2D features derived from 2D encoders with 3D features from 3D encoders, following the prediction of Gaussian locations~\cite{kerbl20233d}. By employing decoders and semantic supervision, CoSSegGaussians effectively imbues Gaussian primitives with semantic information.

{Others focus on \textit{\textbf{incorporating text-visual alignment features}} for open-world understanding~\cite{shi2023language,qin2023langsplat,zuo2024fmgs}.}
A significant challenge is the high dimensionality of CLIP features, which makes direct training and storage difficult compared to original Gaussian attributes. The work~\cite{shi2023language} introduces corresponding continuous semantic vectors into the 3DGS by extracting and discretizing dense features from CLIP~\cite{radford2021learning} and DINO~\cite{caron2021emerging}, which are used to predict semantic indices $m$ in Discrete Feature Space by MLPs as in VQ-VAE~\cite{van2017neural}. {Subsequently, FMGS~\cite{zuo2024fmgs} mitigates the issue of large CLIP feature dimensions by introducing multi-resolution hash encoders~\cite{muller2022instant}}. 

% LangSplat~\cite{qin2023langsplat} compresses scene-specific CLIP features using a trained autoencoder to reduce training memory requirements. To achieve this objective, LangSplat introduces hierarchical semantics—subparts, parts, and wholes—constructed using SAM~\cite{kirillov2023segment}, which addresses point ambiguity across multiple semantic levels and facilitates scene understanding in response to arbitrary text queries. .

\subsection{Physics Simulation}\label{Physics Simulation}

\textit{\textbf{Recent efforts aim to extend 3DGS to simulation.}} Based on the "what you see is what you simulate" philosophy, PhysGaussian~\cite{xie2023physgaussian} reconstructs a static 3DGS as the discretization of the scene to be simulated, and then incorporates continuum mechanics theory along with the Material Point Method~\cite{de2020material} solver to endow 3DGS with physical properties. {Similar approaches have also been explored in PhysDreamer~\cite{zhang2025physdreamer}. Nevertheless, these physics-based simulations typically entail significant computational overhead. VR-GS~\cite{jiang2024vr} design an efficient interactive simulation system for VR, offering users a rich platform for 3D content manipulation. }

% Similar works have also been mentioned in dynamic 3DGS~\ref{4DGS}. For instance, the work~\cite{luiten2023dynamic} discusses regularization terms such as local rigidity, local rotational-similarity, and long-term local-isometry, which guide 4DGS correctly within the spatial domain. Moreover, the work~\cite{wu20234d} references the law of conservation of momentum to smooth dynamic motion trajectories.

{\section{3DGS Components and Improvements}}\label{Technical Improvements}

{3DGS can generally be divided into the following stages, as illustrated in Fig.~\ref{fig12}: initialization, attribute optimization, splatting, regularization, training strategy, and adaptive control. In follow-up studies, researchers have enhanced and refined the fundamental 3DGS components. These technical improvements not only enhance the rendering performance of the original 3DGS but also address specific tasks in derivative works. Consequently, this section delves into the technological improvements in 3DGS, with the aim of providing valuable insights for researchers in related fields.}

\subsection{Initialization}\label{Initialization}

Proper initialization has been shown to be crucial, as it directly affects the optimization process~\cite{jung2024relaxing}. The initialization of 3DGS is typically performed using sparse points derived from Structure-from-Motion (SfM) or through random generation. However, these methods are often unreliable, especially under weak supervision signals. 

\textit{\textbf{Combining pre-trained models}} is an optional approach. Pre-training a 3D model on a limited number of 3D samples and using it as an initialization prior is a viable strategy~\cite{xu2024agg}. This approach can enhance the performance of initialization to some extent, although its effectiveness is contingent upon the data used. To address this limitation, pretrained 3D generative models~\cite{chen2023text, yi2023gaussiandreamer, di2024hyper} or monocular depth estimation models~\cite{chung2023luciddreamer, ouyang2023text2immersion, paliwal2025coherentgs} are also introduced for initialization. Based on these, the work~\cite{chen2023text} also introduces some perturbation points to achieve a more comprehensive geometric representation. 

\textit{\textbf{Improving initialization strategies}} is also important. Based on the analysis of the role of SfM in capturing low-frequency signals within the spectrum, Sparse-large-variance initialization is designed to effectively focus on the low-frequency distribution identified by SfM~\cite{jung2024relaxing}.

\begin{figure}
\begin{center}
\includegraphics[width=0.45\textwidth]{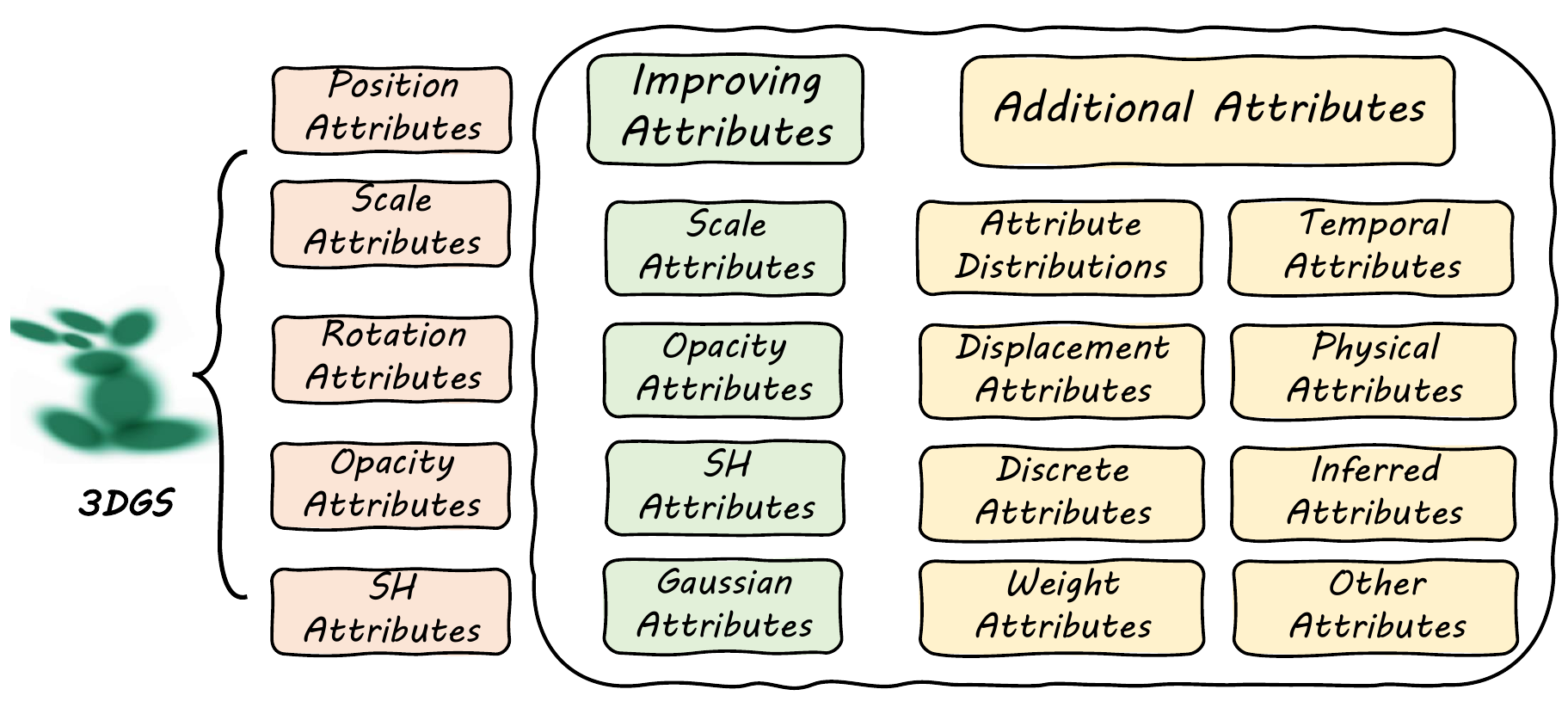}
\vspace{-0.3cm}
\end{center}
\caption{Overview of Attribute Expansion Strategies.}
\label{fig4}
\vspace{-0.6cm}
\end{figure}

\textit{\textbf{Utilizing other representations}} can also enhance initialization capabilities. By determining the Local Volumes from a coarse model, a small number of Gaussians are initialized within each volume, thereby avoiding excessive assumptions~\cite{das2023neural}. {Similarly, some initialization based on Visual Hull~\cite{yang2024gaussianobject}, Flame~\cite{zhao2024psavatar} or Mesh~\cite{teotia2024gaussianheads,gao2024mesh,kuang2024olat} are proposed, enabling the acquisition of structural geometric priors.}

{\textbf{Discussion}: The incorporation of pre-trained or structured information during the initialization of 3DGS is crucial, as high-fidelity initialization can mitigate training instabilities, particularly in under-determined scenarios. It is feasible to determine initialization protocols according to the available prior knowledge.}

\subsection{Attribute Expansion}\label{Attribute Expansion}

The original attributes of 3DGS include the position, scale, rotation, Spherical Harmonic (SH) coefficients, and opacity value. Some works have extended these attributes to make them more suitable for downstream tasks. It can be categorized into improvements of existing attributes or the introduction of novel attributes, as shown in Fig.~\ref{fig4}.

\subsubsection{Improving Attributes}\label{Improving Attributes}

{Certain attributes of vanilla Gaussian can be custom-tailored, thereby making 3DGS suitable for a wider range of tasks.}

\textbf{Scale:} By collapsing the $z$-scale to zero and incorporating additional supervision on depth, normal, or shell maps, the works~\cite{dai2024high, huang20242d, guedon2023sugar, cheng2024gaussianpro, abdal2023gaussian} aim to improve Gaussian primitives to make them flatter and more suitable for surface reconstruction, where $z$ direction can be approximated as the normal direction. Conversely, a scale constraint, which limits the ratio of the major axis length to the minor axis length~\cite{xie2023physgaussian, matsuki2023gaussian, herau20243dgs}, ensures that the Gaussian primitives remain spherical to mitigate the issue of unexpected plush artifacts caused by overly skinny kernels.

\textbf{SH:} By combining hash grids and MLP, the corresponding color attributes are encoded, effectively addressing the storage issues caused by a large number of SH parameters~\cite{lee2023compact}. 

% {Additionally, decomposing SH into a view- and lighting-dependent function represents a common approach for reconstructing complex materials and lighting~\cite{kuang2024olat,bi2024gs3}, which proves effective for anisotropic reflections.}

\textbf{Opacity:} By constraining the transparency to approach either 0 or 1, thereby minimizing the number of semi-transparent Gaussian primitives, the works~\cite{jiang2024gaussianshader, guedon2023sugar} achieve clearer Gaussian surfaces, effectively alleviating artifacts.

\textbf{Gaussian:} By introducing shape parameters, an attempt is made to replace the original Gaussians with a Generalized Exponential Family (GEF) mixture~\cite{hamdi2024ges}. Traditional 3DGS can be viewed as a special case of the GEF mixture ($\beta=2$), enhancing the representational efficiency of Gaussians,
\begin{equation}
     \hat{G}(\bm{x}) = \exp \left\{ -\frac{1}{2} \left( (\bm{x} - \bm{\mu})^\top \Sigma^{-1} (\bm{x} - \bm{\mu}) \right)^{\frac{\beta}{2}} \right\}. \label{eq4}
\end{equation}
\subsubsection{Additional Attributes}\label{Additional Attributes}
{By adding new attributes and corresponding supervisions, the original representation capabilities of 3DGS can be augmented.}

\textbf{Semantic Attributes:} {By introducing them and corresponding supervision, works such as~\cite{yan2024street, zhou2024hugs, li2024sgs, zhu2024semgauss, zhou2023feature, ji2024neds,yue2025improving,choi2025click,gu2025egolifter} are endowed with enhanced spatial semantic awareness, which is crucial for tasks such as SLAM and editing.} After the semantic attributes' splatting, the 3DGS's semantic attributes are supervised using 2D semantic segmentation maps. Additionally, methods to improve the extraction of semantic information~\cite{huang2023point} and introducing high-dimensional semantic-text, such as CLIP and DINO features~\cite{qin2023langsplat, shi2023language, zuo2024fmgs}, have been employed to address a wider range of downstream tasks. Similar to semantic attributes, the identity encoding attributes can group 3DGS that belong to the same instance or stuff~\cite{ye2023gaussian}, which is more effective for multi-object scenes.

\textbf{Attribute Distributions:} Learning position distributions with reparameterization techniques instead of a fixed value is an effective approach to prevent local minima~\cite{li2023gaussiandiffusion} and mitigate its reliance on Adaptive Control of 3DGS~\cite{charatan2023pixelsplat}. In addition to these works focusing on the distribution prediction of position attributes, the distribution of the scale has also been incorporated~\cite{li2023gaussiandiffusion}. By sampling the predicted distributions, Gaussian primitives can be obtained.

\textbf{Temporal Attributes:} Replacing the original static attributes with temporal attributes is key to animating the 3DGS~\cite{yang2023real,pan2024fast,yan2024street,duan20244d}. For 4D attributes, including rotation, scale, and position, existing works render 3DGS on timestep $t$ by either taking time slices~\cite{duan20244d} or decoupling the $t$ dimension from 4D attributes~\cite{pan2024fast,yang2023real}. Also, the introduction of 4D SH is crucial for time-varying color attributes. For this, the Fourier series is typically used as the adopted basis functions to endow SH with temporal capabilities~\cite{yan2024street, yang2023real}. Note that due to involving different timesteps, these attributes often require video-based training. This regularization primarily aims to improve attributes in Gaussian primitives~\cite{matsuki2023gaussian,cheng2024gaussianpro,xie2023physgaussian,zhou2024gala3d, teotia2024gaussianheads}, as in Sec.~\ref{Attribute Expansion}.

\textbf{Displacement Attributes:} They can describe the relationship between the final and initial attributes in Gaussian primitives and be classified based on their condition. Condition-independent displacement attributes are often used to refine coarse attributes, which can be directly optimized in the same manner as other attributes~\cite{szymanowicz2023splatter}. Condition-dependent displacement attributes can describe the changes of static 3DGS, thereby achieving dynamic representations and controllable representations. This approach often involves introducing a small MLP to predict displacement based on timestep $t$~\cite{wu20234d, yang2023deformable, liang2023gaufre}, {expression and other control signals\cite{yu2023cogs, lee2025deblurring,dhamo2025headgas,xu2024gaussian,chen2024monogaussianavatar,ma20243d,xiang2024flashavatar,xu20253d,qian20233dgsavatar}.}

\textbf{Physical Attributes:} They encompass a broad range of properties describing the physical laws governing Gaussian primitives, thus endowing 3DGS with more realistic representation. For instance, shading-related attributes like diffuse color, {direct specular reflection, residual color, shadow, and anisotropic spherical Gaussian can be used for specular reconstruction and relighting~\cite{jiang2024gaussianshader, yang2024spec, meng2024mirror,kuang2024olat,bi2024gs3,duisterhof2023md}. Additionally, the velocity attributes can represent the transient information of Gaussian, essential for describing dynamic objects~\cite{guo2024motion}.} These attributes are typically optimized by considering the influence of physical laws at specific rendering positions~\cite{jiang2024gaussianshader, duisterhof2023md, yang2024spec} or by incorporating supplementary information, such as flow maps~\cite{guo2024motion}.

\textbf{Discrete Attributes:} Utilizing discrete attributes in place of continuous ones is an effective method for compressing high-dimensional representations and representing complex motion. This is often achieved by storing the index values of the VQ codebook~\cite{navaneet2023compact3d, lee2023compact, niedermayr2023compressed, girish2023eagles,roessle2024l3dg} or the motion coefficient for motion basis~\cite{kratimenos2023dynmf} as the discrete attributes in Gaussian primitives. However, discrete attributes inevitable lead to performance degradation; combining them with compressed continuous attributes may be a potential solution~\cite{shi2023language}.

\textbf{Inferred Attributes:} These attributes do not require optimization; they are inferred from other attributes. The \textit{Parameter-Sensitivity attributes} reflects the impact of parameter changes on reconstruction performance and are represented by the gradient of the parameter, guiding compression clustering~\cite{niedermayr2023compressed}. The \textit{Pixel-Coverage attributes} determines the relative size of Gaussian primitives at the current resolution. It is related to the horizontal or vertical size of the Gaussian primitives and guides their scale to meet sampling requirements in multi-scale rendering~\cite{yan2023multi}.

\textbf{Weight Attributes:} They rely on structured representations, such as Local Volumes~\cite{das2023neural}, Gaussian-kernel RBF~\cite{huang2023sc}, Mesh~\cite{gao2024mesh}, and SMPL~\cite{lei2023gart}, to determine the attributes of query points by calculating the weights of structured points.

\textbf{Other Attributes:} The \textit{Uncertainty Attributes} can help maintain training stability by reducing the loss weight in areas with high uncertainty~\cite{shi2023language, swann2024touch}. The \textit{ORB-Features Attributes}, extracted from image frames~\cite{rublee2011orb}, play a crucial role in establishing 2D-to-2D and 2D-to-3D correspondences~\cite{huang2023photo}.

{\textbf{Discussion}: The modification of Gaussian attributes facilitates the execution of a wider range of downstream tasks, offering an efficient approach as it obviates the need for additional structural elements. Moreover, the integration of new attributes with supplementary information constraints also has the potential to significantly enhance the representational efficacy of the original 3DGS. For instance, semantic attributes can, in certain scenarios, yield more precise object boundaries.}

\subsection{Splatting}\label{Splatting}

The role of Splatting is to efficiently transform 3D Gaussian data into high-quality 2D images, ensuring smooth, continuous projections and significantly improving rendering efficiency. As a core technology in traditional computer graphics, there are also efforts aimed at improving it from the perspectives of efficiency and performance.

% \begin{figure}
% \begin{center}
% \includegraphics[width=0.40\textwidth]{Figures/survey_fig10.pdf}
% \vspace{-0.6cm}
% \end{center}
% \caption{The Modified Splatting Strategy~\cite{huang2024gs++}.}
% \label{fig10}
% \vspace{-0.5cm}
% \end{figure}

% Introducing the ADOP~\cite{ruckert2022adop} in 3DGS for real-time rendering, TRIPS~\cite{franke2024trips} utilizes a screen-space image pyramid for point rasterization and employs trilinear write for rendering large points. This approach, combined with front-to-back alpha blending and a lightweight neural network for detail reconstruction, ensures crisp, complete, and alias-free images.

% The work~\cite{huang2024gs++} identifies limitations in prior methods~\cite{kerbl20233d} that use local affine approximations during the projection process, resulting in errors detrimental to rendering quality. 

{Several studies focus on \textit{\textbf{enhancing splatting mechanisms}}. By analyzing the residual errors from the first-order Taylor expansion, the work~\cite{huang2024gs++} establishes a correlation between these errors and the Gaussian mean position and an unified projection plane to mitigate the projection errors through the Unit Sphere Based Rasterizer. Analytic-Splatting~\cite{liang2025analytic} advances pixel-center splatting through the introduction of pixel-window-based approximation, archiving anti-aliasing. To enhance 3DGS performance in complex scenarios, a GPU-accelerated ray tracing algorithm is introduced~\cite{moenne20243d} for semi-transparent particle-based representations, achieving real-time rendering with support for complex effects such as secondary rays, depth of field, and distorted cameras.} {Additional improvements focus on \textit{\textbf{optimizing splat ordering}} during the blending process. StopThePop~\cite{radl2024stopthepop} proposes a hierarchical per-pixel sorting strategy, which eliminates popping artifacts and ensures view-consistent real-time rendering by accurately computing the depth of Gaussians along individual rays.}

\subsection{Regularization}~\label{Regularization}
\begin{figure*}
\begin{center}
\includegraphics[width=0.90\textwidth]{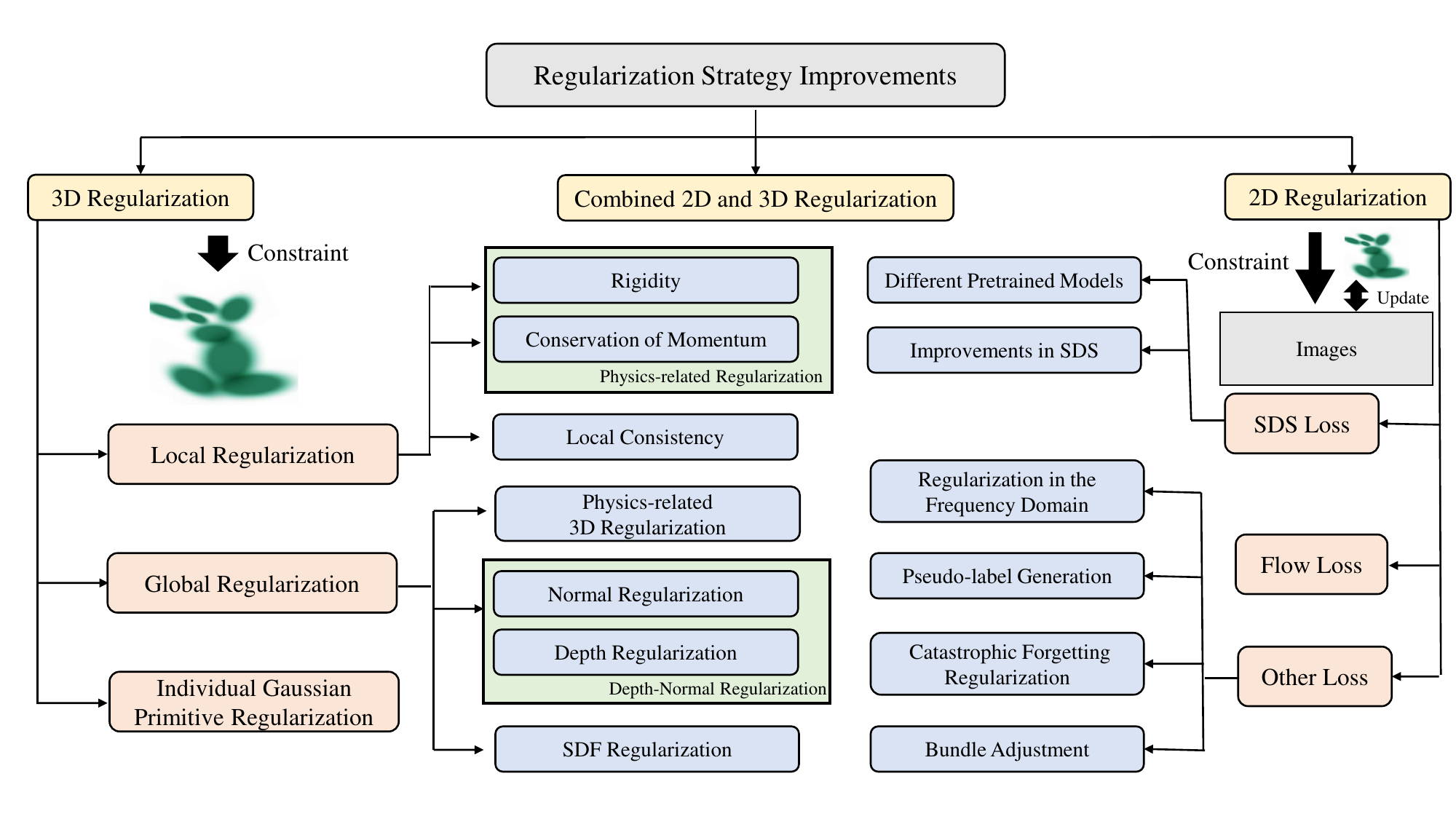}
\vspace{-0.6cm}
\end{center}
\caption{Overview of Regularization Strategy Improvement in existing works.}
\label{fig5}
\vspace{-0.5cm}
\end{figure*}

Regularization is crucial for 3D reconstruction. We categorize the regularization terms into 2D and 3D regularization, as shown in Fig.~\ref{fig5}. The 3D regularization directly constrains 3DGS, while the 2D regularization imposes constraints on the rendered images, thus influencing attribute optimization.

\subsubsection{3D Regularization}~\label{3D Regularization}
The 3D regularization has garnered significant attention due to its intuitive constraint capabilities. These efforts categorized based on their targeted objectives into individual Gaussian primitive, local, and global regularization.

\textbf{Individual Gaussian Primitive Regularization:} 
This regularization primarily aims to improve attributes in Gaussian primitives~\cite{matsuki2023gaussian,cheng2024gaussianpro,xie2023physgaussian,zhou2024gala3d, teotia2024gaussianheads}, as Sec.~\ref{Attribute Expansion}.

\textbf{Local Regularization:} Owing to the explicit representation of 3DGS, it is meaningful to impose constraints on Gaussian primitives within local regions. Such constraints can ensure the continuity and feasibility of Gaussian primitives in the local space. \textit{Physics-related Regularization} is often used to ensure the local rigidity of deformable targets, which includes short-term local-rigidity loss, local rotation similarity loss, and long-term local isometry loss. Short-term local rigidity implies that nearby Gaussians should move following a rigid body transformation between time steps,\\
\begin{equation}
     \mathcal{L}_{i,j}^{\text{rigid}} = w_{i,j} \left\| (\bm{\mu}_{j,t-1} - \bm{\mu}_{i,t-1}) - \bm{R}_{i,t-1} \bm{R}_{i,t}^{-1} (\bm{\mu}_{j,t} - \bm{\mu}_{i,t}) \right\|_2, \label{eq5}
\end{equation}
where $\bm{\mu}$ is Gaussian mean position, $i$ and $j$ are indexing of neighboring points, $t$ is timestep, and $\bm{R}$ reperents rotation; Local rotation similarity enforces that adjacent Gaussian primitives have the same rotation over time steps, \\
\begin{equation}
     \mathcal{L}_{i,j}^{\text{rot}} = w_{i,j} \left\| \hat{\bm{q}}_{j,t} \hat{\bm{q}}_{j,t-1}^{-1} - \hat{\bm{q}}_{i,t} \hat{\bm{q}}_{i,t-1}^{-1} \right\|_2, \label{eq6}
\end{equation}
where $\hat{\bm{q}}$ is the normalized quaternion representation of each Gaussian’s rotation; Long-term local isometry loss prevents elements of the scene from drifting apart,\\
\begin{equation}
     \mathcal{L}_{i,j}^{\text{iso}} =w_{i,j} \left\lvert \|\bm{\mu}_{j,0} - \bm{\mu}_{i,0}\|_2 - \|\bm{\mu}_{j,t} - \bm{\mu}_{i,t}\|_2 \right\rvert, \label{eq7}
\end{equation}
{\cite{luiten2023dynamic,das2023neural,huang2023sc,duisterhof2023md,ling2023align,gao2024gaussianflow,kratimenos2023dynmf,yu2023cogs}.} Subsequently, some works have also adopted similar paradigms to constrain local rigidity{~\cite{lin2023gaussian, zhang2024bags, li2024st}}. 

In addition to rigidity loss, \textit{Conservation of Momentum Regularization} can also be used as a constraint in dynamic scene reconstruction. It encourages a constant velocity vector and applies a low-pass filter effect to the 3D trajectories, thereby smoothing out trajectories with sudden changes~\cite{duisterhof2023md}. In addition, there are some \textit{Local Consistency Regularization} terms that also aim to constrain the Gaussian primitives within local regions to maintain similar attributes, such as {semantic attributes~\cite{ye2023gaussian, shi2023language, lan20232d}, position~\cite{vilesov2023cg3d, yin20234dgen}, time~\cite{lin2023gaussian}, frame~\cite{shaw2023swags}, normal~\cite{li2024geogaussian}, weight~\cite{10.1145/3680528.3687659} and depth~\cite{paliwal2025coherentgs,chung2023depth}.}

\textbf{Global Regularization:} Unlike the local regularization within neighboring regions, global regularization aims to constrain the overall 3DGS. \textit{Physics-related Regularization} introduces real-world physical laws to constrain the state of 3DGS, including gravity loss and contact loss, among others. Gravity loss is used to constrain the relationship between the object and the floor, while contact loss targets the relationships among multiple objects~\cite{vilesov2023cg3d}.

Benefiting from the explicit representation, depth and normal attributes can be directly calculated and constrained during training, particularly for surface reconstruction tasks. \textit{Depth-Normal Regularization} achieves depth-normal consistency by comparing the normal computed from depth values with the predicted normal~\cite{dai2024high,jiang2024gaussianshader,lyu20243dgsr,yu2024gaussian}. This method enforces constraints on both normal and depth simultaneously. Additionally, directly constraining either the normal or the depth is also feasible. \textit{Normal Regularization} often adopts a self-supervised paradigm due to the lack of direct supervision signals, which can be implemented by designing pseudo-labels from gradients~\cite{chen2023neusg}, the shortest axis direction of Gaussian primitives~\cite{jiang2024gaussianshader}, or SDF~\cite{yu2024gsdf,lyu20243dgsr}. Similarly, \textit{Depth Regularization} adopts a similar approach; however, it not only aims for accurate depth values but also seeks to ensure clear surfaces in 3DGS. Depth Distortion loss~\cite{barron2022mip} aggregates Gaussian primitives along the ray,\\
\vspace{-0.3cm}
\begin{equation}
     \mathcal{L}_d = \sum_{i,j} \omega_i \omega_j \lvert z_i - z_j \rvert, \label{eq8}
\end{equation}
\vspace{-0.4cm}

\noindent where $z$ is the intersection depth of Gaussian~\cite{huang20242d,yu2024gaussian}. In addition to self-supervised methods, incorporating additional pre-trained models to estimate normal~\cite{dai2024high} and depth~\cite{meng2024mirror,li2024dngaussian,zhu2023fsgs,xiong2023sparsegs} has proven to be more effective in \textit{Normal Regularization} and \textit{Depth Regularization.} Derived works introduce hard depth and soft depth regularization to address the geometry degradation and achieve more complete surfaces~\cite{li2024dngaussian}. Similarly, SDF Regularization is also a constraint strategy for surface reconstruction. It achieves the desired surface by constraining the SDF that corresponds to 3DGS to an ideal distribution~\cite{guedon2023sugar,chen2023neusg,yu2024gsdf,lyu20243dgsr,bolanos2024gaussian}.

\subsubsection{2D Regularization}~\label{2D Regularization}
Unlike the intuitive constraints in 3D, 2D regularization is often used to address under-constrained situations where original loss functions alone are insufficient.

{\textbf{SDS loss:} An important example is the SDS loss, as shown in Eq.\ref{eq3}, which uses a pre-trained 2D diffusion model to supervise 3DGS training via distillation paradigms~\cite{tang2023dreamgaussian,chen2023gaussianeditor}.} 
\begin{equation}
     \nabla_{\theta } =\mathbb{E} \left [ w_{t}\left ( \epsilon_{\phi}\left ( \bm{x_{t}},t,y \right )-\epsilon   \right ) \frac{\partial \bm{x}}{\partial \theta}  \right ]. \label{eq3}
\end{equation}
This approach extends to distill pre-trained 3D diffusion models~\cite{nichol2022point}, multi-view diffusion models~\cite{liu2023syncdreamer}, image editing models~\cite{brooks2023instructpix2pix}, and video diffusion models. Introducing 3D~\cite{chen2023text,li2024controllable} and {multi-view diffusion models~\cite{ren2023dreamgaussian4d,zhang2023repaint123,zhou2024gala3d,ling2023align,gao2024gaussianflow,yin20234dgen,liu2024sketchdream}} enhances geometry and multi-view consistency. Image editing models~\cite{palandra2024gsedit} enable controllable edits, while video diffusion models~\cite{ling2023align} support dynamic temporal scene generation. Additionally, distillation on multi-modal images, like RGB-Depth~\cite{liu2023humangaussian}, also holds potential, providing more constraints from pre-trained diffusion models.

Some improvements specifically target inherent issues in SDS~\cite{yang2023learn,liang2023luciddreamer}. Interval Score Matching is proposed to address issues of randomness and single-step sampling. \\
\begin{equation}
     \nabla_\theta \mathcal{L}_{\text{ISM}}(\theta) := \mathbb{E}_{t, c} \left[ \omega(t) \left( \epsilon_\phi(\bm{x_t}, t, y) - \epsilon_\phi(\bm{x_s}, s, \emptyset) \right) \frac{\partial g(\theta, c)}{\partial \theta} \right], \label{eq9}
\end{equation}
where $s = t - \delta_T$ and $\delta_T$ denotes a small step size~\cite{liang2023luciddreamer}. Introducing Negative Prompts~\cite{armandpour2023re} is a method~\cite{liu2023humangaussian,liang2023luciddreamer,li2024controllable} to mitigate the impact of random noise $\epsilon$ and enhance stability by replacing random noise with negative prompts $[\epsilon_\phi (\bm{x}_t; y_{\text{neg}})]$. And, LODS incorporates LoRA terms~\cite{hu2021lora} $[\frac{(1 - w) \epsilon_\psi (\bm{x_t}; \emptyset, t)}{w} - \frac{\epsilon}{w}]$ to replace traditional random noise $\epsilon$, thereby alleviating the impact of out-of-distribution~\cite{yang2023learn}.

\textbf{Flow loss:} It is a commonly used regularization term for dynamic 3DGS and uses the output of a pre-trained 2D optical flow estimation model as ground truth. Predicted flow is rendered by calculating the displacement of Gaussian primitives over a unit time and splatting these 3D displacements onto a 2D plane~\cite{katsumata2023efficient,gao2024gaussianflow,zhou2024hugs,li2024st}. However, this approach has a significant gap since optical flow is a 2D attribute and susceptible to noise. Selecting Gaussians with correct depth and introducing uncertainty through KL divergence to constrain optical flow is a potentially feasible method~\cite{guo2024motion}.

% In addition to the rendered images serving as supervision, the rendering results of other attributes can also be supervised. \textit{Depth loss} is a straightforward idea, where the depth map, as a byproduct of 3DGS, can be directly supervised by the ground truth depth image~\cite{yugay2023gaussian}.

\textbf{Other loss:} There are also some 2D regularization terms worth discussing. For example, constraining the differences in amplitude and phase between the rendered image and the ground truth in the frequency domain can serve as a loss function to aid training, thereby alleviating overfitting issues~\cite{zhang2024fregs}. Introducing pseudo-labels for hypothetical viewpoints through noise perturbation can assist training in sparse-view settings~\cite{zhu2023fsgs}. In large-scale scene mapping, constraining the changes in attributes before and after optimization can prevent catastrophic forgetting in 3DGS~\cite{sun2024high}. Additionally, bundle adjustment is usually an important constraint in pose estimation problems~\cite{yan2023gs,huang2023photo,zhu2024semgauss}.

Noted that, whether 2D or 3D regularization is used, overall updating is sometimes suboptimal due to the large number of primitives. Some primitives often have an uncontrollable impact on the results. Therefore, it is necessary to guide the optimization by selecting important primitives using methods such as visibility~\cite{li2024sgs,keetha2023splatam,lei2024gaussnav,ji2024neds}.

{\textbf{Discussion}: The incorporation of regularization terms serves as an effective approach to enhance the reconstruction performance of 3DGS. These regularization terms can impose constraints on various attributes of 3DGS, including geometry and spatial distribution, etc., in accordance with specific task requirements. Furthermore, in under-determined scenarios, the optimization process can be further constrained by introducing additional prior information. For a particular task, it is feasible to incorporate multiple distinct constraints simultaneously.}

\subsection{Training Strategy}\label{Training Strategy}

Training strategy is also an important topic. In this section, we divide it into multi-stage training strategy and end-to-end training strategy, which can be applied to different tasks.

\subsubsection{Multi-stage Training Strategy}

Multi-stage training strategy is a common training paradigm, often involving coarse-to-fine reconstruction. It is widely used for under-determined tasks, such as AIGC, SLAM, etc..

Using different 3D representations in different training stages is a typical multi-stage training paradigm. 3DGS $\rightarrow$ Mesh (training 3DGS first, converting to Mesh, then optimizing Mesh)~\cite{tang2023dreamgaussian,li2024controllable,zhang2023repaint123,ren2023dreamgaussian4d,palandra2024gsedit,tang2024lgm} ensures geometric consistency in the generated 3D model. Additionally, generating multi-view images~\cite{pan2024fast,melas20243d,li2024controllable,feng2024fdgaussian,fang2023gaussianeditor,wu2024gaussctrl,wang2024view} in the first stage to aid reconstruction in the second stage can alleviate optimization difficulties.

Two-stage reconstruction for static and dynamic reconstruction is also important in dynamic 3DGS. This type of work typically involves training a time-independent static 3DGS in the first stage, and then training a time-dependent deformation field in the second stage to characterize dynamic Gaussians~\cite{wu20234d,katsumata2023efficient,yang2023deformable,duisterhof2023md,shao2023control4d,liang2023gaufre}. Additionally, incremental reconstruction of dynamic scenes frame by frame is also a focus in some works, often relying on the performance of previous reconstructions ~\cite{luiten2023dynamic, sun20243dgstream}.

In multi-objective optimization tasks, multi-stage training paradigms can enhance stability and performance. For example, the \textit{coarse-to-fine camera tracking strategy} first obtains a coarse camera pose from a sparse pixel set, then refines it based on optimized rendering results~\cite{yan2023gs, jiang20243dgs}.

Additionally, some works aim to refine the 3DGS trained in the first stage~\cite{yang2024gaussianobject, chen2023text, lan20232d, zhuang2024tip, das2023neural, di2024hyper, xu2024agg, liu2024sketchdream} or endow them with additional capabilities, such as semantics~\cite{dou2024cosseggaussians, lei2024gaussnav} and stylization~\cite{saroha2024gaussian}. There are many such training strategies, which are also effective in maintaining training stability and avoiding local optima~\cite{fan2023lightgaussian}. Furthermore, iterative optimization of the final result to enhance performance is also feasible~\cite{wang2024view, liu2025deceptive}.

\subsubsection{End-to-End Training Strategy} 
\begin{figure}[!tp]
\centering
\subfigure[Progressive Optimization~\cite{yang2024spec}.]{
           \includegraphics[height=0.10\textheight]{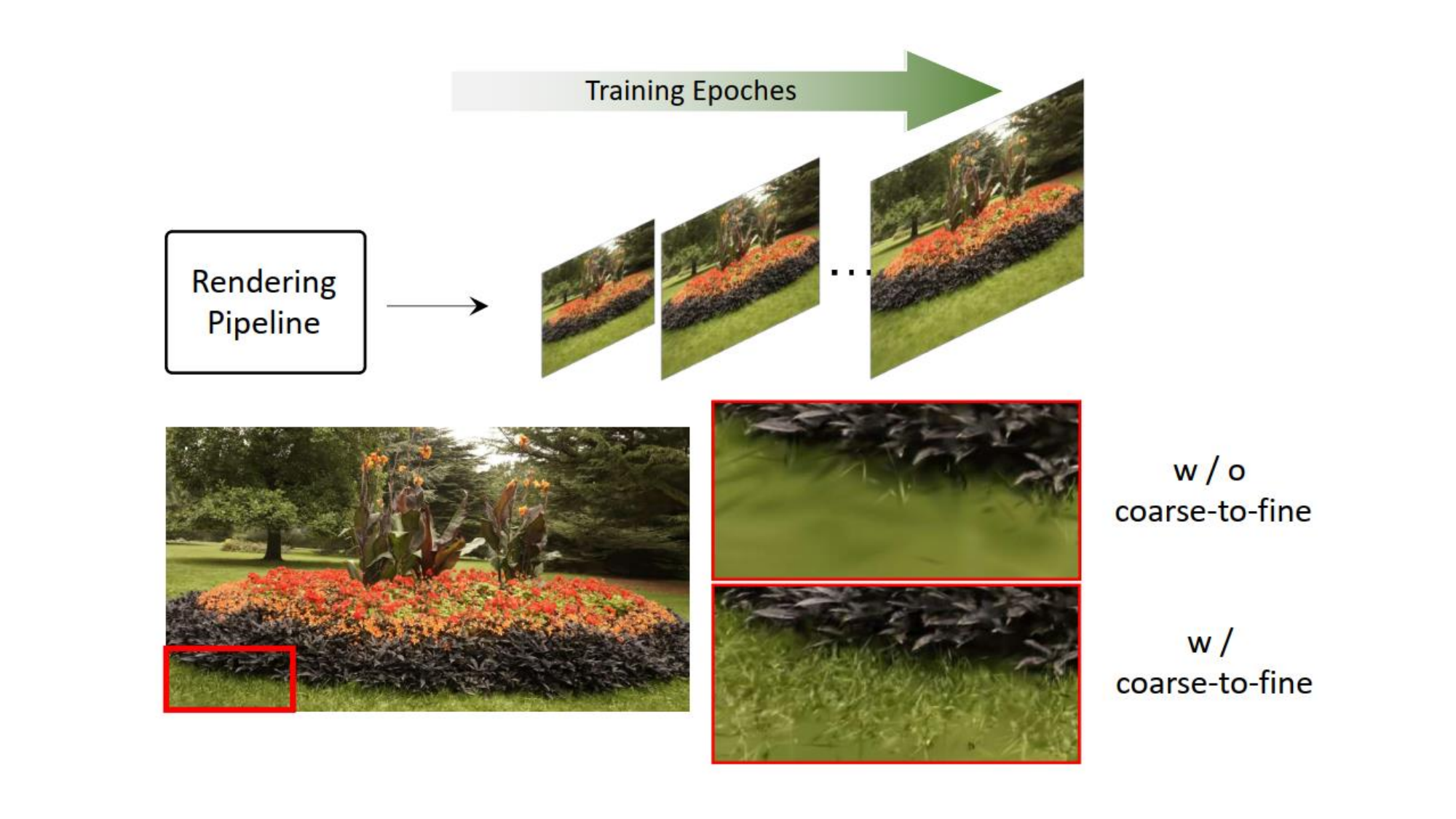}}
\subfigure[Block Optimization~\cite{lei2024gaussnav}.]{
           \includegraphics[height=0.09\textheight]{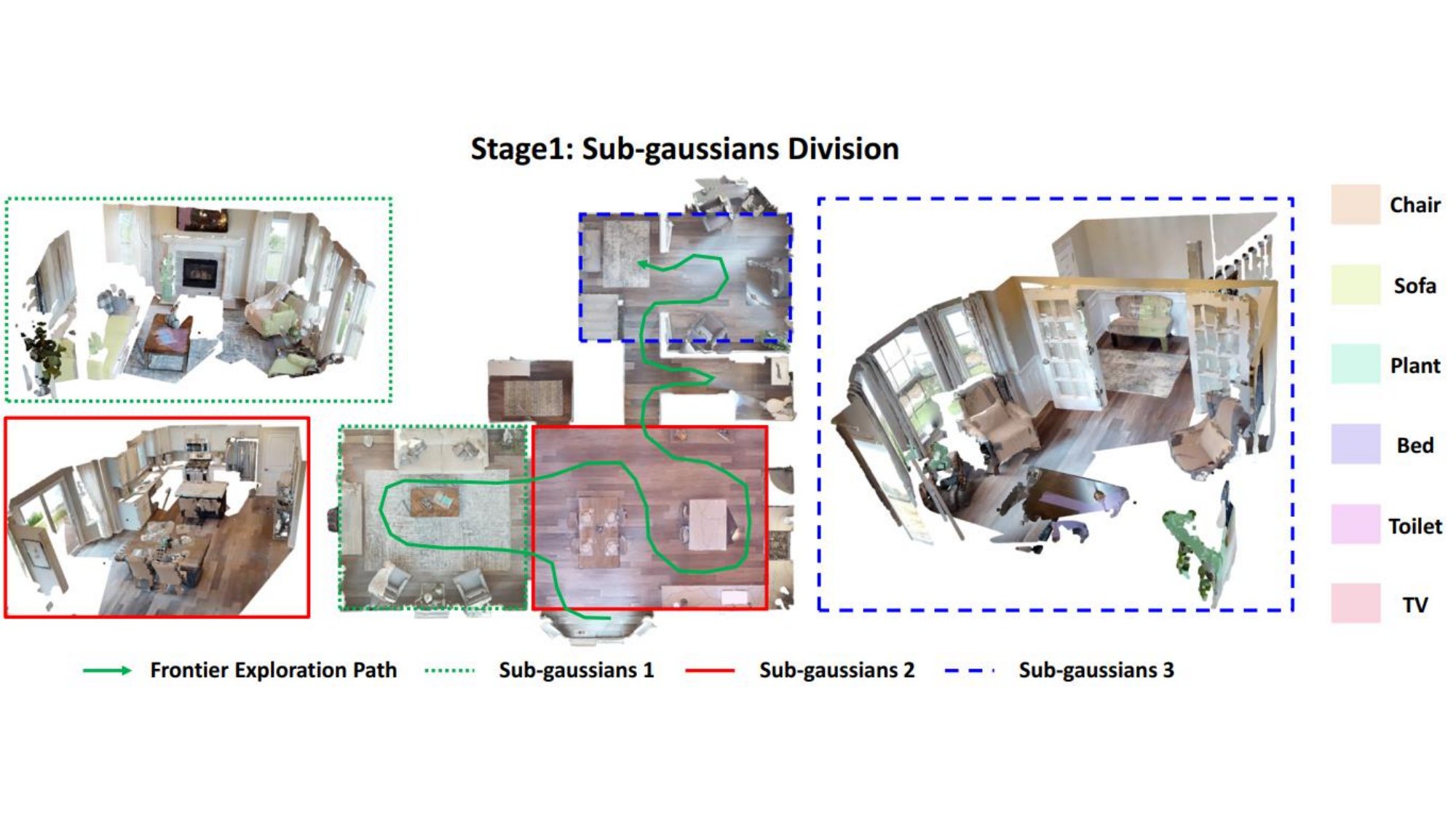}}
\subfigure[Robust Optimization~\cite{yang2024gaussianobject}.]{
           \includegraphics[height=0.10\textheight]{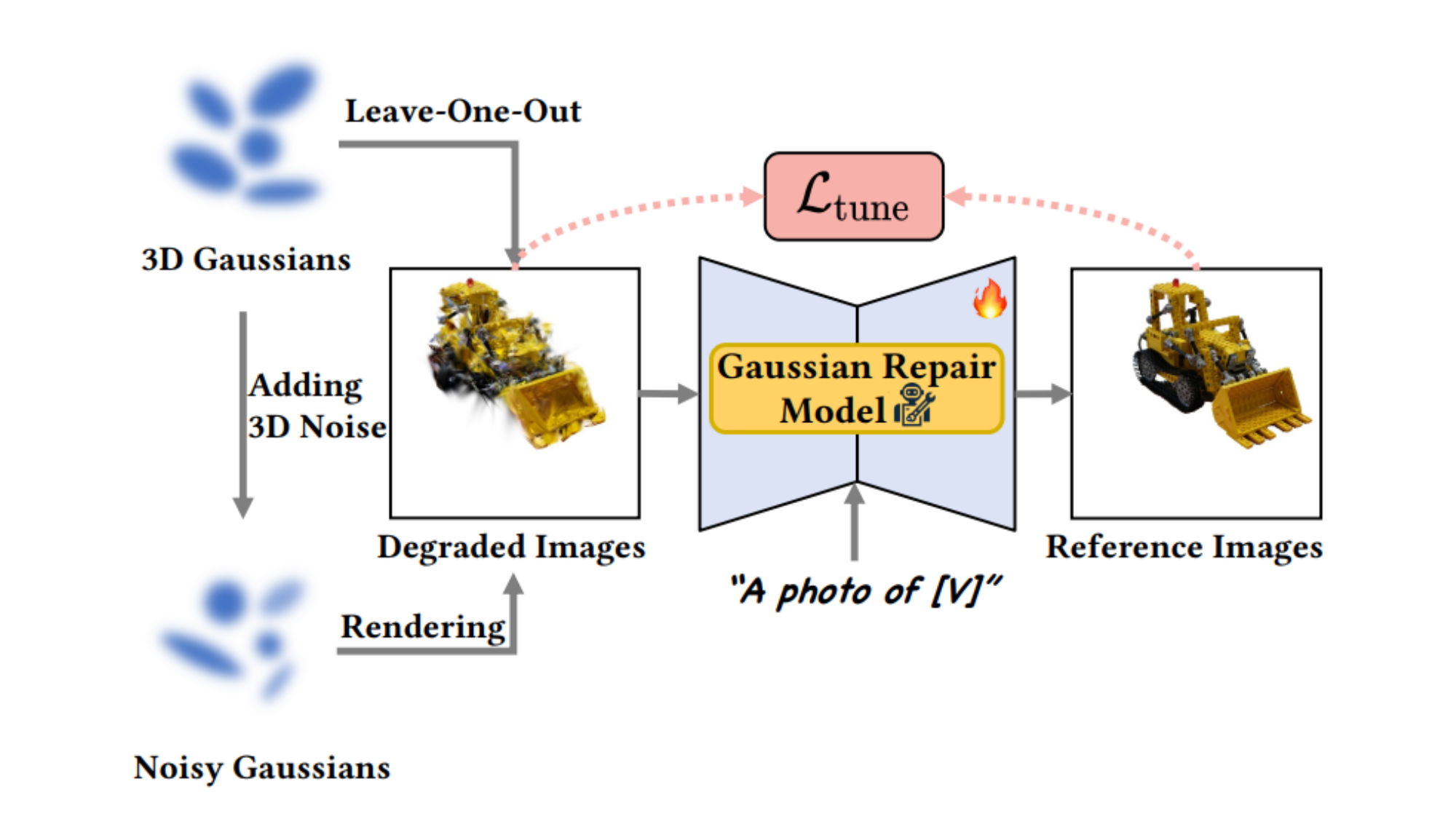}}
\subfigure[Distillation Strategy~\cite{fan2023lightgaussian}.]{
           \includegraphics[height=0.09\textheight]{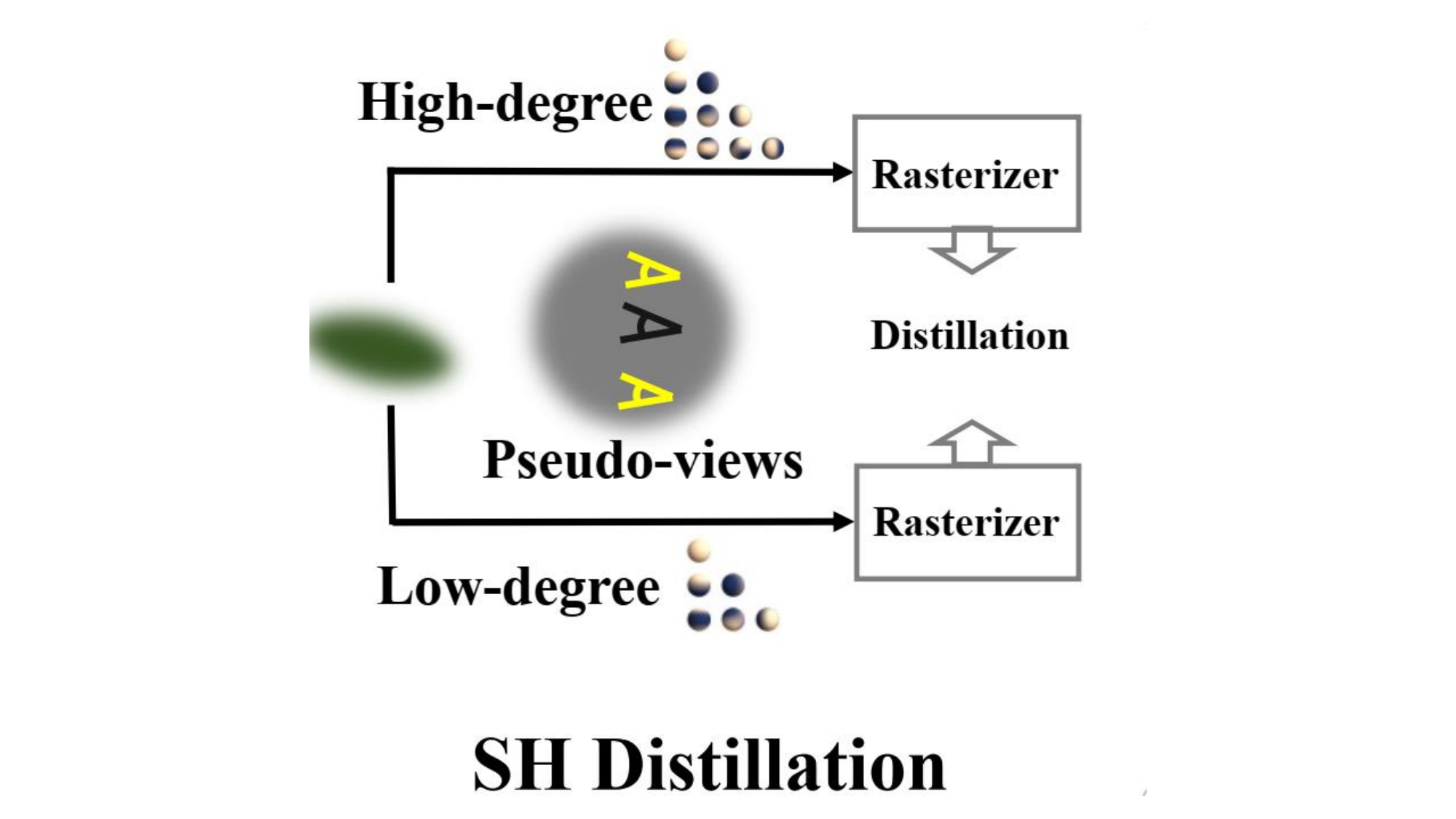}}
\vspace{-0.1cm}
\caption{Four typical End-to-End Training Strategies.}
\label{fig6}
\vspace{-0.6cm}
\end{figure}

These strategies are often more efficient and can be applied to a wider range of downstream tasks. Some typical works are described in Fig.~\ref{fig6}.

% \begin{figure}
%     \centering
%     \begin{subfigure}[b]{0.24\textwidth}
%         \centering
%         \includegraphics[width=\textwidth]{Figures/survey_fig6-1.pdf} % 这里替换为你的第一个小图
%         \caption{Progressive Optimization~\cite{yang2024spec}}
%         \label{Text-3D Objects}
%     \end{subfigure}
%     \hfill
%     \begin{subfigure}[b]{0.24\textwidth}
%         \centering
%         \includegraphics[width=\textwidth]{Figures/survey_fig6-2.pdf} % 这里替换为你的第二个小图
%         \caption{Block Optimization~\cite{lei2024gaussnav}}
%         \label{fig:subfig2}
%     \end{subfigure}
%     \vspace{-0.8cm}
%     \vskip\baselineskip
%     \begin{subfigure}[b]{0.24\textwidth}
%         \centering
%         \includegraphics[width=\textwidth]{Figures/survey_fig6-3.pdf} % 这里替换为你的第三个小图
%         \caption{Robust Optimization~\cite{yang2024gaussianobject}}
%         \label{fig:subfig3}
%     \end{subfigure}
%     \hfill
%     \begin{subfigure}[b]{0.24\textwidth}
%         \centering
%         \includegraphics[width=\textwidth]{Figures/survey_fig6-4.pdf} % 这里替换为你的第四个小图
%         \caption{Distillation-based Strategy~\cite{fan2023lightgaussian}}
%         \label{fig:subfig4}
%     \end{subfigure}
%     \caption{End-to-End Training Strategies}
%     \label{fig6}
%      \vspace{-0.8cm}
% \end{figure}

\textbf{Progressive Optimization Strategy:} This commonly used strategy helps 3DGS prioritize learning global representations before locally optimizing details. In the frequency domain, this can be viewed as progressively learning from low-frequency to high-frequency components. It is often implemented by gradually increasing the proportion of high-frequency signals~\cite{jung2024relaxing, zhang2024fregs} or introducing progressively larger image/feature sizes for supervision~\cite{girish2023eagles, yang2024spec, huang2023photo}, which can also improve efficiency~\cite{herau20243dgs,peng2025bags}. In generative tasks, progressively selecting the camera pose is also an easy-to-difficult training strategy, optimizing from positions close to the initial viewpoint to those further away~\cite{zhang2023repaint123, ouyang2023text2immersion}.

\textbf{Block Optimization Strategy:} This strategy is often used in large-scale scene reconstruction to improve efficiency and alleviate catastrophic forgetting~\cite{yugay2023gaussian, lei2024gaussnav, jiang20243dgs}. However, such paradigms are often influenced by block partitioning and training data selection. {Consequently, several studies have proposed designing Primitives and Data Division strategies to mitigate workload imbalances caused by numerous empty blocks, while enhancing detail reconstruction capabilities~\cite{liu2025citygaussian}. To improve efficiency, introducing Level of Detail and hierarchical reconstruction prove effective, especially in large-scale scene processing~\cite{liu2025citygaussian,kerbl2024hierarchical}.} It can also achieve reconstruction by partitioning the scene into static backgrounds and dynamic objects~\cite{zhou2023drivinggaussian, yan2024street, zhou2024hugs, liang2023gaufre}. Additionally, this approach is applied in AIGC and Semantic Understanding, where refining submap reconstruction quality enhances overall performance~\cite{di2024hyper, qin2023langsplat}. Unlike submaps divided by spatial regions, Gaussians can be categorized into different generations during their densification process, allowing for the application of distinct regularization strategies to each generation, effectively regulating their fluidity~\cite{chen2023gaussianeditor}. Categorizing Gaussians into those on smooth surfaces and independent points is also feasible for geometric representation. By designing distinct initialization and densification strategies, better representation can be achieved~\cite{li2024geogaussian}. Additionally, some works design keyframe (or window) selection strategies based on inter-frame covisibility or geometric overlap ratio in temporal data for reconstructions~\cite{li2024sgs, sun2024high, yan2023gs, matsuki2023gaussian, keetha2023splatam, shaw2023swags}.

\textbf{Robust Optimization Strategy:} Introducing noise perturbations is a common method to enhance the robustness of training~\cite{yang2023deformable, tang2024lgm, yang2024gaussianobject,bao2024distractor}. Such perturbations can target camera poses, timesteps, and images, and can be regarded as a form of data augmentation to prevent overfitting. Additionally, some strategies mitigate catastrophic forgetting by avoiding continuous training from a single viewpoint~\cite{yugay2023gaussian, ha2024rgbd}.

\textbf{Distillation-based Strategy:} To compress model parameters, some distillation strategies use the original 3DGS as the teacher model and a low-dimensional SH 3DGS as the student model, introducing more pseudo views to enhance the performance of the low-dimensional SH~\cite{fan2023lightgaussian}.

{\textbf{Discussion}: Improving training strategies is an efficient way to optimize the training process of 3DGS and can enhance performance in many tasks. End-to-end training strategies, in particular, can improve performance while ensuring efficiency.}

\subsection{Adaptive Control}\label{Adaptive Control}

Adaptive Control of 3DGS is an important process for regulating the number of Gaussian primitives, including cloning, splitting, and pruning. In the following sections, we will summarize existing techniques from the perspectives of densification (cloning and splitting) and pruning.

\subsubsection{Densification} Densification is crucial, especially for detail reconstruction. we will analyze it from the perspectives of "Where to densify" and "How to densify". Additionally, we will discuss how to avoid excessive densification.

\textbf{Where to Densification:} Densification techniques focus on identifying positions requiring densification, governed by gradients in the original 3DGS and extendable to dynamic scene reconstruction~\cite{sun20243dgstream}. Regions with low opacity, silhouette, or high depth-rendered error, considered unreliable, guide densification to fill holes or improve 3D inconsistencies~\cite{sun2024high, yan2023gs, lei2024gaussnav, li2023spacetime, cheng2024gaussianpro, dai2024high}. Some approaches improve based on gradients by weighting the number of pixels covered by each Gaussian in different views to dynamically average view gradients, enhancing point cloud growth conditions~\cite{zhang2024pixel}. Additionally, SDF value, motion masks and neighbor distance are important criteria, with locations closer to the surface, motion regions and lower compactness being more prone to densification~\cite{yu2024gsdf, guedon2023sugar, chen2023text, li2024st}.

\textbf{How to Densification:} 
Numerous works have improved densification methods. Graph structures explore Gaussian relationships and define new Gaussians at edge centers based on proximity scores, mitigating sparse viewpoint impacts~\cite{zhu2023fsgs}. To prevent excessive Gaussian growth, the Candidate Pool Strategy stores pruned Gaussians for densification~\cite{he2024gvgen}. Additionally, work~\cite{feng2024new} introduces three conservation rules for visual consistency and employs integral tensor equations to model densification.

Excessive densification is also unnecessary, as it directly impacts the efficiency of 3DGS. In cases where two Gaussian functions are in close proximity, limiting their densification is a straightforward idea, where the distance between Gaussians can be measured by Gaussian Divergent Significance~\cite{feng2024fdgaussian} (GDS) or Kullback–Leibler divergence~\cite{hu2023gauhuman}, where $\bm{\mu}_1, \ \bm{\Sigma}_1, \ \bm{\mu}_2, \ \bm{\Sigma}_2$ belongs to two adjacent Gaussians.
\begin{align}
&\text{GDS}: \quad \| \bm{\mu}_1 - \bm{\mu}_2 \|^2 + \text{tr} \left( \bm{\Sigma}_1 + \bm{\Sigma}_2 - 2 (\bm{\Sigma}_1^{-1} \bm{\Sigma}_2 \bm{\Sigma}_1^{-1})^{1/2} \right), \notag \\
&\text{KL}: \text{\small \small $\displaystyle \frac{1}{2} \left( \text{tr}(\bm{\Sigma}_2^{-1} \bm{\Sigma}_1) + \ln \frac{\det \bm{\Sigma}_1}{\det \bm{\Sigma}_2} + (\bm{\mu}_2 - \bm{\mu}_1)^T \bm{\Sigma}_2^{-1} (\bm{\mu}_2 - \bm{\mu}_1) - 3 \right).$}\label{eq10}
\end{align}
And DeblurGS~\cite{oh2024deblurgs} incorporates a Gaussian Densification Annealing strategy to prevent the densification of inaccurate Gaussians during the early training stages at imprecise camera motion estimation. Furthermore, in some downstream tasks, densification is sometimes abandoned to prevent 3DGS from overfitting to each image, which could lead to incorrect geometric shapes~\cite{herau20243dgs, yan2023gs,matsuki2023gaussian,keetha2023splatam}.

\subsubsection{Pruning} Removing unimportant Gaussian primitives can ensure efficient representation. In the initial 3DGS framework, opacity was employed as the criterion for determining the significance of a Gaussian. Subsequent research has explored the incorporation of scale as a guiding factor or distractor masks for pruning~\cite{liu2023humangaussian, bao2024distractor}. However, these approaches primarily focus on individual Gaussian primitives, lacking a comprehensive consideration of the global representation. Therefore, subsequent derivative techniques are discussed.

\textbf{Importance scores:} The volume and hit count on training views, along with opacity, can be used to jointly determine the global significance score of a Gaussian primitive~\cite{fan2023lightgaussian}.
\vspace{-0.1cm}
\begin{equation}
     \text{GS}_j = \sum_{i=1}^{MHW} \mathbf{1}(G(\mathbf{X}_j), r_i) \cdot \sigma_j \cdot \gamma(\bm{\Sigma}_j), \label{eq11}
\end{equation}
\vspace{-0.2cm}

\noindent where $\gamma(\bm{\Sigma}_j)$ and $\mathbf{1}(G(\mathbf{X}_j), r_i)$ are volume and hit count, and $M$ represents the number of training views. Subsequently, Gaussians are ranked according to their global scores, and the ones with the lowest scores are pruned. Similar importance scores were improved in other works~\cite{niemeyer2024radsplat,fang2024mini}.

\textbf{Multi-view consistency:} Multi-view consistency is a key criterion for determining whether Gaussians need to be pruned. For example, \cite{matsuki2023gaussian} prunes newly added Gaussians that are not observed by three keyframes within a local keyframe window, while \cite{ji2024neds} prunes Gaussians that are invisible in all virtual views but visible in real views.

\textbf{Distance Metric:} Surface-aware methods often use distance to the surface~\cite{yan2023gs} and SDF values~\cite{yu2024gsdf} to prune Gaussian primitives far from the surface. The distance between Gaussians is also a key metric~\cite{li2024st}. GauHuman~\cite{hu2023gauhuman} aims to merge Gaussians with small scale and low KL divergence, as mentioned in Eq.~\ref{eq10}.

\textbf{Learnable control parameter:} Introducing a learnable mask based on scale and opacity to determine whether Gaussian primitives should be removed effectively prevents 3DGS from becoming overly dense~\cite{lee2023compact}.

{\textbf{Others:} CoR-GS~\cite{zhang2025cor} aims to leverage mismatched regions between two 3DGS models, trained in parallel under identical conditions, as guidance for pruning. }

{\textbf{Discussion}: Adaptive Control strategies play a pivotal role in enhancing rendering fidelity and computational efficiency. However, excessive densification or pruning can adversely affect both the efficiency and performance of 3D Gaussian Splatting. Therefore, it is crucial to examine and establish an optimal balance between these two strategic approaches.}

{\section{Other Technical Discussions}}\label{Other Technical Discussions}
{Several existing works have explored the augmentation of the original 3DGS pipeline through the integration of supplementary components, including post-processing modules and additional information as well as representations. These architectural enhancements typically yield improved performance metrics and facilitate optimization under ill-posed conditions.}

{\subsection{Post-Processing}} \label{Post-Processing}
{Post-processing strategies for pre-trained Gaussians are important, as they can improve the original efficiency and performance. Common post-processing often enhances Gaussian representations through various optimization strategies. This type of work has been discussed in Sec.~\ref{Training Strategy}.}

{\textbf{Representation Conversion:} Pre-trained 3DGS can be converted to Mesh using Poisson reconstruction~\cite{kazhdan2006poisson} on sampled 3D points~\cite{guedon2023sugar, dai2024high}. Similarly, GOF~\cite{yu2024gaussian} uses 3D bounding boxes to convert 3DGS to a Tetrahedral Grid, then extracts meshes using Binary Search of Level Set. Additionally, LGM~\cite{tang2024lgm} converts Pre-trained 3DGS to NeRF, then uses NeRF2Mesh~\cite{tang2023delicate} for Mesh conversion.}

{\textbf{Performance and Efficiency:} Some works enhance 3DGS performance in specific tasks through post-processing, such as multi-scale rendering. SA-GS~\cite{song2024sa} introduces a 2D scale-adaptive filter to dynamically adjust scales based on rendering frequency, enhancing anti-aliasing when zooming out. For efficiency, removing redundant Gaussian primitives from pre-trained 3DGS~\cite{jo2024identifying} or introducing a Gaussian caching mechanism~\cite{li2024ggrt} can improve rendering efficiency.}

% {\textbf{Discussion}: Post-processing strategies represent an effective yet underexplored aspect that can be readily integrated into any pre-trained 3DGS in a plug-and-play manner.}

{\subsection{Integration with Other Representations}}\label{Other Representations} 

{The convertible nature of 3D representations facilitates the integration of 3DGS with other representations, leveraging their advantages to improve the original 3DGS.}
{\subsubsection{Point Clouds}
Point clouds, as a 3D representation related to 3DGS, are often used to initialize positions. Converting point clouds to 3DGS can effectively fill holes~\cite{chung2023luciddreamer, ouyang2023text2immersion} or improve reconstruction details~\cite{lu2024large}, typically after high-precision reconstruction. Conversely, 3DGS can be converted into point clouds, voxelized into 3D voxels, and projected onto 2D BEV grids, which guide navigation tasks~\cite{lei2024gaussnav}. Additionally, anchor points in space can assist 3DGS. These methods use voxel centers as anchor points to represent the scene. Each anchor point comprises a local context feature, a scaling factor, and multiple learnable offsets. By decoding other attributes based on these offsets and features, the anchors transform into local neural Gaussians, which helps mitigate redundant expansion~\cite{lu2023scaffold, yang2024spec, yu2024gsdf}.}
{\subsubsection{Mesh}
Meshes have better geometric representation capabilities and can, to some extent, alleviate artifacts or blurry pixels caused by 3DGS~\cite{xiao2024bridging}. They are still the most widely used 3D representation in downstream tasks~\cite{tang2024lgm}. Much work has discussed converting 3DGS to Mesh, as mentioned in Sec.~\ref{Surface Representation}. Once converted, they can be optimized for better geometry and appearance~\cite{pang2023ash, tang2023dreamgaussian, ren2023dreamgaussian4d, palandra2024gsedit}. Jointly optimizing 3DGS and Mesh is also an optional strategy. 3DGS is suitable for constructing complex geometric structures, while Mesh can be used to reconstruct detailed color appearances on smooth surfaces. Combining the two can enhance reconstruction performance~\cite{xiao2024bridging} and large-scale deformation control~\cite{gao2024mesh}.
\subsubsection{Triplane}
Triplane, known for its compactness and efficient expressiveness~\cite{zou2023triplane}, is often used in generalization tasks. It consists of three orthogonal feature planes: $X$-$Y$, $Y$-$Z$, and $X$-$Z$. Features can be obtained by querying positions in the space, and subsequently decoded to predict Gaussian attributes~\cite{kocabas2023hugs, jiang2024brightdreamer, zou2023triplane, xu2024agg}. Recent works~\cite{wu20234d, yin20234dgen, shao2023control4d,li2024st} extend triplane to 4D space ($XYZ$-$T$) using multi-scale HexPlanes~\cite{cao2023hexplane} or 4D GaussianPlanes~\cite{shao2023control4d} to enhance 4DGS continuity in the spatiotemporal dimension.
\subsubsection{Grid}
Grid is also an efficient representation, as it can access grid corners and interpolate to obtain features or attributes at specific positions. Hash grid~\cite{muller2022instant}, a representative method, {can compress scenes and achieve a more compact and efficient 3DGS~\cite{lee2023compact, herau20243dgs, saroha2024gaussian, zuo2024fmgs, qian20233dgsavatar, he2024gvgen, chen2025hac,roessle2024l3dg}.} Furthermore, Self-Organizing Gaussian~\cite{morgenstern2023compact} maps unstructured 3D Gaussians onto a 2D grid to preserve local spatial relationships, where adjacent Gaussians have similar attribute values, reducing memory storage and maintaining continuity in 3D space. }

{Particularly, GaussianVolumes are also used for generalizable representations~\cite{he2024gvgen}, where a volume is composed of a fixed number of 3DGS. This maintains the efficiency of 3DGS and offers greater manipulability compared to triplane.
\subsubsection{Implicit Representation}
Implicit representations, benefiting from their representational capability, can be used to mitigate the condition difficulty and surface artifacts of 3DGS~\cite{bolanos2024charshadow}. Specifically, introducing NeRF to encode color and opacity can significantly enhance the representation's adjustability~\cite{malarzgaussian}. Moreover, by designing an SDF-to-opacity transformation function~\cite{lyu20243dgsr} or employing mutual geometry supervision~\cite{yu2024gsdf} to jointly optimize 3DGS and SDF representations, the surface reconstruction performance of 3DGS can be improved.}

{\textbf{Discussion}: Given the inherently unstructured characteristics of 3DGS, the incorporation of structured representations emerges as a viable prior, particularly advantageous for tasks such as human body reconstruction, facilitating both cross-representation transformations and geometric reconstruction, thereby enabling enhanced performance in downstream applications. However, due to certain limitations in representational characteristics, this approach may be accompanied by some degradation in rendering performance.}

% \subsubsection{GaussianVolumes}

% GaussianVolumes are also utilized for generalizable representations~\cite{he2024gvgen}, where a volume is composed of a fixed number of 3DGS. This representation maintains the efficiency of Gaussian representations while offering greater manipulability compared to the generalized triplane representation and alleviating the dependence on the accuracy of point cloud predictions.

% \textbf{Discussion}: Depending on different needs, various representations can be introduced. However, efficiently converting between different representations is important.

{\subsection{Guidance by Additional Prior}\label{Guidance by Additional Prior}
When dealing with under-determined problems, such as sparse view settings~\ref{Generalization}, introducing additional priors is a straightforward method to improve 3DGS performance.}

{\textbf{Pre-trained Models:} Introducing pre-trained models is an effective paradigm that can guide the optimization through the model's knowledge. Pre-trained monocular depth models and point cloud prediction models are a common type of priors, where the predicted depth values and positions can be used for the initialization and regularization~\cite{paliwal2025coherentgs, chung2023depth,zhu2023fsgs,ouyang2023text2immersion,ji2024neds,chung2023luciddreamer,swann2024touch}. Pre-trained 2D image (or 3D and video) generative models are also important in some AIGC-related tasks. They can be used not only for optimization in combination with SDS Loss~\cite{chen2023text,li2024controllable,ling2023align} but also for directly generating (or editing) images for training~\cite{chung2023luciddreamer,melas20243d,ouyang2023text2immersion,pan2024fast,liu2025deceptive}. Similarly, some works introduce pre-trained image inpainting networks to alleviate difficulties caused by occlusion as well as overlap~\cite{zhang2023repaint123,chung2023luciddreamer,ouyang2023text2immersion,chen2023gaussianeditor,huang2023point} or super-resolution models for a high level of detail~\cite{ouyang2023text2immersion,shao2023control4d} during the generation process. Additionally, pre-trained ControlNet~\cite{zhang2023adding} or Large Language Models can also be used to guide 3D generation. The former can enhance geometric consistency under depth guidance~\cite{wu2024gaussctrl,zhang2023repaint123, zhou2024gala3d}, while the latter can predict layout maps to guide spatial relationships in multi-object 3D generation scenarios~\cite{zhou2024gala3d}. Notably, certain pre-trained models can endow 3DGS with additional capabilities, such as semantic understanding models, as discussed in Sec.~\ref{Semantic Understanding} and spatial understanding models~\cite{ji2024neds}.}

{\textbf{More Sensors:} Due to the 3D-agnostic nature of 2D images, reconstructing 3DGS can be challenging, especially in large-scale reconstructions such as SLAM and autonomous driving. Therefore, incorporating additional sensors for 3D depth information, including depth sensors~\cite{zhu2024semgauss, ha2024rgbd, ji2024neds, sun2024high, yugay2023gaussian}, audio~\cite{li2025talkinggaussian,cho2024gaussiantalker,yu2024gaussiantalker}, LiDAR~\cite{chen2025g3r,zhou2023drivinggaussian, jiang20243dgs, herau20243dgs,zhao2024tclc}, and optical tactile sensors~\cite{swann2024touch}, has the potential to alleviate this issue.}

{\textbf{Task-specific Priors:} Some reconstruction tasks, such as human reconstruction, target subjects with certain common characteristics. These characteristics, such as template models and Linear Blend Skinning, can be extracted as priors to guide the reconstruction of similar targets. In the reconstruction, animation, and generation of non-rigid objects, many works utilize SMPL \cite{Loper2015smpl} and SMAL \cite{zuffi20173d} to provide strong priors for representing the motion and deformation of non-rigid objects like humans \cite{kocabas2023hugs,moreau2023human,liu2023humangaussian,hu2023gauhuman,qian20233dgsavatar} and animals \cite{lei2023gart,zhang2024bags}. Subsequently, based on the SMPL template, Shell Maps \cite{porumbescu2005shell} and template meshes are also introduced in combination with 3DGS to address issues of low efficiency in 3DGAN \cite{abdal2023gaussian,kirschstein2024gghead} and unclear geometry \cite{li2023animatable,pang2023ash}. Similarly, in head and face reconstruction and animation tasks, some works \cite{qian2023gaussianavatars,xu2023gaussian} also use the FLAME model \cite{siggraphAsia2017flame} as a prior. Linear Blend Skinning \cite{sumner2007embedded} is also employed as prior knowledge to assist in the prediction of 3DGS motion \cite{xu2023gaussian,huang2023sc}. Additionally, in 3D urban scene reconstruction tasks, HUGS \cite{zhou2024hugs} introduces the Unicycle Model to model the motion of vehicles, thereby making the motion modeling of moving objects smoother.}

{\textbf{Discussion}: Accessible auxiliary information has the capability to enhance the performance of 3DGS across numerous tasks, serving as prior knowledge to facilitate spatial comprehension, particularly in inherently ill-posed problems. Although certain priors or sensors may lead to increased computational overhead and costs, they have the capability to significantly enhance the representational capacity of 3DGS
.}

\begin{table*}
    \centering
    \caption{The Relationships among Challenges, Tasks, and Technological Improvement, where the first column represents the core challenges, while the second and third columns denote the related downstream tasks and the corresponding technological advancements associated with these challenges, respectively}
    \vspace{-.2cm}
    \begin{tabular}{
        >{\columncolor[gray]{0.9}}p{2cm}|
        >{\columncolor[gray]{0.9}}p{9cm}|
        >{\columncolor[gray]{0.9}}p{6cm}
    }
        \toprule
         Challenges & Major Tasks & Major Technological Improvements  \\
        \midrule
        \rowcolor[HTML]{FFFFE0} % Light Yellow
        \textit{Suboptimal Data \quad \quad Challenges}~\ref{Suboptimal Data Challenges}     & \textbf{Limited Number}: Sparse Views ~\ref{Sparse Views}, Autonomous Driving Reconstruction ~\ref{Autonomous Driving Scene Reconstruction}, Dynamic 3DGS~\ref{4DGS} (Monocular Video Part~\ref{Monocular Video}), AIGC~\ref{AIGC} and Editable 3DGS~\ref{Editable}. \textbf{Limited Quality}: Slam~\ref{SLAM} and Reconstruction under blurred images (Sec.\ref{Photorealism}) or without poses~\cite{fan2024instantsplat,sun2023icomma}   & \textbf{Limited Number}: Initialization~\ref{Initialization}, Regularization~\ref{Regularization}, Adaptive Control~\ref{Adaptive Control} Training Strategies~\ref{Training Strategy}, and Guidance by Additional Prior~\ref{Guidance by Additional Prior}.  \textbf{Limited Quality}: Training Strategies~\ref{Training Strategy} and Integration with Other Representations (Sec.~\ref{Other Representations})   \\
        \midrule
        \rowcolor[HTML]{FFFFE0} % Light Yellow
       \textit{Generalization Challenges}~\ref{Generalization Challenges}    & Generalization~\ref{Generalization}, and Generalization-related tasks in the Human Reconstruction~\ref{Human Reconstruction} and AIGC~\ref{AIGC}   & Initialization~\ref{Initialization}, Adaptive Control~\ref{Adaptive Control}, and Integration with Other Representations~\ref{Other Representations}      \\
        \midrule
        \rowcolor[HTML]{FFFFE0} % Light Yellow
        \textit{Physics Challenges}~\ref{Physics Challenges}    & \textbf{Physical Motion}: Dynamic 3DGS~\ref{4DGS}, Physics Simulation~\ref{Physics Simulation}, Animation~\ref{Animation}, Dynamic Human~\ref{Human Reconstruction} and Autonomous Driving Reconstruction ~\ref{Autonomous Driving Scene Reconstruction}. \textbf{Physical Rendering}: Photorealism~\ref{Photorealism} and Physics Simulation~\ref{Physics Simulation}.  & \textbf{Physical Motion and Rendering}: Attribute Expansion~\ref{Attribute Expansion}, Regularization Strategy~\ref{Regularization}, and Guidance by Additional Prior~\ref{Guidance by Additional Prior}.   \\
        \midrule
        \rowcolor[HTML]{FFFFE0} % Light Yellow
       \textit{Realness and Efficiency Challenges} \ref{Realness and Efficiency Challenges}    & \textbf{Realness}: Photorealism~\ref{Photorealism}, Surface Reconstruction~\ref{Surface Representation}, Semantic Understanding~\ref{Semantic Understanding}, some AIGC-related~\ref{AIGC} and Autonomous Driving~\ref{Autonomous Driving} works. \textbf{Efficiency}: Efficiency~\ref{Efficiency}, some works in Autonomous Driving~\ref{Autonomous Driving} and Semantic Understanding~\ref{Semantic Understanding} & \textbf{Realness}: Most of the technologies in~\ref{Technical Improvements}. \textbf{Efficiency}: Attribute Expansion~\ref{Attribute Expansion}, Post-Processing~\ref{Post-Processing}, Adaptive Control~\ref{Adaptive Control} and Splatting~\ref{Splatting}.  \\
        \bottomrule
    \end{tabular}
    \label{tab2}
    \vspace{-0.5cm}
\end{table*}

{\section{Challenges and Opportunities}}\label{Challenges and Opportunities}
{The preceding discussion indicates that various 3DGS-related tasks share similar technical approaches, which stems from common challenges across different tasks. To provide readers with a deeper understanding of this phenomenon, this section examines the commonalities among different tasks, summarizes four core challenges as well as their corresponding technical solutions, and outlines future opportunities.}

{\subsection{Interrelationships}\label{Interrelationships}
We have extensively discussed various 3DGS-related tasks in Sec.\ref{Optimization}, Sec.\ref{Applications}, and Sec.~\ref{Extensions}, revealing common challenges and techniques across these tasks. As illustrated in Tab.\ref{tab2}, we categorize existing tasks according to four core challenges, demonstrating that solutions from different tasks can be mutually instructive. Furthermore, there are some interrelationships between different tasks that have not been mentioned. For instance, Surface Reconstruction techniques (Sec.\ref{Surface Representation}) are often referenced in the context of Editable 3DGS (Sec.\ref{Editable}), etc. We anticipate that this analysis will offer valuable insights for future research endeavors in related tasks.}

\subsection{Suboptimal Data}\label{Suboptimal Data Challenges}

\textbf{Challenges.} In real-world scenarios, collecting large volumes of high-quality training data is often impractical. Without access to 3D data and sufficient multi-view images, relying on limited 2D image supervision is insufficient for accurate 3DGS reconstruction. For example, inferring the back appearance from only a frontal image is highly challenging. Additionally, data quality is critical, as accurate poses and clear images directly influence reconstruction performance.

{\textbf{Opportunities.} An ideal 3DGS training process requires sufficient high-quality data, but this is often excessively challenging in practical applications. Although introducing priors can mitigate this problem to some extent, optimizing a large number of Gaussians under underconstrained conditions remains difficult. A potential solution is to reduce the number of Gaussian primitives based on their uncertainty while enhancing the representational capacity of individual primitives~\cite{hamdi2024ges}. This involves finding a trade-off between the number of Gaussians and rendering performance, thereby improving the efficiency of utilizing sparse samples. Then, poor-quality data should also be taken into consideration. Unconstrained in-the-wild images are a typical case, encompassing transient occlusions and dynamic appearance changes, such as varying sky, weather, and lighting, which have been extensively discussed in NeRF~\cite{martin2021nerf, chen2022hallucinated, yang2023cross}. To enhance efficiency, existing works have addressed this issue in the context of 3DGS~\cite{dahmani2024swag, zhang2024gaussian}, attempting to model appearance changes and handle transient objects. However, their performance struggles, especially in scenes with complex lighting changes and frequent occlusions. Thanks to the explicit representation characteristics of 3DGS, decoupling geometric representations and introducing geometric consistency constraints across different scenes is a promising approach to mitigate instability during the training process.}
% Such problems are discussed across multiple tasks, such as Sparse Views Setting (Sec.\ref{Generalization}), Autonomous Driving (Sec.\ref{Autonomous Driving}), Dynamic 3DGS (Sec.\ref{4DGS}) (Monocular Video), AIGC (Sec.\ref{AIGC}), and Editable 3DGS (Sec.\ref{Editable}). Numerous works have discussed how to improve initialization (Sec.\ref{Initialization}), regularization (Sec.\ref{Regularization}), adaptive control (Sec.\ref{Adaptive Control}), and training strategies (Sec.\ref{Training Strategy}), or introduce additional priors (Sec.\ref{Guidance by Additional Prior}) in the context of sparse or missing multi-view training images. Additionally, reconstruction with missing accurate poses can also be considered an underdetermined problem. This is discussed in the SLAM literature (Sec.\ref{SLAM}) and reconstruction under blurred images (Sec.\ref{Photorealism}) or without poses~\cite{fan2024instantsplat,sun2023icomma}, where new training strategies (Sec.~\ref{Training Strategy}) and other representations (Sec.~\ref{Other Representations}) are often introduced to mitigate it.
\subsection{Generalization}\label{Generalization Challenges}
\textbf{Challenges.} Despite the improved training efficiency compared to NeRF, the scene-specific training paradigm remains a major bottleneck for the application of 3DGS. It is hard to imagine having to train for each target or scene individually, especially in multi-target and scene reconstruction (generation).

{\textbf{Opportunities.} Although existing generalization-related works can directly obtain scene representations through forward inference, their performance is often unsatisfactory and limited by the type of scene~\cite{chen2024mvsplat,jiang2024brightdreamer,szymanowicz2023splatter,zou2023triplane}. We hypothesize that this is due to the difficulty of feedforward networks in performing the adaptive control of 3DGS, as also mentioned in~\cite{charatan2023pixelsplat}. In future research, designing a reference-feature-based feedforward adaptive control strategy is a potential solution, which can predict the positions requiring adaptive control through reference features and be plug-and-play into existing generalization-related works. Additionally, existing generalization-related works rely on accurate poses, which are often difficult to obtain in practical applications~\cite{fu2023colmap,sun2023icomma,cai2024gs}. Therefore, discussing generalizable 3DGS under pose-missing conditions is also promising~\cite{li2024ggrt}.}
\subsection{Physics Reconstruction and Rendering}\label{Physics Challenges}
\textbf{Challenges.} Traditional 3DGS only considers static rendering and neglects the laws of physical motion, which are important in simulations~\cite{xie2023physgaussian}. Additionally, Physically-based rendering is a significant step towards applying 3DGS to simulate the physical world and achieve more realistic effects.

{\textbf{Opportunities.} Ensuring that the 3DGS's motion adheres to physical laws is essential for unifying simulation and rendering~\cite{xie2023physgaussian}. Although rigidity-related regularization have been introduced, as described in Sec.~\ref{3D Regularization}, most existing works focus on animating 3DGS while neglecting the physical attributes of the Gaussian primitives themselves (Sec.~\ref{4DGS}). Some pioneering works attempt to introduce velocity attributes~\cite{guo2024motion} and Newtonian dynamics rules~\cite{xie2023physgaussian}, but this is not sufficient to fully describe the physical motion of 3DGS in space. A potential solution is to introduce more physical attributes in Gaussian primitives, such as material~\cite{liang2024gs}, acceleration, and force distribution, which can be regularized by priors from certain simulation tools and physics knowledge. Physically-based rendering is also a direction worth attention, as it enables 3DGS to handle relighting and material editing, producing outstanding inverse rendering results~\cite{gao2023relightable}. Future works can explore decoupling geometry and appearance in 3DGS, conducting research from the perspectives of normal reconstruction and the modeling of illumination and materials~\cite{liang2024gs, saito2024relightable, bolanos2024charshadow}.}

\subsection{Realness and Efficiency}\label{Realness and Efficiency Challenges}
\textbf{Challenges.} Realness and efficiency challenges are fundamental issues. They are investigated in various works and have been discussed in Sec.~\ref{Optimization}. In this part, we discuss downstream tasks and techniques optimized for performance and efficiency.

{\textbf{Opportunities.} The difficulty in reconstructing clear surfaces has always been a significant challenge affecting rendering realism. As discussed in Sec.~\ref{Surface Representation}, some works have addressed it by attempting to represent surfaces with planar Gaussians. However, this can result in a decline in rendering performance, possibly due to the reduced representational capacity of planar Gaussian primitives or the training instability. Therefore, designing Gaussian primitives better suited for surface representation and introducing a multi-stage training paradigm along with regularization are potential solutions. Storage efficiency is one of the critical bottlenecks of 3DGS. Existing works focus on introducing VQ techniques and compressing SH parameters, as discussed in Sec.~\ref{Storage Efficiency}. However, such approaches inevitably affect performance. Therefore, exploring how to design more efficient representations based on 3DGS is a potential way to enhance efficiency~\cite{lu2023scaffold, hamdi2024ges} while maintaining performance.}

\subsection{Other Technical Challenges and Opportunities}
% \textbf{Interrelationships.} There are some relationships between different domains that have not been mentioned. For instance, Surface Reconstruction techniques (Sec.\ref{Surface Representation}) are often referenced in the context of Editable 3DGS (Sec.\ref{Editable}), etc.

{An increasing number of research and engineering projects have found that initialization is important in 3DGS. Traditional SfM is not suitable for constrained scenarios, such as sparse view settings, AIGC, and low-light reconstruction. Therefore, more robust initialization methods should be designed to replace original initialization in these scenarios. And
Splatting also plays an important role in 3DGS, but it is rarely mentioned in existing works~\cite{franke2024trips, huang2024gs++}. Designing efficient parallel splatting strategies on pretrained 3DGS has the potential to impact rendering performance and efficiency. }

\section{Conclusion}\label{Conclusion}
The burgeoning interest in the field of 3D Gaussian Splatting (3DGS) has precipitated the emergence of a myriad of related downstream tasks and technologies, thereby contributing to increasing complexity and confusion within the domain. This complexity manifests in various forms, including similar motivations across different works, the incorporation of similar technologies across disparate tasks, and the nuanced differences and interconnections between various technologies. This survey systematically categorizes existing works based on their motivations and critically discusses associated technologies. Our objective is to elucidate common challenges across tasks and technologies, providing a coherent framework for understanding this evolving field. Additionally, we highlight prospective research avenues to inspire continued innovation in 3DGS.

% \section*{Acknowledgments}
% This work was supported in part by the National Natural Science Foundation of China under Grant 62276128, Grant 62192783, and Grant 62106100; in part by the Jiangsu Natural Science Foundation under Grant BK20221441; in part by the Young Elite Scientists Sponsorship Program by CAST under Grant 2023QNRC001; in part by the Collaborative Innovation Center of Novel Software Technology and Industrialization. 

\bibliographystyle{IEEEtran}
\bibliography{egbib}

\begin{thebibliography}{100}
\providecommand{\url}[1]{#1}
\csname url@samestyle\endcsname
\providecommand{\newblock}{\relax}
\providecommand{\bibinfo}[2]{#2}
\providecommand{\BIBentrySTDinterwordspacing}{\spaceskip=0pt\relax}
\providecommand{\BIBentryALTinterwordstretchfactor}{4}
\providecommand{\BIBentryALTinterwordspacing}{\spaceskip=\fontdimen2\font plus
\BIBentryALTinterwordstretchfactor\fontdimen3\font minus \fontdimen4\font\relax}
\providecommand{\BIBforeignlanguage}[2]{{%
\expandafter\ifx\csname l@#1\endcsname\relax
\typeout{** WARNING: IEEEtran.bst: No hyphenation pattern has been}%
\typeout{** loaded for the language `#1'. Using the pattern for}%
\typeout{** the default language instead.}%
\else
\language=\csname l@#1\endcsname
\fi
#2}}
\providecommand{\BIBdecl}{\relax}
\BIBdecl

\bibitem{mildenhall2021nerf}
B.~Mildenhall, P.~P. Srinivasan, M.~Tancik, J.~T. Barron, R.~Ramamoorthi, and R.~Ng, ``Nerf: Representing scenes as neural radiance fields for view synthesis,'' \emph{Commun. ACM}, vol.~65, no.~1, pp. 99--106, 2021.

\bibitem{chen2024survey}
G.~Chen and W.~Wang, ``A survey on 3d gaussian splatting,'' \emph{arXiv:2401.03890}, 2024.

\bibitem{wu2024recent}
T.~Wu, Y.-J. Yuan, L.-X. Zhang, J.~Yang, Y.-P. Cao, L.-Q. Yan, and L.~Gao, ``Recent advances in 3d gaussian splatting,'' \emph{Comput. Vis. Media}, pp. 1--30, 2024.

\bibitem{fei20243d}
B.~Fei, J.~Xu, R.~Zhang, Q.~Zhou, W.~Yang, and Y.~He, ``3d gaussian splatting as new era: A survey,'' \emph{IEEE Trans. Vis. Comput. Graph.}, 2024.

\bibitem{liao2024ov}
G.~Liao, K.~Zhou, Z.~Bao, K.~Liu, and Q.~Li, ``Ov-nerf: Open-vocabulary neural radiance fields with vision and language foundation models for 3d semantic understanding,'' \emph{IEEE Trans. Circuits Syst. Video Technol.}, 2024.

\bibitem{zhang2021progressive}
P.~Zhang, X.~Wang, L.~Ma, S.~Wang, S.~Kwong, and J.~Jiang, ``Progressive point cloud upsampling via differentiable rendering,'' \emph{IEEE Trans. Circuits Syst. Video Technol.}, vol.~31, no.~12, pp. 4673--4685, 2021.

\bibitem{kerbl20233d}
B.~Kerbl, G.~Kopanas, T.~Leimk{\"u}hler, and G.~Drettakis, ``3d gaussian splatting for real-time radiance field rendering,'' \emph{ACM Trans. Graph.}, vol.~42, no.~4, pp. 1--14, 2023.

\bibitem{zwicker2001ewa}
M.~Zwicker, H.~Pfister, J.~Van~Baar, and M.~Gross, ``Ewa volume splatting,'' in \emph{Proc. IEEE Vis.}\hskip 1em plus 0.5em minus 0.4em\relax IEEE, 2001, pp. 29--538.

\bibitem{ding2024ray}
J.~Ding, Y.~He, B.~Yuan, Z.~Yuan, P.~Zhou, J.~Yu, and X.~Lou, ``Ray reordering for hardware-accelerated neural volume rendering,'' \emph{IEEE Trans. Circuits Syst. Video Technol.}, 2024.

\bibitem{girish2023eagles}
S.~Girish, K.~Gupta, and A.~Shrivastava, ``Eagles: Efficient accelerated 3d gaussians with lightweight encodings,'' \emph{arXiv:2312.04564}, 2023.

\bibitem{navaneet2023compact3d}
K.~Navaneet, K.~P. Meibodi, S.~A. Koohpayegani, and H.~Pirsiavash, ``Compact3d: Compressing gaussian splat radiance field models with vector quantization,'' \emph{arXiv:2311.18159}, 2023.

\bibitem{niedermayr2023compressed}
S.~Niedermayr, J.~Stumpfegger, and R.~Westermann, ``Compressed 3d gaussian splatting for accelerated novel view synthesis,'' \emph{arXiv:2401.02436}, 2023.

\bibitem{equitz1989new}
W.~H. Equitz, ``A new vector quantization clustering algorithm,'' \emph{IEEE Trans. Acoust. Speech Signal Process.}, vol.~37, no.~10, pp. 1568--1575, 1989.

\bibitem{lee2023compact}
J.~C. Lee, D.~Rho, X.~Sun, J.~H. Ko, and E.~Park, ``Compact 3d gaussian representation for radiance field,'' \emph{arXiv:2311.13681}, 2023.

\bibitem{fan2023lightgaussian}
Z.~Fan, K.~Wang, K.~Wen, Z.~Zhu, D.~Xu, and Z.~Wang, ``Lightgaussian: Unbounded 3d gaussian compression with 15x reduction and 200+ fps,'' \emph{arXiv:2311.17245}, 2023.

\bibitem{lu2023scaffold}
T.~Lu, M.~Yu, L.~Xu, Y.~Xiangli, L.~Wang, D.~Lin, and B.~Dai, ``Scaffold-gs: Structured 3d gaussians for view-adaptive rendering,'' \emph{arXiv:2312.00109}, 2023.

\bibitem{hamdi2024ges}
A.~Hamdi, L.~Melas-Kyriazi, J.~Mai, G.~Qian, R.~Liu, C.~Vondrick, B.~Ghanem, and A.~Vedaldi, ``Ges: Generalized exponential splatting for efficient radiance field rendering,'' in \emph{Proc. IEEE/CVF Conf. Comput. Vis. Pattern Recognit.}, 2024, pp. 19\,812--19\,822.

\bibitem{morgenstern2023compact}
W.~Morgenstern, F.~Barthel, A.~Hilsmann, and P.~Eisert, ``Compact 3d scene representation via self-organizing gaussian grids,'' \emph{arXiv:2312.13299}, 2023.

\bibitem{chen2025hac}
Y.~Chen, Q.~Wu, W.~Lin, M.~Harandi, and J.~Cai, ``Hac: Hash-grid assisted context for 3d gaussian splatting compression,'' in \emph{Eur. Conf. Comput. Vis.}\hskip 1em plus 0.5em minus 0.4em\relax Springer, 2025, pp. 422--438.

\bibitem{barron2022mip}
J.~T. Barron, B.~Mildenhall, D.~Verbin, P.~P. Srinivasan, and P.~Hedman, ``Mip-nerf 360: Unbounded anti-aliased neural radiance fields,'' in \emph{Proc. IEEE/CVF Conf. Comput. Vis. Pattern Recognit.}, 2022, pp. 5470--5479.

\bibitem{lee2024compact}
J.~C. Lee, D.~Rho, X.~Sun, J.~H. Ko, and E.~Park, ``Compact 3d gaussian representation for radiance field,'' in \emph{Proc. IEEE/CVF Conf. Comput. Vis. Pattern Recognit.}, 2024, pp. 21\,719--21\,728.

\bibitem{durvasula2023distwar}
S.~Durvasula, A.~Zhao, F.~Chen, R.~Liang, P.~K. Sanjaya, and N.~Vijaykumar, ``Distwar: Fast differentiable rendering on raster-based rendering pipelines,'' \emph{arXiv:2401.05345}, 2023.

\bibitem{jo2024identifying}
J.~Jo, H.~Kim, and J.~Park, ``Identifying unnecessary 3d gaussians using clustering for fast rendering of 3d gaussian splatting,'' \emph{arXiv:2402.13827}, 2024.

\bibitem{lee2024gscore}
J.~Lee, S.~Lee, J.~Lee, J.~Park, and J.~Sim, ``Gscore: Efficient radiance field rendering via architectural support for 3d gaussian splatting,'' in \emph{Proc. 29th ACM Int. Conf. Archit. Support Program. Lang. Oper. Syst., Volume 3}, 2024, pp. 497--511.

\bibitem{verbin2022ref}
D.~Verbin, P.~Hedman, B.~Mildenhall, T.~Zickler, J.~T. Barron, and P.~P. Srinivasan, ``Ref-nerf: Structured view-dependent appearance for neural radiance fields,'' in \emph{Proc. IEEE/CVF Conf. Comput. Vis. Pattern Recognit.}\hskip 1em plus 0.5em minus 0.4em\relax IEEE, 2022, pp. 5481--5490.

\bibitem{cheng2024gaussianpro}
K.~Cheng, X.~Long, K.~Yang, Y.~Yao, W.~Yin, Y.~Ma, W.~Wang, and X.~Chen, ``Gaussianpro: 3d gaussian splatting with progressive propagation,'' \emph{arXiv:2402.14650}, 2024.

\bibitem{zhang2024fregs}
J.~Zhang, F.~Zhan, M.~Xu, S.~Lu, and E.~Xing, ``Fregs: 3d gaussian splatting with progressive frequency regularization,'' \emph{arXiv:2403.06908}, 2024.

\bibitem{radl2024stopthepop}
L.~Radl, M.~Steiner, M.~Parger, A.~Weinrauch, B.~Kerbl, and M.~Steinberger, ``Stopthepop: Sorted gaussian splatting for view-consistent real-time rendering,'' \emph{ACM Trans. Graph.}, vol.~43, no.~4, pp. 1--17, 2024.

\bibitem{diolatzis2024n}
S.~Diolatzis, T.~Zirr, A.~Kuznetsov, G.~Kopanas, and A.~Kaplanyan, ``N-dimensional gaussians for fitting of high dimensional functions,'' in \emph{ACM SIGGRAPH Conf. Papers}, 2024, pp. 1--11.

\bibitem{yan2023multi}
Z.~Yan, W.~F. Low, Y.~Chen, and G.~H. Lee, ``Multi-scale 3d gaussian splatting for anti-aliased rendering,'' \emph{arXiv:2311.17089}, 2023.

\bibitem{yu2024mip}
Z.~Yu, A.~Chen, B.~Huang, T.~Sattler, and A.~Geiger, ``Mip-splatting: Alias-free 3d gaussian splatting,'' in \emph{Proc. IEEE/CVF Conf. Comput. Vis. Pattern Recognit.}, 2024, pp. 19\,447--19\,456.

\bibitem{song2024sa}
X.~Song, J.~Zheng, S.~Yuan, H.-a. Gao, J.~Zhao, X.~He, W.~Gu, and H.~Zhao, ``Sa-gs: Scale-adaptive gaussian splatting for training-free anti-aliasing,'' \emph{arXiv:2403.19615}, 2024.

\bibitem{liang2025analytic}
Z.~Liang, Q.~Zhang, W.~Hu, L.~Zhu, Y.~Feng, and K.~Jia, ``Analytic-splatting: Anti-aliased 3d gaussian splatting via analytic integration,'' in \emph{Eur. Conf. Comput. Vis.}\hskip 1em plus 0.5em minus 0.4em\relax Springer, 2025, pp. 281--297.

\bibitem{jiang2024gaussianshader}
Y.~Jiang, J.~Tu, Y.~Liu, X.~Gao, X.~Long, W.~Wang, and Y.~Ma, ``Gaussianshader: 3d gaussian splatting with shading functions for reflective surfaces,'' in \emph{Proc. IEEE/CVF Conf. Comput. Vis. Pattern Recognit.}, 2024, pp. 5322--5332.

\bibitem{meng2024mirror}
J.~Meng, H.~Li, Y.~Wu, Q.~Gao, S.~Yang, J.~Zhang, and S.~Ma, ``Mirror-3dgs: Incorporating mirror reflections into 3d gaussian splatting,'' \emph{arXiv:2404.01168}, 2024.

\bibitem{yang2024spec}
Z.~Yang, X.~Gao, Y.~Sun, Y.~Huang, X.~Lyu, W.~Zhou, S.~Jiao, X.~Qi, and X.~Jin, ``Spec-gaussian: Anisotropic view-dependent appearance for 3d gaussian splatting,'' \emph{arXiv:2402.15870}, 2024.

\bibitem{ye20243d}
K.~Ye, Q.~Hou, and K.~Zhou, ``3d gaussian splatting with deferred reflection,'' in \emph{ACM SIGGRAPH Conf. Papers}, 2024, pp. 1--10.

\bibitem{10.1145/3680528.3687659}
\BIBentryALTinterwordspacing
X.~Wu, J.~Xu, C.~Wang, Y.~Peng, Q.~Huang, J.~Tompkin, and W.~Xu, ``Local gaussian density mixtures for unstructured lumigraph rendering,'' in \emph{ACM SIGGRAPH Conf. Papers}, ser. SA '24.\hskip 1em plus 0.5em minus 0.4em\relax New York, NY, USA: Association for Computing Machinery, 2024. [Online]. Available: \url{https://doi.org/10.1145/3680528.3687659}
\BIBentrySTDinterwordspacing

\bibitem{liu2025mirrorgaussian}
J.~Liu, X.~Tang, F.~Cheng, R.~Yang, Z.~Li, J.~Liu, Y.~Huang, J.~Lin, S.~Liu, X.~Wu \emph{et~al.}, ``Mirrorgaussian: Reflecting 3d gaussians for reconstructing mirror reflections,'' in \emph{Eur. Conf. Comput. Vis.}\hskip 1em plus 0.5em minus 0.4em\relax Springer, 2025, pp. 377--393.

\bibitem{oh2024deblurgs}
J.~Oh, J.~Chung, D.~Lee, and K.~M. Lee, ``Deblurgs: Gaussian splatting for camera motion blur,'' \emph{arXiv:2404.11358}, 2024.

\bibitem{lee2025deblurring}
B.~Lee, H.~Lee, X.~Sun, U.~Ali, and E.~Park, ``Deblurring 3d gaussian splatting,'' in \emph{Eur. Conf. Comput. Vis.}\hskip 1em plus 0.5em minus 0.4em\relax Springer, 2025, pp. 127--143.

\bibitem{seiskari2025gaussian}
O.~Seiskari, J.~Ylilammi, V.~Kaatrasalo, P.~Rantalankila, M.~Turkulainen, J.~Kannala, E.~Rahtu, and A.~Solin, ``Gaussian splatting on the move: Blur and rolling shutter compensation for natural camera motion,'' in \emph{Eur. Conf. Comput. Vis.}\hskip 1em plus 0.5em minus 0.4em\relax Springer, 2025, pp. 160--177.

\bibitem{peng2025bags}
C.~Peng, Y.~Tang, Y.~Zhou, N.~Wang, X.~Liu, D.~Li, and R.~Chellappa, ``Bags: Blur agnostic gaussian splatting through multi-scale kernel modeling,'' in \emph{Eur. Conf. Comput. Vis.}\hskip 1em plus 0.5em minus 0.4em\relax Springer, 2025, pp. 293--310.

\bibitem{walter2007microfacet}
B.~Walter, S.~R. Marschner, H.~Li, and K.~E. Torrance, ``Microfacet models for refraction through rough surfaces,'' in \emph{Proc. 18th Eurographics Conf. Rendering Tech.}, 2007, pp. 195--206.

\bibitem{szymanowicz2023splatter}
S.~Szymanowicz, C.~Rupprecht, and A.~Vedaldi, ``Splatter image: Ultra-fast single-view 3d reconstruction,'' \emph{arXiv:2312.13150}, 2023.

\bibitem{charatan2023pixelsplat}
D.~Charatan, S.~Li, A.~Tagliasacchi, and V.~Sitzmann, ``Pixelsplat: 3d gaussian splats from image pairs for scalable generalizable 3d reconstruction,'' \emph{arXiv:2312.12337}, 2023.

\bibitem{chen2024mvsplat}
Y.~Chen, H.~Xu, C.~Zheng, B.~Zhuang, M.~Pollefeys, A.~Geiger, T.-J. Cham, and J.~Cai, ``Mvsplat: Efficient 3d gaussian splatting from sparse multi-view images,'' \emph{arXiv:2403.14627}, 2024.

\bibitem{wang2024freesplat}
Y.~Wang, T.~Huang, H.~Chen, and G.~H. Lee, ``Freesplat: Generalizable 3d gaussian splatting towards free-view synthesis of indoor scenes,'' \emph{arXiv:2405.17958}, 2024.

\bibitem{bao2024distractor}
Y.~Bao, J.~Liao, J.~Huo, and Y.~Gao, ``Distractor-free generalizable 3d gaussian splatting,'' \emph{arXiv:2411.17605}, 2024.

\bibitem{chen2025g3r}
Y.~Chen, J.~Wang, Z.~Yang, S.~Manivasagam, and R.~Urtasun, ``G3r: Gradient guided generalizable reconstruction,'' in \emph{Eur. Conf. Comput. Vis.}\hskip 1em plus 0.5em minus 0.4em\relax Springer, 2025, pp. 305--323.

\bibitem{zou2023triplane}
Z.-X. Zou, Z.~Yu, Y.-C. Guo, Y.~Li, D.~Liang, Y.-P. Cao, and S.-H. Zhang, ``Triplane meets gaussian splatting: Fast and generalizable single-view 3d reconstruction with transformers,'' \emph{arXiv:2312.09147}, 2023.

\bibitem{xu2024agg}
D.~Xu, Y.~Yuan, M.~Mardani, S.~Liu, J.~Song, Z.~Wang, and A.~Vahdat, ``Agg: Amortized generative 3d gaussians for single image to 3d,'' \emph{arXiv:2401.04099}, 2024.

\bibitem{gslrm2024}
K.~Zhang, S.~Bi, H.~Tan, Y.~Xiangli, N.~Zhao, K.~Sunkavalli, and Z.~Xu, ``Gs-lrm: Large reconstruction model for 3d gaussian splatting,'' \emph{Eur. Conf. Comput. Vis.}, 2024.

\bibitem{xu2024grm}
Y.~Xu, Z.~Shi, W.~Yifan, H.~Chen, C.~Yang, S.~Peng, Y.~Shen, and G.~Wetzstein, ``Grm: Large gaussian reconstruction model for efficient 3d reconstruction and generation,'' \emph{arXiv:2403.14621}, 2024.

\bibitem{guo2024depth}
S.~Guo, Q.~Wang, Y.~Gao, R.~Xie, L.~Li, F.~Zhu, and L.~Song, ``Depth-guided robust point cloud fusion nerf for sparse input views,'' \emph{IEEE Trans. Circuits Syst. Video Technol.}, 2024.

\bibitem{chung2023depth}
J.~Chung, J.~Oh, and K.~M. Lee, ``Depth-regularized optimization for 3d gaussian splatting in few-shot images,'' \emph{arXiv:2311.13398}, 2023.

\bibitem{zhu2023fsgs}
Z.~Zhu, Z.~Fan, Y.~Jiang, and Z.~Wang, ``Fsgs: Real-time few-shot view synthesis using gaussian splatting,'' \emph{arXiv:2312.00451}, 2023.

\bibitem{swann2024touch}
A.~Swann, M.~Strong, W.~K. Do, G.~S. Camps, M.~Schwager, and M.~Kennedy~III, ``Touch-gs: Visual-tactile supervised 3d gaussian splatting,'' \emph{arXiv:2403.09875}, 2024.

\bibitem{li2024dngaussian}
J.~Li, J.~Zhang, X.~Bai, J.~Zheng, X.~Ning, J.~Zhou, and L.~Gu, ``Dngaussian: Optimizing sparse-view 3d gaussian radiance fields with global-local depth normalization,'' \emph{arXiv:2403.06912}, 2024.

\bibitem{liu2024georgs}
Z.~Liu, J.~Su, G.~Cai, Y.~Chen, B.~Zeng, and Z.~Wang, ``Georgs: Geometric regularization for real-time novel view synthesis from sparse inputs,'' \emph{IEEE Trans. Circuits Syst. Video Technol.}, 2024.

\bibitem{liu2025deceptive}
X.~Liu, J.~Chen, S.-H. Kao, Y.-W. Tai, and C.-K. Tang, ``Deceptive-nerf/3dgs: Diffusion-generated pseudo-observations for high-quality sparse-view reconstruction,'' in \emph{Eur. Conf. Comput. Vis.}\hskip 1em plus 0.5em minus 0.4em\relax Springer, 2025, pp. 337--355.

\bibitem{paliwal2025coherentgs}
A.~Paliwal, W.~Ye, J.~Xiong, D.~Kotovenko, R.~Ranjan, V.~Chandra, and N.~K. Kalantari, ``Coherentgs: Sparse novel view synthesis with coherent 3d gaussians,'' in \emph{Eur. Conf. Comput. Vis.}\hskip 1em plus 0.5em minus 0.4em\relax Springer, 2025, pp. 19--37.

\bibitem{zhang2025cor}
J.~Zhang, J.~Li, X.~Yu, L.~Huang, L.~Gu, J.~Zheng, and X.~Bai, ``Cor-gs: sparse-view 3d gaussian splatting via co-regularization,'' in \emph{Eur. Conf. Comput. Vis.}\hskip 1em plus 0.5em minus 0.4em\relax Springer, 2025, pp. 335--352.

\bibitem{yang2024gaussianobject}
C.~Yang, S.~Li, J.~Fang, R.~Liang, L.~Xie, X.~Zhang, W.~Shen, and Q.~Tian, ``Gaussianobject: Just taking four images to get a high-quality 3d object with gaussian splatting,'' \emph{arXiv:2402.10259}, 2024.

\bibitem{moreau2023human}
A.~Moreau, J.~Song, H.~Dhamo, R.~Shaw, Y.~Zhou, and E.~P{\'e}rez-Pellitero, ``Human gaussian splatting: Real-time rendering of animatable avatars,'' \emph{arXiv:2311.17113}, 2023.

\bibitem{zheng2024gps-gaussian}
S.~Zheng, B.~Zhou, R.~Shao, B.~Liu, S.~Zhang, L.~Nie, and Y.~Liu, ``{GPS}-{Gaussian}: {Generalizable} {Pixel}-wise {3D} {Gaussian} {Splatting} for {Real}-time {Human} {Novel} {View} {Synthesis},'' 2024.

\bibitem{kocabas2023hugs}
M.~Kocabas, J.-H.~R. Chang, J.~Gabriel, O.~Tuzel, and A.~Ranjan, ``Hugs: Human gaussian splats,'' \emph{arXiv:2311.17910}, 2023.

\bibitem{hu2023gaussianavatar}
L.~Hu, H.~Zhang, Y.~Zhang, B.~Zhou, B.~Liu, S.~Zhang, and L.~Nie, ``Gaussianavatar: Towards realistic human avatar modeling from a single video via animatable 3d gaussians,'' \emph{arXiv:2312.02134}, 2023.

\bibitem{hu2023gauhuman}
S.~Hu and Z.~Liu, ``Gauhuman: Articulated gaussian splatting from monocular human videos,'' \emph{arXiv:2312.02973}, 2023.

\bibitem{qian20233dgsavatar}
Z.~Qian, S.~Wang, M.~Mihajlovic, A.~Geiger, and S.~Tang, ``3dgs-avatar: Animatable avatars via deformable 3d gaussian splatting,'' \emph{arXiv:2312.09228}, 2023.

\bibitem{pang2023ash}
H.~Pang, H.~Zhu, A.~Kortylewski, C.~Theobalt, and M.~Habermann, ``Ash: Animatable gaussian splats for efficient and photoreal human rendering,'' \emph{arXiv:2312.05941}, 2023.

\bibitem{li2023animatable}
Z.~Li, Z.~Zheng, L.~Wang, and Y.~Liu, ``Animatable gaussians: Learning pose-dependent gaussian maps for high-fidelity human avatar modeling,'' \emph{arXiv:2311.16096}, 2023.

\bibitem{sheng2024open}
Z.~Sheng, F.~Liu, M.~Liu, F.~Zheng, and L.~Nie, ``Open-set synthesis for free-viewpoint human body reenactment of novel poses,'' \emph{IEEE Trans. Circuits Syst. Video Technol.}, 2024.

\bibitem{Loper2015smpl}
M.~Loper, N.~Mahmood, J.~Romero, G.~Pons-Moll, and M.~J. Black, ``{SMPL}: A skinned multi-person linear model,'' \emph{ACM Trans. Graph. (Proc. SIGGRAPH Asia)}, vol.~34, no.~6, pp. 248:1--248:16, oct 2015.

\bibitem{pavlakos2019smplx}
G.~Pavlakos, V.~Choutas, N.~Ghorbani, T.~Bolkart, A.~A.~A. Osman, D.~Tzionas, and M.~J. Black, ``Expressive body capture: {3D} hands, face, and body from a single image,'' in \emph{Proc. IEEE Conf. Comput. Vis. Pattern Recognit.}, 2019, pp. 10\,975--10\,985.

\bibitem{chen2024monogaussianavatar}
Y.~Chen, L.~Wang, Q.~Li, H.~Xiao, S.~Zhang, H.~Yao, and Y.~Liu, ``Monogaussianavatar: Monocular gaussian point-based head avatar,'' in \emph{ACM SIGGRAPH Conf. Papers}, 2024, pp. 1--9.

\bibitem{qian2023gaussianavatars}
S.~Qian, T.~Kirschstein, L.~Schoneveld, D.~Davoli, S.~Giebenhain, and M.~Nie{\ss}ner, ``Gaussianavatars: Photorealistic head avatars with rigged 3d gaussians,'' \emph{arXiv:2312.02069}, 2023.

\bibitem{zhao2024psavatar}
Z.~Zhao, Z.~Bao, Q.~Li, G.~Qiu, and K.~Liu, ``Psavatar: A point-based morphable shape model for real-time head avatar creation with 3d gaussian splatting,'' \emph{arXiv:2401.12900}, 2024.

\bibitem{dhamo2025headgas}
H.~Dhamo, Y.~Nie, A.~Moreau, J.~Song, R.~Shaw, Y.~Zhou, and E.~P{\'e}rez-Pellitero, ``Headgas: Real-time animatable head avatars via 3d gaussian splatting,'' in \emph{Eur. Conf. Comput. Vis.}\hskip 1em plus 0.5em minus 0.4em\relax Springer, 2025, pp. 459--476.

\bibitem{teotia2024gaussianheads}
K.~Teotia, H.~Kim, P.~Garrido, M.~Habermann, M.~Elgharib, and C.~Theobalt, ``Gaussianheads: End-to-end learning of drivable gaussian head avatars from coarse-to-fine representations,'' \emph{ACM Trans. Graph.}, vol.~43, no.~6, pp. 1--12, 2024.

\bibitem{siggraphAsia2017flame}
\BIBentryALTinterwordspacing
T.~Li, T.~Bolkart, M.~J. Black, H.~Li, and J.~Romero, ``Learning a model of facial shape and expression from {4D} scans,'' \emph{ACM Trans. Graph. (Proc. SIGGRAPH Asia)}, vol.~36, no.~6, pp. 194:1--194:17, 2017. [Online]. Available: \url{https://doi.org/10.1145/3130800.3130813}
\BIBentrySTDinterwordspacing

\bibitem{xu2023gaussian}
Y.~Xu, B.~Chen, Z.~Li, H.~Zhang, L.~Wang, Z.~Zheng, and Y.~Liu, ``Gaussian head avatar: Ultra high-fidelity head avatar via dynamic gaussians,'' \emph{arXiv:2312.03029}, 2023.

\bibitem{xu2024gaussian}
------, ``Gaussian head avatar: Ultra high-fidelity head avatar via dynamic gaussians,'' in \emph{Proc. IEEE/CVF Conf. Comput. Vis. Pattern Recognit.}, 2024, pp. 1931--1941.

\bibitem{xu20253d}
Y.~Xu, L.~Wang, Z.~Zheng, Z.~Su, and Y.~Liu, ``3d gaussian parametric head model,'' in \emph{Eur. Conf. Comput. Vis.}\hskip 1em plus 0.5em minus 0.4em\relax Springer, 2025, pp. 129--147.

\bibitem{ma20243d}
S.~Ma, Y.~Weng, T.~Shao, and K.~Zhou, ``3d gaussian blendshapes for head avatar animation,'' in \emph{ACM SIGGRAPH Conf. Papers}, 2024, pp. 1--10.

\bibitem{xiang2024flashavatar}
J.~Xiang, X.~Gao, Y.~Guo, and J.~Zhang, ``Flashavatar: High-fidelity head avatar with efficient gaussian embedding,'' in \emph{Proc. IEEE/CVF Conf. Comput. Vis. Pattern Recognit.}, 2024, pp. 1802--1812.

\bibitem{li2025talkinggaussian}
J.~Li, J.~Zhang, X.~Bai, J.~Zheng, X.~Ning, J.~Zhou, and L.~Gu, ``Talkinggaussian: Structure-persistent 3d talking head synthesis via gaussian splatting,'' in \emph{Eur. Conf. Comput. Vis.}\hskip 1em plus 0.5em minus 0.4em\relax Springer, 2025, pp. 127--145.

\bibitem{cho2024gaussiantalker}
K.~Cho, J.~Lee, H.~Yoon, Y.~Hong, J.~Ko, S.~Ahn, and S.~Kim, ``Gaussiantalker: Real-time talking head synthesis with 3d gaussian splatting,'' in \emph{Proc. ACM Int. Conf. Multimedia}, 2024, pp. 10\,985--10\,994.

\bibitem{yu2024gaussiantalker}
H.~Yu, Z.~Qu, Q.~Yu, J.~Chen, Z.~Jiang, Z.~Chen, S.~Zhang, J.~Xu, F.~Wu, C.~Lv \emph{et~al.}, ``Gaussiantalker: Speaker-specific talking head synthesis via 3d gaussian splatting,'' in \emph{Proc. ACM Int. Conf. Multimedia}, 2024, pp. 3548--3557.

\bibitem{luo2024gaussianhair}
H.~Luo, M.~Ouyang, Z.~Zhao, S.~Jiang, L.~Zhang, Q.~Zhang, W.~Yang, L.~Xu, and J.~Yu, ``Gaussianhair: Hair modeling and rendering with light-aware gaussians,'' \emph{arXiv:2402.10483}, 2024.

\bibitem{bolanos2024charshadow}
L.~Bolanos, S.-Y. Su, and H.~Rhodin, ``Gaussian {Shadow} {Casting} for {Neural} {Characters},'' 2024.

\bibitem{abdal2023gaussian}
R.~Abdal, W.~Yifan, Z.~Shi, Y.~Xu, R.~Po, Z.~Kuang, Q.~Chen, D.-Y. Yeung, and G.~Wetzstein, ``Gaussian shell maps for efficient 3d human generation,'' \emph{arXiv:2311.17857}, 2023.

\bibitem{liu2023humangaussian}
X.~Liu, X.~Zhan, J.~Tang, Y.~Shan, G.~Zeng, D.~Lin, X.~Liu, and Z.~Liu, ``Humangaussian: Text-driven 3d human generation with gaussian splatting,'' \emph{arXiv:2311.17061}, 2023.

\bibitem{kirschstein2024gghead}
T.~Kirschstein, S.~Giebenhain, J.~Tang, M.~Georgopoulos, and M.~Nie{\ss}ner, ``Gghead: Fast and generalizable 3d gaussian heads,'' in \emph{ACM SIGGRAPH Asia Conf. Papers}, 2024, pp. 1--11.

\bibitem{tang2023dreamgaussian}
J.~Tang, J.~Ren, H.~Zhou, Z.~Liu, and G.~Zeng, ``Dreamgaussian: Generative gaussian splatting for efficient 3d content creation,'' \emph{arXiv:2309.16653}, 2023.

\bibitem{ren2023dreamgaussian4d}
J.~Ren, L.~Pan, J.~Tang, C.~Zhang, A.~Cao, G.~Zeng, and Z.~Liu, ``Dreamgaussian4d: Generative 4d gaussian splatting,'' \emph{arXiv:2312.17142}, 2023.

\bibitem{zhang2023repaint123}
J.~Zhang, Z.~Tang, Y.~Pang, X.~Cheng, P.~Jin, Y.~Wei, W.~Yu, M.~Ning, and L.~Yuan, ``Repaint123: Fast and high-quality one image to 3d generation with progressive controllable 2d repainting,'' \emph{arXiv:2312.13271}, 2023.

\bibitem{zhou2024gala3d}
X.~Zhou, X.~Ran, Y.~Xiong, J.~He, Z.~Lin, Y.~Wang, D.~Sun, and M.-H. Yang, ``Gala3d: Towards text-to-3d complex scene generation via layout-guided generative gaussian splatting,'' \emph{arXiv:2402.07207}, 2024.

\bibitem{poole2022dreamfusion}
B.~Poole, A.~Jain, J.~T. Barron, and B.~Mildenhall, ``Dreamfusion: Text-to-3d using 2d diffusion,'' \emph{arXiv:2209.14988}, 2022.

\bibitem{chen2023text}
Z.~Chen, F.~Wang, and H.~Liu, ``Text-to-3d using gaussian splatting,'' \emph{arXiv:2309.16585}, 2023.

\bibitem{yi2023gaussiandreamer}
T.~Yi, J.~Fang, J.~Wang, G.~Wu, L.~Xie, X.~Zhang, W.~Liu, Q.~Tian, and X.~Wang, ``Gaussiandreamer: Fast generation from text to 3d gaussians by bridging 2d and 3d diffusion models,'' \emph{arXiv preprint arXiv}, vol. 2310, 2023.

\bibitem{shen2025vista3d}
Q.~Shen, X.~Yang, M.~B. Mi, and X.~Wang, ``Vista3d: Unravel the 3d darkside of a single image,'' in \emph{Eur. Conf. Comput. Vis.}\hskip 1em plus 0.5em minus 0.4em\relax Springer, 2025, pp. 405--421.

\bibitem{li2023gaussiandiffusion}
X.~Li, H.~Wang, and K.-K. Tseng, ``Gaussiandiffusion: 3d gaussian splatting for denoising diffusion probabilistic models with structured noise,'' \emph{arXiv:2311.11221}, 2023.

\bibitem{liang2023luciddreamer}
Y.~Liang, X.~Yang, J.~Lin, H.~Li, X.~Xu, and Y.~Chen, ``Luciddreamer: Towards high-fidelity text-to-3d generation via interval score matching,'' \emph{arXiv:2311.11284}, 2023.

\bibitem{di2024hyper}
D.~Di, J.~Yang, C.~Luo, Z.~Xue, W.~Chen, X.~Yang, and Y.~Gao, ``Hyper-3dg: Text-to-3d gaussian generation via hypergraph,'' \emph{arXiv:2403.09236}, 2024.

\bibitem{yang2023learn}
X.~Yang, Y.~Chen, C.~Chen, C.~Zhang, Y.~Xu, X.~Yang, F.~Liu, and G.~Lin, ``Learn to optimize denoising scores for 3d generation: A unified and improved diffusion prior on nerf and 3d gaussian splatting,'' \emph{arXiv:2312.04820}, 2023.

\bibitem{zhuo2025vividdreamer}
W.~Zhuo, F.~Ma, H.~Fan, and Y.~Yang, ``Vividdreamer: Invariant score distillation for hyper-realistic text-to-3d generation,'' in \emph{Eur. Conf. Comput. Vis.}\hskip 1em plus 0.5em minus 0.4em\relax Springer, 2025, pp. 122--139.

\bibitem{song2020denoising}
J.~Song, C.~Meng, and S.~Ermon, ``Denoising diffusion implicit models,'' \emph{arXiv:2010.02502}, 2020.

\bibitem{melas20243d}
L.~Melas-Kyriazi, I.~Laina, C.~Rupprecht, N.~Neverova, A.~Vedaldi, O.~Gafni, and F.~Kokkinos, ``Im-3d: Iterative multiview diffusion and reconstruction for high-quality 3d generation,'' \emph{arXiv:2402.08682}, 2024.

\bibitem{voleti2025sv3d}
V.~Voleti, C.-H. Yao, M.~Boss, A.~Letts, D.~Pankratz, D.~Tochilkin, C.~Laforte, R.~Rombach, and V.~Jampani, ``Sv3d: Novel multi-view synthesis and 3d generation from a single image using latent video diffusion,'' in \emph{Eur. Conf. Comput. Vis.}\hskip 1em plus 0.5em minus 0.4em\relax Springer, 2025, pp. 439--457.

\bibitem{gao2024cat3d}
R.~Gao, A.~Holynski, P.~Henzler, A.~Brussee, R.~Martin-Brualla, P.~Srinivasan, J.~T. Barron, and B.~Poole, ``Cat3d: Create anything in 3d with multi-view diffusion models,'' \emph{arXiv:2405.10314}, 2024.

\bibitem{han2025vfusion3d}
J.~Han, F.~Kokkinos, and P.~Torr, ``Vfusion3d: Learning scalable 3d generative models from video diffusion models,'' in \emph{Eur. Conf. Comput. Vis.}\hskip 1em plus 0.5em minus 0.4em\relax Springer, 2025, pp. 333--350.

\bibitem{yang2024hi3d}
H.~Yang, Y.~Chen, Y.~Pan, T.~Yao, Z.~Chen, C.-W. Ngo, and T.~Mei, ``Hi3d: Pursuing high-resolution image-to-3d generation with video diffusion models,'' in \emph{Proc. ACM Int. Conf. Multimedia}, 2024, pp. 6870--6879.

\bibitem{jiang2024brightdreamer}
L.~Jiang and L.~Wang, ``Brightdreamer: Generic 3d gaussian generative framework for fast text-to-3d synthesis,'' \emph{arXiv:2403.11273}, 2024.

\bibitem{tang2024lgm}
J.~Tang, Z.~Chen, X.~Chen, T.~Wang, G.~Zeng, and Z.~Liu, ``Lgm: Large multi-view gaussian model for high-resolution 3d content creation,'' \emph{arXiv:2402.05054}, 2024.

\bibitem{zhang2024gaussiancube}
B.~Zhang, Y.~Cheng, J.~Yang, C.~Wang, F.~Zhao, Y.~Tang, D.~Chen, and B.~Guo, ``Gaussiancube: Structuring gaussian splatting using optimal transport for 3d generative modeling,'' \emph{arXiv:2403.19655}, 2024.

\bibitem{liu2024sketchdream}
F.-L. Liu, H.~Fu, Y.-K. Lai, and L.~Gao, ``Sketchdream: Sketch-based text-to-3d generation and editing,'' \emph{ACM Trans. Graph.}, vol.~43, no.~4, pp. 1--13, 2024.

\bibitem{he2024gvgen}
X.~He, J.~Chen, S.~Peng, D.~Huang, Y.~Li, X.~Huang, C.~Yuan, W.~Ouyang, and T.~He, ``Gvgen: Text-to-3d generation with volumetric representation,'' \emph{arXiv:2403.12957}, 2024.

\bibitem{lu2024large}
L.~Lu, H.~Gao, T.~Dai, Y.~Zha, Z.~Hou, J.~Wu, and S.-T. Xia, ``Large point-to-gaussian model for image-to-3d generation,'' in \emph{Proc. ACM Int. Conf. Multimedia}, 2024, pp. 10\,843--10\,852.

\bibitem{roessle2024l3dg}
B.~Roessle, N.~M{\"u}ller, L.~Porzi, S.~Rota~Bul{\`o}, P.~Kontschieder, A.~Dai, and M.~Nie{\ss}ner, ``L3dg: Latent 3d gaussian diffusion,'' in \emph{ACM SIGGRAPH Asia Conf. Papers}, 2024, pp. 1--11.

\bibitem{yu2024get3dgs}
H.~Yu, W.~Gong, J.~Chen, and H.~Ma, ``Get3dgs: Generate 3d gaussians based on points deformation fields,'' \emph{IEEE Trans. Circuits Syst. Video Technol.}, 2024.

\bibitem{yuan2024gavatar}
Y.~Yuan, X.~Li, Y.~Huang, S.~De~Mello, K.~Nagano, J.~Kautz, and U.~Iqbal, ``Gavatar: Animatable 3d gaussian avatars with implicit mesh learning,'' in \emph{Proc. IEEE/CVF Conf. Comput. Vis. Pattern Recognit.}, 2024, pp. 896--905.

\bibitem{feng2024fdgaussian}
Q.~Feng, Z.~Xing, Z.~Wu, and Y.-G. Jiang, ``Fdgaussian: Fast gaussian splatting from single image via geometric-aware diffusion model,'' \emph{arXiv:2403.10242}, 2024.

\bibitem{chen2025cascade}
Y.~Chen, J.~Fang, Y.~Huang, T.~Yi, X.~Zhang, L.~Xie, X.~Wang, W.~Dai, H.~Xiong, and Q.~Tian, ``Cascade-zero123: One image to highly consistent 3d with self-prompted nearby views,'' in \emph{Eur. Conf. Comput. Vis.}\hskip 1em plus 0.5em minus 0.4em\relax Springer, 2025, pp. 311--330.

\bibitem{liu2023zero}
R.~Liu, R.~Wu, B.~Van~Hoorick, P.~Tokmakov, S.~Zakharov, and C.~Vondrick, ``Zero-1-to-3: Zero-shot one image to 3d object,'' in \emph{Proc. IEEE/CVF Int. Conf. Comput. Vis.}, 2023, pp. 9298--9309.

\bibitem{vilesov2023cg3d}
A.~Vilesov, P.~Chari, and A.~Kadambi, ``Cg3d: Compositional generation for text-to-3d via gaussian splatting,'' \emph{arXiv:2311.17907}, 2023.

\bibitem{li2025dreamscene}
H.~Li, H.~Shi, W.~Zhang, W.~Wu, Y.~Liao, L.~Wang, L.-h. Lee, and P.~Y. Zhou, ``Dreamscene: 3d gaussian-based text-to-3d scene generation via formation pattern sampling,'' in \emph{Eur. Conf. Comput. Vis.}\hskip 1em plus 0.5em minus 0.4em\relax Springer, 2025, pp. 214--230.

\bibitem{chen2025comboverse}
Y.~Chen, T.~Wang, T.~Wu, X.~Pan, K.~Jia, and Z.~Liu, ``Comboverse: Compositional 3d assets creation using spatially-aware diffusion guidance,'' in \emph{Eur. Conf. Comput. Vis.}\hskip 1em plus 0.5em minus 0.4em\relax Springer, 2025, pp. 128--146.

\bibitem{chung2023luciddreamer}
J.~Chung, S.~Lee, H.~Nam, J.~Lee, and K.~M. Lee, ``Luciddreamer: Domain-free generation of 3d gaussian splatting scenes,'' \emph{arXiv:2311.13384}, 2023.

\bibitem{ouyang2023text2immersion}
H.~Ouyang, K.~Heal, S.~Lombardi, and T.~Sun, ``Text2immersion: Generative immersive scene with 3d gaussians,'' \emph{arXiv:2312.09242}, 2023.

\bibitem{zhou2025dreamscene360}
S.~Zhou, Z.~Fan, D.~Xu, H.~Chang, P.~Chari, T.~Bharadwaj, S.~You, Z.~Wang, and A.~Kadambi, ``Dreamscene360: Unconstrained text-to-3d scene generation with panoramic gaussian splatting,'' in \emph{Eur. Conf. Comput. Vis.}\hskip 1em plus 0.5em minus 0.4em\relax Springer, 2025, pp. 324--342.

\bibitem{lugmayr2022repaint}
A.~Lugmayr, M.~Danelljan, A.~Romero, F.~Yu, R.~Timofte, and L.~Van~Gool, ``Repaint: Inpainting using denoising diffusion probabilistic models,'' in \emph{Proc. IEEE/CVF Conf. Comput. Vis. Pattern Recognit.}, 2022, pp. 11\,461--11\,471.

\bibitem{ling2023align}
H.~Ling, S.~W. Kim, A.~Torralba, S.~Fidler, and K.~Kreis, ``Align your gaussians: Text-to-4d with dynamic 3d gaussians and composed diffusion models,'' \emph{arXiv:2312.13763}, 2023.

\bibitem{gao2024gaussianflow}
Q.~Gao, Q.~Xu, Z.~Cao, B.~Mildenhall, W.~Ma, L.~Chen, D.~Tang, and U.~Neumann, ``Gaussianflow: Splatting gaussian dynamics for 4d content creation,'' \emph{arXiv:2403.12365}, 2024.

\bibitem{bahmani20244d}
S.~Bahmani, I.~Skorokhodov, V.~Rong, G.~Wetzstein, L.~Guibas, P.~Wonka, S.~Tulyakov, J.~J. Park, A.~Tagliasacchi, and D.~B. Lindell, ``4d-fy: Text-to-4d generation using hybrid score distillation sampling,'' in \emph{Proc. IEEE/CVF Conf. Comput. Vis. Pattern Recognit.}, 2024, pp. 7996--8006.

\bibitem{bahmani2025tc4d}
S.~Bahmani, X.~Liu, W.~Yifan, I.~Skorokhodov, V.~Rong, Z.~Liu, X.~Liu, J.~J. Park, S.~Tulyakov, G.~Wetzstein \emph{et~al.}, ``Tc4d: Trajectory-conditioned text-to-4d generation,'' in \emph{Eur. Conf. Comput. Vis.}\hskip 1em plus 0.5em minus 0.4em\relax Springer, 2025, pp. 53--72.

\bibitem{zheng2024unified}
Y.~Zheng, X.~Li, K.~Nagano, S.~Liu, O.~Hilliges, and S.~De~Mello, ``A unified approach for text-and image-guided 4d scene generation,'' in \emph{Proc. IEEE/CVF Conf. Comput. Vis. Pattern Recognit.}, 2024, pp. 7300--7309.

\bibitem{shi2023mvdream}
Y.~Shi, P.~Wang, J.~Ye, M.~Long, K.~Li, and X.~Yang, ``Mvdream: Multi-view diffusion for 3d generation,'' \emph{arXiv:2308.16512}, 2023.

\bibitem{yin20234dgen}
Y.~Yin, D.~Xu, Z.~Wang, Y.~Zhao, and Y.~Wei, ``4dgen: Grounded 4d content generation with spatial-temporal consistency,'' \emph{arXiv:2312.17225}, 2023.

\bibitem{pan2024fast}
Z.~Pan, Z.~Yang, X.~Zhu, and L.~Zhang, ``Fast dynamic 3d object generation from a single-view video,'' \emph{arXiv:2401.08742}, 2024.

\bibitem{sun2024eg4d}
Q.~Sun, Z.~Guo, Z.~Wan, J.~N. Yan, S.~Yin, W.~Zhou, J.~Liao, and H.~Li, ``Eg4d: Explicit generation of 4d object without score distillation,'' \emph{arXiv:2405.18132}, 2024.

\bibitem{zeng2025stag4d}
Y.~Zeng, Y.~Jiang, S.~Zhu, Y.~Lu, Y.~Lin, H.~Zhu, W.~Hu, X.~Cao, and Y.~Yao, ``Stag4d: Spatial-temporal anchored generative 4d gaussians,'' in \emph{Eur. Conf. Comput. Vis.}\hskip 1em plus 0.5em minus 0.4em\relax Springer, 2025, pp. 163--179.

\bibitem{cao2023hexplane}
A.~Cao and J.~Johnson, ``Hexplane: A fast representation for dynamic scenes,'' in \emph{Proc. IEEE/CVF Conf. Comput. Vis. Pattern Recognit.}, 2023, pp. 130--141.

\bibitem{zhang2024bags}
T.~Zhang, Q.~Gao, W.~Li, L.~Liu, and B.~Chen, ``Bags: Building animatable gaussian splatting from a monocular video with diffusion priors,'' \emph{arXiv:2403.11427}, 2024.

\bibitem{wu2025sc4d}
Z.~Wu, C.~Yu, Y.~Jiang, C.~Cao, F.~Wang, and X.~Bai, ``Sc4d: Sparse-controlled video-to-4d generation and motion transfer,'' in \emph{Eur. Conf. Comput. Vis.}\hskip 1em plus 0.5em minus 0.4em\relax Springer, 2025, pp. 361--379.

\bibitem{zhou2023drivinggaussian}
X.~Zhou, Z.~Lin, X.~Shan, Y.~Wang, D.~Sun, and M.-H. Yang, ``Drivinggaussian: Composite gaussian splatting for surrounding dynamic autonomous driving scenes,'' \emph{arXiv:2312.07920}, 2023.

\bibitem{yan2024street}
Y.~Yan, H.~Lin, C.~Zhou, W.~Wang, H.~Sun, K.~Zhan, X.~Lang, X.~Zhou, and S.~Peng, ``Street gaussians for modeling dynamic urban scenes,'' \emph{arXiv:2401.01339}, 2024.

\bibitem{zhou2024hugs}
H.~Zhou, J.~Shao, L.~Xu, D.~Bai, W.~Qiu, B.~Liu, Y.~Wang, A.~Geiger, and Y.~Liao, ``Hugs: Holistic urban 3d scene understanding via gaussian splatting,'' \emph{arXiv:2403.12722}, 2024.

\bibitem{zhao2024tclc}
C.~Zhao, S.~Sun, R.~Wang, Y.~Guo, J.-J. Wan, Z.~Huang, X.~Huang, Y.~V. Chen, and L.~Ren, ``Tclc-gs: Tightly coupled lidar-camera gaussian splatting for surrounding autonomous driving scenes,'' \emph{arXiv:2404.02410}, 2024.

\bibitem{herau20243dgs}
Q.~Herau, M.~Bennehar, A.~Moreau, N.~Piasco, L.~Roldao, D.~Tsishkou, C.~Migniot, P.~Vasseur, and C.~Demonceaux, ``3dgs-calib: 3d gaussian splatting for multimodal spatiotemporal calibration,'' \emph{arXiv:2403.11577}, 2024.

\bibitem{yan2023gs}
C.~Yan, D.~Qu, D.~Wang, D.~Xu, Z.~Wang, B.~Zhao, and X.~Li, ``Gs-slam: Dense visual slam with 3d gaussian splatting,'' \emph{arXiv:2311.11700}, 2023.

\bibitem{huang2023photo}
H.~Huang, L.~Li, H.~Cheng, and S.-K. Yeung, ``Photo-slam: Real-time simultaneous localization and photorealistic mapping for monocular, stereo, and rgb-d cameras,'' \emph{arXiv:2311.16728}, 2023.

\bibitem{keetha2023splatam}
N.~Keetha, J.~Karhade, K.~M. Jatavallabhula, G.~Yang, S.~Scherer, D.~Ramanan, and J.~Luiten, ``Splatam: Splat, track \& map 3d gaussians for dense rgb-d slam,'' \emph{arXiv:2312.02126}, 2023.

\bibitem{yugay2023gaussian}
V.~Yugay, Y.~Li, T.~Gevers, and M.~R. Oswald, ``Gaussian-slam: Photo-realistic dense slam with gaussian splatting,'' \emph{arXiv:2312.10070}, 2023.

\bibitem{hu2025cg}
J.~Hu, X.~Chen, B.~Feng, G.~Li, L.~Yang, H.~Bao, G.~Zhang, and Z.~Cui, ``Cg-slam: Efficient dense rgb-d slam in a consistent uncertainty-aware 3d gaussian field,'' in \emph{Eur. Conf. Comput. Vis.}\hskip 1em plus 0.5em minus 0.4em\relax Springer, 2025, pp. 93--112.

\bibitem{ha2024rgbd}
S.~Ha, J.~Yeon, and H.~Yu, ``Rgbd gs-icp slam,'' \emph{arXiv:2403.12550}, 2024.

\bibitem{rublee2011orb}
E.~Rublee, V.~Rabaud, K.~Konolige, and G.~Bradski, ``Orb: An efficient alternative to sift or surf,'' in \emph{Proc. Int. Conf. Comput. Vis. (ICCV)}.\hskip 1em plus 0.5em minus 0.4em\relax IEEE, 2011, pp. 2564--2571.

\bibitem{more2006levenberg}
J.~J. Mor{\'e}, ``The levenberg-marquardt algorithm: implementation and theory,'' in \emph{Proc. Dundee Biennial Conf. Numer. Anal.}\hskip 1em plus 0.5em minus 0.4em\relax Springer, 2006, pp. 105--116.

\bibitem{segal2009generalized}
A.~Segal, D.~Haehnel, and S.~Thrun, ``Generalized-icp,'' in \emph{Robotics: Sci. Syst. (RSS)}, vol.~2, no.~4.\hskip 1em plus 0.5em minus 0.4em\relax Seattle, WA, 2009, p. 435.

\bibitem{li2024sgs}
M.~Li, S.~Liu, and H.~Zhou, ``Sgs-slam: Semantic gaussian splatting for neural dense slam,'' \emph{arXiv:2402.03246}, 2024.

\bibitem{zhu2024semgauss}
S.~Zhu, R.~Qin, G.~Wang, J.~Liu, and H.~Wang, ``Semgauss-slam: Dense semantic gaussian splatting slam,'' \emph{arXiv:2403.07494}, 2024.

\bibitem{ji2024neds}
Y.~Ji, Y.~Liu, G.~Xie, B.~Ma, and Z.~Xie, ``Neds-slam: A novel neural explicit dense semantic slam framework using 3d gaussian splatting,'' \emph{arXiv:2403.11679}, 2024.

\bibitem{jiang20243dgs}
P.~Jiang, G.~Pandey, and S.~Saripalli, ``3dgs-reloc: 3d gaussian splatting for map representation and visual relocalization,'' \emph{arXiv:2403.11367}, 2024.

\bibitem{lei2024gaussnav}
X.~Lei, M.~Wang, W.~Zhou, and H.~Li, ``Gaussnav: Gaussian splatting for visual navigation,'' \emph{arXiv:2403.11625}, 2024.

\bibitem{straub2019replica}
J.~Straub, T.~Whelan, L.~Ma, Y.~Chen, E.~Wijmans, S.~Green, J.~J. Engel, R.~Mur-Artal, C.~Ren, S.~Verma \emph{et~al.}, ``The replica dataset: A digital replica of indoor spaces,'' \emph{arXiv:1906.05797}, 2019.

\bibitem{yan2024gs}
C.~Yan, D.~Qu, D.~Xu, B.~Zhao, Z.~Wang, D.~Wang, and X.~Li, ``Gs-slam: Dense visual slam with 3d gaussian splatting,'' in \emph{Proc. IEEE/CVF Conf. Comput. Vis. Pattern Recognit.}, 2024, pp. 19\,595--19\,604.

\bibitem{guo2024motiongs}
X.~Guo, W.~Zhang, R.~Liu, P.~Han, and H.~Chen, ``Motiongs: Compact gaussian splatting slam by motion filter,'' \emph{arXiv:2405.11129}, 2024.

\bibitem{zhu2024loopsplat}
L.~Zhu, Y.~Li, E.~Sandstr{\"o}m, S.~Huang, K.~Schindler, and I.~Armeni, ``Loopsplat: Loop closure by registering 3d gaussian splats,'' \emph{arXiv:2408.10154}, 2024.

\bibitem{luiten2023dynamic}
J.~Luiten, G.~Kopanas, B.~Leibe, and D.~Ramanan, ``Dynamic 3d gaussians: Tracking by persistent dynamic view synthesis,'' \emph{arXiv:2308.09713}, 2023.

\bibitem{sun20243dgstream}
J.~Sun, H.~Jiao, G.~Li, Z.~Zhang, L.~Zhao, and W.~Xing, ``3dgstream: On-the-fly training of 3d gaussians for efficient streaming of photo-realistic free-viewpoint videos,'' \emph{arXiv:2403.01444}, 2024.

\bibitem{muller2022instant}
T.~M{\"u}ller, A.~Evans, C.~Schied, and A.~Keller, ``Instant neural graphics primitives with a multiresolution hash encoding,'' \emph{ACM Trans. Graph.}, vol.~41, no.~4, pp. 1--15, 2022.

\bibitem{shaw2023swags}
R.~Shaw, J.~Song, A.~Moreau, M.~Nazarczuk, S.~Catley-Chandar, H.~Dhamo, and E.~Perez-Pellitero, ``Swags: Sampling windows adaptively for dynamic 3d gaussian splatting,'' \emph{arXiv:2312.13308}, 2023.

\bibitem{xiao2024bridging}
Y.~Xiao, X.~Wang, J.~Li, H.~Cai, Y.~Fan, N.~Xue, M.~Yang, Y.~Shen, and S.~Gao, ``Bridging 3d gaussian and mesh for freeview video rendering,'' \emph{arXiv:2403.11453}, 2024.

\bibitem{yang2023deformable}
Z.~Yang, X.~Gao, W.~Zhou, S.~Jiao, Y.~Zhang, and X.~Jin, ``Deformable 3d gaussians for high-fidelity monocular dynamic scene reconstruction,'' \emph{arXiv:2309.13101}, 2023.

\bibitem{liang2023gaufre}
Y.~Liang, N.~Khan, Z.~Li, T.~Nguyen-Phuoc, D.~Lanman, J.~Tompkin, and L.~Xiao, ``Gaufre: Gaussian deformation fields for real-time dynamic novel view synthesis,'' \emph{arXiv:2312.11458}, 2023.

\bibitem{wu20234d}
G.~Wu, T.~Yi, J.~Fang, L.~Xie, X.~Zhang, W.~Wei, W.~Liu, Q.~Tian, and X.~Wang, ``4d gaussian splatting for real-time dynamic scene rendering,'' \emph{arXiv:2310.08528}, 2023.

\bibitem{duisterhof2023md}
B.~P. Duisterhof, Z.~Mandi, Y.~Yao, J.-W. Liu, M.~Z. Shou, S.~Song, and J.~Ichnowski, ``Md-splatting: Learning metric deformation from 4d gaussians in highly deformable scenes,'' \emph{arXiv:2312.00583}, 2023.

\bibitem{guo2024motion}
Z.~Guo, W.~Zhou, L.~Li, M.~Wang, and H.~Li, ``Motion-aware 3d gaussian splatting for efficient dynamic scene reconstruction,'' \emph{IEEE Trans. Circuits Syst. Video Technol.}, 2024.

\bibitem{li2024st}
D.~Li, S.-S. Huang, Z.~Lu, X.~Duan, and H.~Huang, ``St-4dgs: Spatial-temporally consistent 4d gaussian splatting for efficient dynamic scene rendering,'' in \emph{ACM SIGGRAPH Conf. Papers}, 2024, pp. 1--11.

\bibitem{lu20243d}
Z.~Lu, X.~Guo, L.~Hui, T.~Chen, M.~Yang, X.~Tang, F.~Zhu, and Y.~Dai, ``3d geometry-aware deformable gaussian splatting for dynamic view synthesis,'' in \emph{Proc. IEEE/CVF Conf. Comput. Vis. Pattern Recognit.}, 2024, pp. 8900--8910.

\bibitem{katsumata2023efficient}
K.~Katsumata, D.~M. Vo, and H.~Nakayama, ``An efficient 3d gaussian representation for monocular/multi-view dynamic scenes,'' \emph{arXiv:2311.12897}, 2023.

\bibitem{kratimenos2023dynmf}
A.~Kratimenos, J.~Lei, and K.~Daniilidis, ``Dynmf: Neural motion factorization for real-time dynamic view synthesis with 3d gaussian splatting,'' \emph{arXiv:2312.00112}, 2023.

\bibitem{lin2023gaussian}
Y.~Lin, Z.~Dai, S.~Zhu, and Y.~Yao, ``Gaussian-flow: 4d reconstruction with dynamic 3d gaussian particles,'' \emph{arXiv:2312.03431}, 2023.

\bibitem{huang2023sc}
Y.-H. Huang, Y.-T. Sun, Z.~Yang, X.~Lyu, Y.-P. Cao, and X.~Qi, ``Sc-gs: Sparse-controlled gaussian splatting for editable dynamic scenes,'' \emph{arXiv:2312.14937}, 2023.

\bibitem{li2023spacetime}
Z.~Li, Z.~Chen, Z.~Li, and Y.~Xu, ``Spacetime gaussian feature splatting for real-time dynamic view synthesis,'' \emph{arXiv:2312.16812}, 2023.

\bibitem{pumarola2021d}
A.~Pumarola, E.~Corona, G.~Pons-Moll, and F.~Moreno-Noguer, ``D-nerf: Neural radiance fields for dynamic scenes,'' in \emph{Proc. IEEE/CVF Conf. Comput. Vis. Pattern Recognit.}, 2021, pp. 10\,318--10\,327.

\bibitem{yu2023cogs}
H.~Yu, J.~Julin, Z.~{\'A}. Milacski, K.~Niinuma, and L.~A. Jeni, ``Cogs: Controllable gaussian splatting,'' \emph{arXiv:2312.05664}, 2023.

\bibitem{yang2023real}
Z.~Yang, H.~Yang, Z.~Pan, X.~Zhu, and L.~Zhang, ``Real-time photorealistic dynamic scene representation and rendering with 4d gaussian splatting,'' \emph{arXiv:2310.10642}, 2023.

\bibitem{duan20244d}
Y.~Duan, F.~Wei, Q.~Dai, Y.~He, W.~Chen, and B.~Chen, ``4d-rotor gaussian splatting: towards efficient novel view synthesis for dynamic scenes,'' in \emph{ACM SIGGRAPH Conf. Papers}, 2024, pp. 1--11.

\bibitem{lu2025manigaussian}
G.~Lu, S.~Zhang, Z.~Wang, C.~Liu, J.~Lu, and Y.~Tang, ``Manigaussian: Dynamic gaussian splatting for multi-task robotic manipulation,'' in \emph{Eur. Conf. Comput. Vis.}\hskip 1em plus 0.5em minus 0.4em\relax Springer, 2025, pp. 349--366.

\bibitem{xie2023physgaussian}
T.~Xie, Z.~Zong, Y.~Qiu, X.~Li, Y.~Feng, Y.~Yang, and C.~Jiang, ``Physgaussian: Physics-integrated 3d gaussians for generative dynamics,'' \emph{arXiv:2311.12198}, 2023.

\bibitem{guedon2023sugar}
A.~Gu{\'e}don and V.~Lepetit, ``Sugar: Surface-aligned gaussian splatting for efficient 3d mesh reconstruction and high-quality mesh rendering,'' \emph{arXiv:2311.12775}, 2023.

\bibitem{chen2023neusg}
H.~Chen, C.~Li, and G.~H. Lee, ``Neusg: Neural implicit surface reconstruction with 3d gaussian splatting guidance,'' \emph{arXiv:2312.00846}, 2023.

\bibitem{lyu20243dgsr}
X.~Lyu, Y.-T. Sun, Y.-H. Huang, X.~Wu, Z.~Yang, Y.~Chen, J.~Pang, and X.~Qi, ``3dgsr: Implicit surface reconstruction with 3d gaussian splatting,'' \emph{arXiv:2404.00409}, 2024.

\bibitem{yu2024gsdf}
M.~Yu, T.~Lu, L.~Xu, L.~Jiang, Y.~Xiangli, and B.~Dai, ``Gsdf: 3dgs meets sdf for improved rendering and reconstruction,'' \emph{arXiv:2403.16964}, 2024.

\bibitem{wang2021neus}
P.~Wang, L.~Liu, Y.~Liu, C.~Theobalt, T.~Komura, and W.~Wang, ``Neus: Learning neural implicit surfaces by volume rendering for multi-view reconstruction,'' \emph{arXiv:2106.10689}, 2021.

\bibitem{dai2024high}
P.~Dai, J.~Xu, W.~Xie, X.~Liu, H.~Wang, and W.~Xu, ``High-quality surface reconstruction using gaussian surfels,'' \emph{arXiv:2404.17774}, 2024.

\bibitem{yu2024gaussian}
Z.~Yu, T.~Sattler, and A.~Geiger, ``Gaussian opacity fields: Efficient and compact surface reconstruction in unbounded scenes,'' \emph{arXiv:2404.10772}, 2024.

\bibitem{huang20242d}
B.~Huang, Z.~Yu, A.~Chen, A.~Geiger, and S.~Gao, ``2d gaussian splatting for geometrically accurate radiance fields,'' \emph{arXiv:2403.17888}, 2024.

\bibitem{reiser2024binary}
C.~Reiser, S.~Garbin, P.~Srinivasan, D.~Verbin, R.~Szeliski, B.~Mildenhall, J.~Barron, P.~Hedman, and A.~Geiger, ``Binary opacity grids: Capturing fine geometric detail for mesh-based view synthesis,'' \emph{ACM Trans. Graph.}, vol.~43, no.~4, pp. 1--14, 2024.

\bibitem{chen2023gaussianeditor}
Y.~Chen, Z.~Chen, C.~Zhang, F.~Wang, X.~Yang, Y.~Wang, Z.~Cai, L.~Yang, H.~Liu, and G.~Lin, ``Gaussianeditor: Swift and controllable 3d editing with gaussian splatting,'' \emph{arXiv:2311.14521}, 2023.

\bibitem{palandra2024gsedit}
F.~Palandra, A.~Sanchietti, D.~Baieri, and E.~Rodol{\`a}, ``Gsedit: Efficient text-guided editing of 3d objects via gaussian splatting,'' \emph{arXiv:2403.05154}, 2024.

\bibitem{brooks2023instructpix2pix}
T.~Brooks, A.~Holynski, and A.~A. Efros, ``Instructpix2pix: Learning to follow image editing instructions,'' in \emph{Proc. IEEE/CVF Conf. Comput. Vis. Pattern Recognit.}, 2023, pp. 18\,392--18\,402.

\bibitem{fang2023gaussianeditor}
J.~Fang, J.~Wang, X.~Zhang, L.~Xie, and Q.~Tian, ``Gaussianeditor: Editing 3d gaussians delicately with text instructions,'' \emph{arXiv:2311.16037}, 2023.

\bibitem{wu2024gaussctrl}
J.~Wu, J.-W. Bian, X.~Li, G.~Wang, I.~Reid, P.~Torr, and V.~A. Prisacariu, ``Gaussctrl: Multi-view consistent text-driven 3d gaussian splatting editing,'' \emph{arXiv:2403.08733}, 2024.

\bibitem{wang2024view}
Y.~Wang, X.~Yi, Z.~Wu, N.~Zhao, L.~Chen, and H.~Zhang, ``View-consistent 3d editing with gaussian splatting,'' \emph{arXiv:2403.11868}, 2024.

\bibitem{zhang2023adding}
L.~Zhang, A.~Rao, and M.~Agrawala, ``Adding conditional control to text-to-image diffusion models,'' in \emph{Proc. IEEE/CVF Int. Conf. Comput. Vis.}, 2023, pp. 3836--3847.

\bibitem{shao2023control4d}
R.~Shao, J.~Sun, C.~Peng, Z.~Zheng, B.~Zhou, H.~Zhang, and Y.~Liu, ``Control4d: Dynamic portrait editing by learning 4d gan from 2d diffusion-based editor,'' \emph{arXiv:2305.20082}, vol.~2, no.~6, p.~16, 2023.

\bibitem{goodfellow2020generative}
I.~Goodfellow, J.~Pouget-Abadie, M.~Mirza, B.~Xu, D.~Warde-Farley, S.~Ozair, A.~Courville, and Y.~Bengio, ``Generative adversarial networks,'' \emph{Commun. ACM}, vol.~63, no.~11, pp. 139--144, 2020.

\bibitem{zhuang2024tip}
J.~Zhuang, D.~Kang, Y.-P. Cao, G.~Li, L.~Lin, and Y.~Shan, ``Tip-editor: An accurate 3d editor following both text-prompts and image-prompts,'' \emph{arXiv:2401.14828}, 2024.

\bibitem{huang2023point}
J.~Huang and H.~Yu, ``Point'n move: Interactive scene object manipulation on gaussian splatting radiance fields,'' \emph{arXiv:2311.16737}, 2023.

\bibitem{feng2024evsplitting}
Q.-Y. Feng, G.-C. Cao, H.-X. Chen, Q.-C. Xu, T.-J. Mu, R.~Martin, and S.-M. Hu, ``Evsplitting: An efficient and visually consistent splitting algorithm for 3d gaussian splatting,'' in \emph{SIGGRAPH Asia 2024 Conf. Papers}, 2024, pp. 1--11.

\bibitem{saroha2024gaussian}
A.~Saroha, M.~Gladkova, C.~Curreli, T.~Yenamandra, and D.~Cremers, ``Gaussian splatting in style,'' \emph{arXiv:2403.08498}, 2024.

\bibitem{huang2022stylizednerf}
Y.-H. Huang, Y.~He, Y.-J. Yuan, Y.-K. Lai, and L.~Gao, ``Stylizednerf: consistent 3d scene stylization as stylized nerf via 2d-3d mutual learning,'' in \emph{Proc. IEEE/CVF Conf. Comput. Vis. Pattern Recognit.}, 2022, pp. 18\,342--18\,352.

\bibitem{gao2023relightable}
J.~Gao, C.~Gu, Y.~Lin, H.~Zhu, X.~Cao, L.~Zhang, and Y.~Yao, ``Relightable 3d gaussian: Real-time point cloud relighting with brdf decomposition and ray tracing,'' \emph{arXiv:2311.16043}, 2023.

\bibitem{liang2024gs}
Z.~Liang, Q.~Zhang, Y.~Feng, Y.~Shan, and K.~Jia, ``Gs-ir: 3d gaussian splatting for inverse rendering,'' in \emph{Proc. IEEE/CVF Conf. Comput. Vis. Pattern Recognit.}, 2024, pp. 21\,644--21\,653.

\bibitem{kuang2024olat}
Z.~Kuang, Y.~Yang, S.~Dong, J.~Ma, H.~Fu, and Y.~Zheng, ``Olat gaussians for generic relightable appearance acquisition,'' in \emph{ACM SIGGRAPH Asia Conf. Papers}, 2024, pp. 1--11.

\bibitem{bi2024gs3}
Z.~Bi, Y.~Zeng, C.~Zeng, F.~Pei, X.~Feng, K.~Zhou, and H.~Wu, ``Gs3: Efficient relighting with triple gaussian splatting,'' in \emph{ACM SIGGRAPH Asia Conf. Papers}, 2024, pp. 1--12.

\bibitem{ye2023gaussian}
M.~Ye, M.~Danelljan, F.~Yu, and L.~Ke, ``Gaussian grouping: Segment and edit anything in 3d scenes,'' \emph{arXiv:2312.00732}, 2023.

\bibitem{zhou2023feature}
S.~Zhou, H.~Chang, S.~Jiang, Z.~Fan, Z.~Zhu, D.~Xu, P.~Chari, S.~You, Z.~Wang, and A.~Kadambi, ``Feature 3dgs: Supercharging 3d gaussian splatting to enable distilled feature fields,'' \emph{arXiv:2312.03203}, 2023.

\bibitem{lan20232d}
K.~Lan, H.~Li, H.~Shi, W.~Wu, Y.~Liao, L.~Wang, and P.~Zhou, ``2d-guided 3d gaussian segmentation,'' \emph{arXiv:2312.16047}, 2023.

\bibitem{dou2024cosseggaussians}
B.~Dou, T.~Zhang, Y.~Ma, Z.~Wang, and Z.~Yuan, ``Cosseggaussians: Compact and swift scene segmenting 3d gaussians,'' \emph{arXiv:2401.05925}, 2024.

\bibitem{gu2025egolifter}
Q.~Gu, Z.~Lv, D.~Frost, S.~Green, J.~Straub, and C.~Sweeney, ``Egolifter: Open-world 3d segmentation for egocentric perception,'' in \emph{Eur. Conf. Comput. Vis.}\hskip 1em plus 0.5em minus 0.4em\relax Springer, 2025, pp. 382--400.

\bibitem{choi2025click}
S.~Choi, H.~Song, J.~Kim, T.~Kim, and H.~Do, ``Click-gaussian: Interactive segmentation to any 3d gaussians,'' in \emph{Eur. Conf. Comput. Vis.}\hskip 1em plus 0.5em minus 0.4em\relax Springer, 2025, pp. 289--305.

\bibitem{yue2025improving}
Y.~Yue, A.~Das, F.~Engelmann, S.~Tang, and J.~E. Lenssen, ``Improving 2d feature representations by 3d-aware fine-tuning,'' in \emph{Eur. Conf. Comput. Vis}.\hskip 1em plus 0.5em minus 0.4em\relax Springer, 2025, pp. 57--74.

\bibitem{shi2023language}
J.-C. Shi, M.~Wang, H.-B. Duan, and S.-H. Guan, ``Language embedded 3d gaussians for open-vocabulary scene understanding,'' \emph{arXiv:2311.18482}, 2023.

\bibitem{qin2023langsplat}
M.~Qin, W.~Li, J.~Zhou, H.~Wang, and H.~Pfister, ``Langsplat: 3d language gaussian splatting,'' \emph{arXiv:2312.16084}, 2023.

\bibitem{zuo2024fmgs}
X.~Zuo, P.~Samangouei, Y.~Zhou, Y.~Di, and M.~Li, ``Fmgs: Foundation model embedded 3d gaussian splatting for holistic 3d scene understanding,'' \emph{arXiv:2401.01970}, 2024.

\bibitem{radford2021learning}
A.~Radford, J.~W. Kim, C.~Hallacy, A.~Ramesh, G.~Goh, S.~Agarwal, G.~Sastry, A.~Askell, P.~Mishkin, J.~Clark \emph{et~al.}, ``Learning transferable visual models from natural language supervision,'' in \emph{Proc. Int. Conf. Mach. Learn.}\hskip 1em plus 0.5em minus 0.4em\relax PMLR, 2021, pp. 8748--8763.

\bibitem{caron2021emerging}
M.~Caron, H.~Touvron, I.~Misra, H.~J{\'e}gou, J.~Mairal, P.~Bojanowski, and A.~Joulin, ``Emerging properties in self-supervised vision transformers,'' in \emph{Proc. IEEE/CVF Int. Conf. Comput. Vis.}, 2021, pp. 9650--9660.

\bibitem{van2017neural}
A.~Van Den~Oord, O.~Vinyals \emph{et~al.}, ``Neural discrete representation learning,'' \emph{Adv. Neural Inf. Process. Syst.}, vol.~30, 2017.

\bibitem{de2020material}
A.~De~Vaucorbeil, V.~P. Nguyen, S.~Sinaie, and J.~Y. Wu, ``Material point method after 25 years: Theory, implementation, and applications,'' \emph{Adv. Appl. Mech.}, vol.~53, pp. 185--398, 2020.

\bibitem{zhang2025physdreamer}
T.~Zhang, H.-X. Yu, R.~Wu, B.~Y. Feng, C.~Zheng, N.~Snavely, J.~Wu, and W.~T. Freeman, ``Physdreamer: Physics-based interaction with 3d objects via video generation,'' in \emph{Eur. Conf. Comput. Vis}.\hskip 1em plus 0.5em minus 0.4em\relax Springer, 2025, pp. 388--406.

\bibitem{jiang2024vr}
Y.~Jiang, C.~Yu, T.~Xie, X.~Li, Y.~Feng, H.~Wang, M.~Li, H.~Lau, F.~Gao, Y.~Yang \emph{et~al.}, ``Vr-gs: A physical dynamics-aware interactive gaussian splatting system in virtual reality,'' in \emph{ACM SIGGRAPH 2024 Conf. Papers}, 2024, pp. 1--1.

\bibitem{jung2024relaxing}
J.~Jung, J.~Han, H.~An, J.~Kang, S.~Park, and S.~Kim, ``Relaxing accurate initialization constraint for 3d gaussian splatting,'' \emph{arXiv:2403.09413}, 2024.

\bibitem{das2023neural}
D.~Das, C.~Wewer, R.~Yunus, E.~Ilg, and J.~E. Lenssen, ``Neural parametric gaussians for monocular non-rigid object reconstruction,'' \emph{arXiv:2312.01196}, 2023.

\bibitem{gao2024mesh}
L.~Gao, J.~Yang, B.-T. Zhang, J.-M. Sun, Y.-J. Yuan, H.~Fu, and Y.-K. Lai, ``Mesh-based gaussian splatting for real-time large-scale deformation,'' \emph{arXiv:2402.04796}, 2024.

\bibitem{matsuki2023gaussian}
H.~Matsuki, R.~Murai, P.~H. Kelly, and A.~J. Davison, ``Gaussian splatting slam,'' \emph{arXiv:2312.06741}, 2023.

\bibitem{lei2023gart}
J.~Lei, Y.~Wang, G.~Pavlakos, L.~Liu, and K.~Daniilidis, ``Gart: Gaussian articulated template models,'' \emph{arXiv:2311.16099}, 2023.

\bibitem{huang2024gs++}
L.~Huang, J.~Bai, J.~Guo, and Y.~Guo, ``Gs++: Error analyzing and optimal gaussian splatting,'' \emph{arXiv:2402.00752}, 2024.

\bibitem{moenne20243d}
N.~Moenne-Loccoz, A.~Mirzaei, O.~Perel, R.~de~Lutio, J.~Martinez~Esturo, G.~State, S.~Fidler, N.~Sharp, and Z.~Gojcic, ``3d gaussian ray tracing: Fast tracing of particle scenes,'' \emph{ACM Trans. Graph.}, vol.~43, no.~6, pp. 1--19, 2024.

\bibitem{li2024geogaussian}
Y.~Li, C.~Lyu, Y.~Di, G.~Zhai, G.~H. Lee, and F.~Tombari, ``Geogaussian: Geometry-aware gaussian splatting for scene rendering,'' \emph{arXiv:2403.11324}, 2024.

\bibitem{xiong2023sparsegs}
H.~Xiong, S.~Muttukuru, R.~Upadhyay, P.~Chari, and A.~Kadambi, ``Sparsegs: Real-time 360° sparse view synthesis using gaussian splatting,'' \emph{arXiv:2312.00206}, 2023.

\bibitem{bolanos2024gaussian}
L.~Bolanos, S.-Y. Su, and H.~Rhodin, ``Gaussian shadow casting for neural characters,'' in \emph{Proc. IEEE/CVF Conf. Comput. Vis. Pattern Recognit.}, 2024, pp. 20\,997--21\,006.

\bibitem{nichol2022point}
A.~Nichol, H.~Jun, P.~Dhariwal, P.~Mishkin, and M.~Chen, ``Point-e: A system for generating 3d point clouds from complex prompts,'' \emph{arXiv:2212.08751}, 2022.

\bibitem{liu2023syncdreamer}
Y.~Liu, C.~Lin, Z.~Zeng, X.~Long, L.~Liu, T.~Komura, and W.~Wang, ``Syncdreamer: Generating multiview-consistent images from a single-view image,'' \emph{arXiv:2309.03453}, 2023.

\bibitem{li2024controllable}
Z.~Li, Y.~Chen, L.~Zhao, and P.~Liu, ``Controllable text-to-3d generation via surface-aligned gaussian splatting,'' \emph{arXiv:2403.09981}, 2024.

\bibitem{armandpour2023re}
M.~Armandpour, A.~Sadeghian, H.~Zheng, A.~Sadeghian, and M.~Zhou, ``Re-imagine the negative prompt algorithm: Transform 2d diffusion into 3d, alleviate janus problem and beyond,'' \emph{arXiv:2304.04968}, 2023.

\bibitem{hu2021lora}
E.~J. Hu, Y.~Shen, P.~Wallis, Z.~Allen-Zhu, Y.~Li, S.~Wang, L.~Wang, and W.~Chen, ``Lora: Low-rank adaptation of large language models,'' \emph{arXiv:2106.09685}, 2021.

\bibitem{sun2024high}
S.~Sun, M.~Mielle, A.~J. Lilienthal, and M.~Magnusson, ``High-fidelity slam using gaussian splatting with rendering-guided densification and regularized optimization,'' \emph{arXiv:2403.12535}, 2024.

\bibitem{liu2025citygaussian}
Y.~Liu, C.~Luo, L.~Fan, N.~Wang, J.~Peng, and Z.~Zhang, ``Citygaussian: Real-time high-quality large-scale scene rendering with gaussians,'' in \emph{Eur. Conf. Comput. Vis}.\hskip 1em plus 0.5em minus 0.4em\relax Springer, 2025, pp. 265--282.

\bibitem{kerbl2024hierarchical}
B.~Kerbl, A.~Meuleman, G.~Kopanas, M.~Wimmer, A.~Lanvin, and G.~Drettakis, ``A hierarchical 3d gaussian representation for real-time rendering of very large datasets,'' \emph{ACM Trans. Graph.}, vol.~43, no.~4, pp. 1--15, 2024.

\bibitem{zhang2024pixel}
Z.~Zhang, W.~Hu, Y.~Lao, T.~He, and H.~Zhao, ``Pixel-gs: Density control with pixel-aware gradient for 3d gaussian splatting,'' \emph{arXiv:2403.15530}, 2024.

\bibitem{feng2024new}
Q.~Feng, G.~Cao, H.~Chen, T.-J. Mu, R.~R. Martin, and S.-M. Hu, ``A new split algorithm for 3d gaussian splatting,'' \emph{arXiv:2403.09143}, 2024.

\bibitem{niemeyer2024radsplat}
M.~Niemeyer, F.~Manhardt, M.-J. Rakotosaona, M.~Oechsle, D.~Duckworth, R.~Gosula, K.~Tateno, J.~Bates, D.~Kaeser, and F.~Tombari, ``Radsplat: Radiance field-informed gaussian splatting for robust real-time rendering with 900+ fps,'' \emph{arXiv:2403.13806}, 2024.

\bibitem{fang2024mini}
G.~Fang and B.~Wang, ``Mini-splatting: Representing scenes with a constrained number of gaussians,'' \emph{arXiv:2403.14166}, 2024.

\bibitem{kazhdan2006poisson}
M.~Kazhdan, M.~Bolitho, and H.~Hoppe, ``Poisson surface reconstruction,'' in \emph{Proc. Eurographics Symp. Geom. Process. (SGP)}, vol.~7, no.~4, 2006.

\bibitem{tang2023delicate}
J.~Tang, H.~Zhou, X.~Chen, T.~Hu, E.~Ding, J.~Wang, and G.~Zeng, ``Delicate textured mesh recovery from nerf via adaptive surface refinement,'' in \emph{Proc. IEEE/CVF Int. Conf. Comput. Vis.}, 2023, pp. 17\,739--17\,749.

\bibitem{li2024ggrt}
H.~Li, Y.~Gao, D.~Zhang, C.~Wu, Y.~Dai, C.~Zhao, H.~Feng, E.~Ding, J.~Wang, and J.~Han, ``Ggrt: Towards generalizable 3d gaussians without pose priors in real-time,'' \emph{arXiv:2403.10147}, 2024.

\bibitem{malarzgaussian}
D.~Malarz, W.~Smolak, J.~Tabor, S.~Tadeja, and P.~Spurek, ``Gaussian splatting with nerf-based color and opacity.''

\bibitem{zuffi20173d}
S.~Zuffi, A.~Kanazawa, D.~W. Jacobs, and M.~J. Black, ``3d menagerie: Modeling the 3d shape and pose of animals,'' in \emph{Proc. IEEE Conf. Comput. Vis. Pattern Recognit.}, 2017, pp. 6365--6373.

\bibitem{porumbescu2005shell}
S.~D. Porumbescu, B.~Budge, L.~Feng, and K.~I. Joy, ``Shell maps,'' \emph{ACM Trans. Graph.}, vol.~24, no.~3, pp. 626--633, 2005.

\bibitem{sumner2007embedded}
R.~W. Sumner, J.~Schmid, and M.~Pauly, ``Embedded deformation for shape manipulation,'' in \emph{ACM SIGGRAPH 2007 Papers}, 2007, pp. 80--es.

\bibitem{fan2024instantsplat}
Z.~Fan, W.~Cong, K.~Wen, K.~Wang, J.~Zhang, X.~Ding, D.~Xu, B.~Ivanovic, M.~Pavone, G.~Pavlakos \emph{et~al.}, ``Instantsplat: Unbounded sparse-view pose-free gaussian splatting in 40 seconds,'' \emph{IEEE Trans. Circuits Syst. Video Technol.}, 2024.

\bibitem{sun2023icomma}
Y.~Sun, X.~Wang, Y.~Zhang, J.~Zhang, C.~Jiang, Y.~Guo, and F.~Wang, ``icomma: Inverting 3d gaussians splatting for camera pose estimation via comparing and matching,'' \emph{arXiv:2312.09031}, 2023.

\bibitem{martin2021nerf}
R.~Martin-Brualla, N.~Radwan, M.~S. Sajjadi, J.~T. Barron, A.~Dosovitskiy, and D.~Duckworth, ``Nerf in the wild: Neural radiance fields for unconstrained photo collections,'' in \emph{Proc. IEEE/CVF Conf. Comput. Vis. Pattern Recognit.}, 2021, pp. 7210--7219.

\bibitem{chen2022hallucinated}
X.~Chen, Q.~Zhang, X.~Li, Y.~Chen, Y.~Feng, X.~Wang, and J.~Wang, ``Hallucinated neural radiance fields in the wild,'' in \emph{Proc. IEEE/CVF Conf. Comput. Vis. Pattern Recognit.}, 2022, pp. 12\,943--12\,952.

\bibitem{yang2023cross}
Y.~Yang, S.~Zhang, Z.~Huang, Y.~Zhang, and M.~Tan, ``Cross-ray neural radiance fields for novel-view synthesis from unconstrained image collections,'' in \emph{Proc. IEEE/CVF Int. Conf. Comput. Vis. (ICCV)}, 2023, pp. 15\,901--15\,911.

\bibitem{dahmani2024swag}
H.~Dahmani, M.~Bennehar, N.~Piasco, L.~Roldao, and D.~Tsishkou, ``Swag: Splatting in the wild images with appearance-conditioned gaussians,'' \emph{arXiv:2403.10427}, 2024.

\bibitem{zhang2024gaussian}
D.~Zhang, C.~Wang, W.~Wang, P.~Li, M.~Qin, and H.~Wang, ``Gaussian in the wild: 3d gaussian splatting for unconstrained image collections,'' \emph{arXiv:2403.15704}, 2024.

\bibitem{fu2023colmap}
Y.~Fu, S.~Liu, A.~Kulkarni, J.~Kautz, A.~A. Efros, and X.~Wang, ``Colmap-free 3d gaussian splatting,'' \emph{arXiv:2312.07504}, 2023.

\bibitem{cai2024gs}
D.~Cai, J.~Heikkil{\"a}, and E.~Rahtu, ``Gs-pose: Cascaded framework for generalizable segmentation-based 6d object pose estimation,'' \emph{arXiv:2403.10683}, 2024.

\bibitem{saito2024relightable}
S.~Saito, G.~Schwartz, T.~Simon, J.~Li, and G.~Nam, ``Relightable gaussian codec avatars,'' in \emph{Proc. IEEE/CVF Conf. Comput. Vis. Pattern Recognit.}, 2024, pp. 130--141.

\bibitem{franke2024trips}
L.~Franke, D.~Rückert, L.~Fink, and M.~Stamminger, ``Trips: Trilinear point splatting for real-time radiance field rendering,'' in \emph{Comput. Graph. Forum}.\hskip 1em plus 0.5em minus 0.4em\relax Wiley Online Library, 2024, p. e15012.

\end{thebibliography}

\vfill

\end{document}